\documentclass[12pt]{iopart}
\usepackage{hyperref, amssymb}
\usepackage{multirow}
\usepackage{graphicx}
\usepackage{caption}
\usepackage{enumerate}
\usepackage{mdwmath}
\usepackage{mdwtab}

\usepackage{algorithmic}
\usepackage{amssymb}
\usepackage{array}
\usepackage{fixltx2e}
\usepackage{tabularx}
\usepackage{xcolor}

  \expandafter\let\csname equation*\endcsname\relax
  \expandafter\let\csname endequation*\endcsname\relax
\usepackage[cmex10]{amsmath}

\begin{document}
\nocite{*}

\title[FRIST Learning and Applications]{FRIST - Flipping and Rotation Invariant Sparsifying Transform Learning and Applications}


\author{Bihan~Wen$^{1}$, Saiprasad~Ravishankar $^{2}$, and Yoram~Bresler$^{1}$}

\address{$^{1}$ Department of Electrical and Computer Engineering and the Coordinated Science Laboratory, University of Illinois at Urbana-Champaign, Champaign, IL 61801, USA.\\
$^{2}$ Department of Electrical Engineering and Computer Science \\
University of Michigan, Ann Arbor, MI 48109, USA.}
\ead{\mailto{bwen3@illinois.edu}, \mailto{ravisha@umich.edu}, and \mailto{ybresler@illinois.edu}}

\vspace{10pt}
\begin{indented}
\item[]April 2017
\end{indented}

\begin{abstract}
Features based on sparse representation, especially using the synthesis dictionary model, have been heavily exploited in signal processing and computer vision. However, synthesis dictionary learning typically involves NP-hard sparse coding and expensive learning steps. Recently, sparsifying transform learning received interest for its cheap computation and its optimal updates in the alternating algorithms. In this work, we develop a methodology for learning Flipping and Rotation Invariant Sparsifying Transforms, dubbed FRIST, to better represent natural images that contain textures with various geometrical directions. The proposed alternating FRIST learning algorithm involves efficient optimal updates. 
We provide a convergence guarantee, and demonstrate the empirical convergence behavior of the proposed FRIST learning approach. Preliminary experiments show the promising performance of FRIST learning for sparse image representation, segmentation, denoising, robust inpainting, and compressed sensing-based magnetic resonance image reconstruction.
\end{abstract}

%
\vspace{2pc}

\section{Introduction} \label{intro}

\subsection{Sparse Modeling}
Sparse representation of natural signals in a certain transform domain or dictionary has been widely exploited. Various sparse signal models, such as the synthesis model \cite{ambruck, elmiru} and the transform model \cite{tfcode} have been studied. The popular \textit{synthesis model} suggests that a signal $y \in \mathbb{R}^{n}$ can be sparsely represented as $y = D x + \eta$, where $D \in \mathbb{R}^{n \times m}$ is a synthesis dictionary, $x \in \mathbb{R}^{m}$ is a sparse code, and $\eta$ is a small approximation error in the signal domain. 
Synthesis dictionary learning methods \cite{eng, elad} that adapt the dictionary based on training data typically involve a synthesis sparse coding step which is often NP-hard \cite{npa}, so that approximate solutions \cite{pati, mp2, chen2} are widely used.
Various dictionary learning algorithms \cite{eng, Yagh, skret, Mai} have been proposed and are popular in numerous applications such as denoising, inpainting, deblurring, and demosaicing \cite{elad2, elad3, elad5}.
For example, the well-known K-SVD method \cite{elad} generalizes the K-means clustering process to a dictionary learning algorithm, and alternates between updating the sparse codes of training signals (sparse coding step) and the dictionary (dictionary or codebook update step). K-SVD updates both the dictionary atoms (columns) and the non-zero entries in the sparse codes (with fixed support) in the dictionary update step using singular value decompositions (SVD).
However, the dictionary learning algorithms such as K-SVD are usually computationally expensive for large-scale problems. Moreover, methods such as KSVD lack convergence guarantees, and can get easily caught in local minima, or saddle points \cite{Rubd}.

The alternative \textit{transform model} suggests that the signal $y$ is approximately sparsifiable using a transform $W \in \mathbb{R}^{m \times n}$, i.e., $ W y = x + e $, with $x \in \mathbb{R}^{m} $ sparse in some sense and $ e $ a small approximation error in the transform domain (rather than in the signal domain). It is well-known that natural images are sparsifiable by analytical transforms such as the discrete cosine transform (DCT), or wavelet transform \cite{wav}. Furthermore, recent works proposed learning square sparsifying transforms (SST) \cite{sabres}, which turn out to be advantageous in various applications such as image denoising, magnetic resonance imaging (MRI), and computed tomography (CT) \cite{sabres, syber, luke4, Pfister2015}. Alternating minimization algorithms for learning SST have been proposed with cheap and optimal updates \cite{sabres3}.

Since SST learning is restricted to one adaptive square transform for all the data, the diverse patches of natural images may not be sufficiently sparsified in the SST model. 
Recent work focused on learning a union of unstructured sparsifying transforms \cite{wensabres, wenICIP2014}, dubbed OCTOBOS (for OverComplete
TransfOrm with BlOck coSparsity constraint -- cf. \cite{wensabres}), to sparsify images with diverse contents, features and textures. 
Given a signal $y \in \mathbb{R}^{n}$ and a union of transforms $\left \{ W_k \right \}_{k=1}^{K}$, where each $W_k \in \mathbb{R}^{m \times n}$, the OCTOBOS model selects the best matching transform for $y$ as the one providing the minimum transform-domain modeling error. The OCTOBOS sparse coding problem is the following:
\begin{align} 
 \nonumber (\mathrm{P0}) \;\; & \min_{1 \leq k \leq K}\: \min_{x^{k}}\: \left \| W_k \, y - x^{k} \right \|_{2}^{2} \;\; s.t.\; \:  \left \| x^{k} \right \|_{0}\leq s\; \: \forall \,\,  k,
\end{align} 
where $x^{k}$ denotes the sparse representation for $y$ in the transform $W_k$, the $\ell_{0}$ ``norm" counts the number of non-zeros in a vector, and $s$ is a given sparsity level. 
However, learning such an unstructured OCTOBOS model \cite{wensabres} (that has many free parameters) especially from noisy or limited data could suffer from overfitting to noise/artifacts, thereby degrading performance in various applications and inverse problem settings.

Instead, in this paper, we consider the use of transformation symmetries to constrain the multiple learned transforms, thus reducing the number of free parameters and avoiding overfitting. While previous works exploited transformation symmetries in synthesis model sparse coding \cite{rotSC}, and applied rotational operators with analytical transforms \cite{zhan2015fast}, the usefulness of the rotational invariance property in learning adaptive sparse signal models has not been explored. Here, we propose a Flipping and Rotation Invariant Sparsifying Transform (FRIST) learning scheme, and show that it can provide better sparse representations by capturing the ``optimal" orientations of patches in natural images. As such, it serves as an effective regularizer for image recovery in various inverse problems. Preliminary experiments in this paper show the usefulness of adaptive sparse representation by FRIST for sparse image representation, segmentation, denoising, robust inpainting, and compressed sensing-based magnetic resonance image (MRI) reconstruction with promising performances.

\subsection{Highlights and Organization}
We summarize some important features and contributions of this work as follows:
\begin{itemize}
	\item We propose a FRIST model that exploits the flipping and rotation invariance property of natural images, i.e., image patches typically contain edges and features at different orientations, and hence a (single) common transform could be learned for (optimally) flipped or rotated versions of patches. Compared to a general overcomplete synthesis dictionary model or OCTOBOS, FRIST is much more constrained (with the constraints reflecting commonly observed image properties), which proves beneficial in various applications and inverse problem settings involving limited or highly corrupted measurements.
\vspace{-0.05in}	
	\item We propose a novel problem formulation and an efficient algorithm for learning FRIST. All steps of our alternating minimization algorithm involve simple optimal updates. We also provide a convergence analysis of the proposed FRIST learning algorithm.
\vspace{-0.05in}
	\item We propose various adaptive FRIST-based inverse problem formulations along with algorithms for these problems, including for image denoising, inpainting, and compressed sensing-based image reconstruction. The proposed FRIST-based algorithms outperform the previous adaptive sparse modeling methods.
\end{itemize}

The adaptive FRIST algorithms in this work involve clustering the image patches. Such clustering naturally arises in the proposed alternating minimization algorithms. There are other prior works such as the Structured Sparse Model Selection (SSMS) \cite{yu2010pca, yu2012pca} approach that also involve clustering, but the sub-dictionary in each cluster in SSMS is obtained by conventional Principal Component Analysis (PCA). Though both SSMS and FRIST impose more structure to reduce the degrees of freedom in the underlying model, the proposed FRIST learning involves a variational formulation that automatically enables joint sparse coding, clustering (based on flipping and rotations), and (a single, small parent) transform learning from training signals. Moreover, our FRIST learning algorithm comes with convergence guarantees.

We organize the rest of the paper as follows. Section \ref{learning} introduces the FRIST model and the learning formulation. In Section \ref{algoConv}, we present efficient FRIST learning algorithms to solve the proposed learning problem along with convergence analysis. Section \ref{app} describes various applications based on FRIST learning, including image denoising, inpainting and compressed sensing-based (MR) image reconstruction. Section \ref{exp} provides experimental results demonstrating the promise of FRIST learning, including for image segmentation, sparse representation, denoising, inpainting, and MRI.

\section{FRIST Model and Its Learning Formulation} \label{learning}

\subsection{FRIST Model}
The learning of the sparsifying transform model \cite{sabres} has been proposed recently. 
Extending this approach, we propose a \textit{FRIST model} that first applies a flipping and rotation (FR) operator $\Phi \in \mathbb{R}^{n \times n}$ to a signal $y \in \mathbb{R}^{n}$,
and models $\Phi y$ as approximately sparsifiable by some sparsifying transform $W \in \mathbb{R}^{m \times n}$, i.e., $W \Phi y = x + e$, with $x \in \mathbb{R}^{m} $ sparse in some sense, and $ e $ is a small deviation term. A finite set of flipping and rotation operators $\left \{ \Phi_{k} \right \}_{k=1}^{K}$ is considered, and the sparse coding problem in the FRIST model is as follows:
\begin{align} 
 \nonumber (\mathrm{P1}) \;\; & \min_{1 \leq k \leq K}\: \min_{x^{k}}\: \left \| W\,\Phi_k \, y - x^{k} \right \|_{2}^{2} \;\; \text{s.t.}\; \:  \left \| x^{k} \right \|_{0}\leq s\; \: \forall \,\,  k.
\end{align} 
Thus, $x^{k}$ denotes the sparse code of $\Phi_k\,y$ in the transform $W$ domain, with maximum sparsity $s$. Equivalently, the optimal $\hat{x}^{\hat{k}}$ (achieving the minimum over all $k$) is called the optimal sparse code in the FRIST domain.
We further decompose the FR matrix as $\Phi_k \triangleq G_q\,F$, where $F$ can be either an identity matrix, or (for 2D signals) a left-to-right flipping permutation matrix. Though there are various methods of formulating the rotation operator $G$ with arbitrary angles \cite{lowe1999sift, ke2004pcasift}, rotating image patches by an angle $\theta$ that is not a multiple of $90^{\circ}$ requires interpolation, and may result in misalignment with the pixel grid. 
Here, we adopt the matrix $G_{q} \triangleq G(\theta_q)$ that permutes the pixels in an image patch approximating rotation by angle $\theta_q$ without interpolation.  Constructions of such $ \left \{ G_{q} \right \}$ have been proposed before \cite{le2005bandelet, qu2012undersampled, zhan2015fast}. With such implementation,  the number of possible permutations  $ \left \{ G_{q} \right \}$ denoted by $\tilde{Q}$, is finite and grows linearly with the signal dimension $n$. Accounting for the flipping operation, the total number of possible FR operators is $\tilde{K} = 2\tilde{Q}$.

In practice, one can select a subset $ \left \{ \Phi_{k} \right \}_{k=1}^{K}$, containing $K < \tilde{K}$ of FR candidates, from which the optimal $\hat{\Phi} = \Phi_{\hat{k}}$ is chosen in $(\mathrm{P1})$.
For each $\Phi_k$, the optimal sparse code $\hat{x}^{k}$ in Problem $(\mathrm{P1})$ can be solved exactly as $\hat{x}^{k} = H_{s}(W \Phi_k y)$, where $H_{s}(\cdot)$ is the projector onto the $s$-$\ell_{0}$ ball \cite{doubsp2l}, i.e., $H_{s}(b)$ zeros out all but the $s$ elements of largest magnitude in $b \in \mathbb{R}^{m}$. The optimal FR operator $\Phi_{\hat{k}}$ is selected to provide the smallest sparsification (modeling) error $\left \| W\,\Phi_{k} \, y - H_{s}(W\Phi_{k} y) \right \|_{2}^{2}$ over $k$ in Problem $(\mathrm{P1})$.

The FRIST model can be interpreted as a structured union-of-transforms model, or a structured OCTOBOS model \cite{wensabres}, i.e., compared to OCTOBOS, FRIST is much more constrained, with fewer free parameters. In particular, the OCTOBOS model involves a collection (or union) of sparsifying transforms $\left\{ W_{k}\right\} _{k=1}^{K}$ such that for each candidate signal, there is a transform in the collection that is best matched (or that provides the lowest sparsification error) to the signal. The FRIST model involves a collection of transforms $\left\{ W\Phi_{k}\right\} _{k=1}^{K}$ (as in Problem (P1)) that are related to each other by rotation and flip operators (and involving a single parent transform $W$). 
The transforms in the collection all share a common transform $W$. We call the shared common transform $W$ the \textbf{parent transform}, and each generated $W_k = W \Phi_k$ is called a \textbf{child transform}. 
Clearly, the collection of transforms in FRIST is more constrained than in the OCTOBOS model.
The constraints that are imposed by FRIST are devised to reflect commonly observed properties of natural image patches, i.e., image patches tend to have edges and features at various orientations, and optimally rotating (or flipping) each patch would allow it to be well-sparsified by a common sparsifying transform $W$ (as in $(\mathrm{P1})$).
This property turns out to be useful in inverse problems such as denoising and inpainting, preventing the overfitting of the model in the presence of limited or highly corrupted data or measurements.

Problem $\mathrm{(P1)}$ is similar to the OCTOBOS sparse coding problem \cite{wensabres}, where each $W_k = W \Phi_k$ corresponds to a block of OCTOBOS. 
Similar to the clustering procedure in OCTOBOS, Problem $\mathrm{(P1)}$ matches a signal $y$ to a particular child transform $W_k$ with its directional FR operator $\Phi_k$. Thus, FRIST is potentially capable of automatically clustering a collection of signals (e.g., image patches), but according to their geometric orientations. When the parent transform $W$ is unitary, FRIST is also equivalent to an overcomplete synthesis dictionary with block sparsity \cite{blockSparse}, with $W_{k}^{T}$ denoting the $k$th block of the equivalent overcomplete dictionary.

\subsection{FRIST Learning Formulation} 
Generally, the parent transform $W$ can be overcomplete \cite{lost2013, wensabres, Pfister2015}. In this work, we restrict ourselves to learning FRIST with a square parent transform $W$ (i.e., $m=n$), which leads to a highly efficient learning algorithm with optimal updates. Note that the FRIST model is still overcomplete, even with a square parent $W$, because of the additional FR operators.
Given the training data $Y \in \mathbb{R}^{n \times N}$, we formulate the FRIST learning problem as follows:
\begin{align} 
 \nonumber (\mathrm{P2}) & \min_{W, \left \{X_{i}\right \}, \left \{C_{k} \right \}}\: \sum _{k=1}^K  \sum_{i \in C_{k}} \left \| W \Phi_k Y_{i}-X_{i} \right \|_{2}^{2} + \lambda Q(W)\\
 \nonumber & \,\,\,\,\,\,\,\,\,\, \text{s.t.}\; \:  \left \| X_{i} \right \|_{0}\leq s\; \: \forall \,\,  i, \,\,\, \left\{C_{k}\right\} \in \Gamma
\end{align} 
where $\left \{ X_i \right \}$ represent the FRIST-domain sparse codes of the corresponding columns $\left \{ Y_i \right \}$ of $Y$, and $X \in \mathbb{R}^{n \times N}$ with columns $X_i$ denotes the transform sparse code matrix of $Y$. 
The $\left\{C_{k}\right\}_{k = 1}^K$ indicate a clustering of the signals $\left \{ Y_{i} \right \}_{i = 1}^N$ such that $C_j$ contains the indices of signals in the $j$th cluster (corresponding to the child transform $W \Phi_j$), and each signal $Y_i$ is associated exactly with one FR operator $\Phi_{k}$. The set $\Gamma$ is the set of all possible partitions (into $K$ subsets) of the set of integers $\left \{1, 2, ..., N  \right \}$, which enforces all of the $C_{k}$'s to be disjoint \cite{wensabres}.

Problem $\mathrm{(P2)}$ is to minimize the FRIST learning objective that includes the modeling or sparsification error $\sum _{k=1}^K  \sum_{i \in C_{k}} \left \| W \Phi_k Y_{i}-X_{i} \right \|_{2}^{2}$ for $Y$ as well as the regularizer $Q(W)= - \log \,\left | \mathrm{det \,} W \right |  + \left \| W \right \|_{F}^{2}$ to prevent trivial solutions \cite{sabres}. Here, the negative log-determinant penalty $- \log \,\left | \mathrm{det \,} W \right |$ enforces full rank on $W$, and the $\left \| W \right \|_{F}^{2}$ penalty helps remove a `scale ambiguity' in the solution. The regularizer $Q(W)$ fully controls the condition number and scaling of the learned parent transform \cite{sabres}. The regularizer weight $\lambda$ is chosen as $\lambda = \lambda_{0} \left \| Y \right \|_{F}^{2}$ with $\lambda_{0} >0$, in order to scale with the first term in $\mathrm{(P2)}$. Previous works \cite{sabres} showed that the condition number and spectral norm of the optimal parent transform $\hat{W}$ approach $1$ and $1/\sqrt{2}$ respectively, as $\lambda_{0} \to \infty $ in $\mathrm{(P2)}$.

Problem $\mathrm{(P2)}$ imposes an $\ell_{0}$ sparsity constraint $\left \| X_{i} \right \|_{0}\leq s$ on the sparse code of each signal or image patch. One can also impose an overall (or aggregate) sparsity constraint on the entire sparse code matrix $X$ to allow variable sparsity levels across the signals (see Section \ref{MRIform}). 
Alternatively, a sparsity penalty method can be used, instead of imposing sparsity constraints, which also leads to efficient algorithms (see Section \ref{inpaintingForm}). We will demonstrate the use of various sparsity methods in Section \ref{app}.

Note that in the overcomplete synthesis dictionary model, sparse coding with an $\ell_{0}$ ``norm" constraint is NP-hard in general, and convex $\ell_{1}$ relaxations of the synthesis sparse coding problem have been popular, and solving such an $\ell_{1}$ (relaxed) problem is known to provide the sparsest solution under certain conditions on the dictionary. 
On the other hand, in the sparsifying transform model (including in the FRIST model), the sparse coding problem can be solved exactly and cheaply by thresholding operations, irrespective of whether an $\ell_{0}$ penalty or constraint (resulting in hard thresholding-type solution) or an $\ell_{1}$ penalty (resulting in soft thresholding solution) is used. 
Thus there is not a computational benefit for employing the $\ell_{1}$ norm in case of the transform model. More importantly, in practice, we have observed that transform learning with $\ell_0$ sparsity leads to better performance in applications compared to $\ell_{1}$ norm-based learning.

\section{FRIST Learning Algorithm and Convergence Analysis} \label{algoConv}

\subsection{FRIST Learning Algorithm} \label{algorithm}
We propose an efficient algorithm for solving $\mathrm{(P2)}$, which alternates between a \emph{sparse coding and clustering} step, and a \emph{transform update} step.

\textbf{Sparse Coding and Clustering}. Given the training matrix $Y$, and a fixed parent transform $W$, we solve the following Problem $\mathrm{(P3)}$  for the sparse codes and clusters:
\begin{align}
 \nonumber (\mathrm{P3}) \;\; \min_{\left \{ C_{k} \right \}, \left \{ X_{i} \right \}}\: \sum _{k=1}^K  \sum_{i \in C_{k}} \left \| W \Phi_k Y_{i}-X_{i} \right \|_{2}^{2} \;\,\,\, s.t.\; \:  \left \| X_{i} \right \|_{0}\leq s\; \: \forall \,\,  i, \,\,\, \left\{C_{k}\right\} \in \Gamma.
\end{align} 
The modeling error $ \left \| W \Phi_k Y_{i}-X_{i} \right \|_{2}^{2}$ serves as the clustering measure corresponding to signal $Y_i$, where the best sparse code corresponding to FR permutation $\Phi_k$ \footnote{The FR operator is $\Phi_k = G_q F$, where both $G_q$ and $F$ are permutation matrices. Therefore the composite operator $\Phi_k$ is a permutation matrix.} is $X_i = H_s (W \Phi_k Y_i)$.  Problem $\mathrm{(P3)}$ is clearly equivalent to finding the ``optimal" FR permutation $\Phi_{\hat{k_i}}$ independently for each data vector $Y_i$ 
by solving the following optimization problems:
\begin{align} 
\min_{1 \leq k \leq K}\: \left \| W \Phi_k Y_{i} - H_s (W \Phi_k Y_i) \right \|_{2}^{2}  \;\,\;  \forall \,\,  i \label{clustering}
\end{align} 
where the minimization over $k$ for each $Y_i$ determines the optimal $\Phi_{\hat{k}_i}$, or the cluster $C_{\hat{k}_i}$ to which $Y_i$ belongs. The corresponding optimal sparse code for $Y_i$ in $\mathrm{(P3)}$ is thus $\hat{X}_i = H_s (W \Phi_{\hat{k}_i} Y_i)$. Given the sparse code \footnote{The stored sparse code includes the value of $\hat{X}_i$, as well as the membership index $\hat{k}_i$ which adds just $\text{log}_{2}K$ bits to the code storage.}, one can also easily recover a least squares estimate of each signal as $\hat{Y_i} = \Phi_{\hat{k}_i}^{T}W^{-1}\hat{X}_i$. Since the $\Phi_k$'s are permutation matrices, applying and computing $\Phi_{k}^{T}$ (which is also a permutation matrix) is cheap.

\textbf{Transform Update Step}. Here, we solve for $W$ in $\mathrm{(P2)}$ with fixed $\left\{C_{k}\right\}$ and $\left\{X_{i}\right\}$, which leads to the following problem:
\begin{align} 
 \nonumber (\mathrm{P4}) \;\; & \min_{W}\: \left \| W \tilde{Y} - X \right \|_{F}^{2} + \lambda Q(W)  
\end{align} 
where $\tilde{Y} = \begin{bmatrix}\Phi_{\hat{k}_1}Y_{1}\mid \Phi_{\hat{k}_2}Y_{2}\mid &...& \mid \Phi_{\hat{k}_N}Y_{N}\end{bmatrix}$ contains signals after applying their optimal (as determined in the preceding sparse coding and clustering step) FR operations, and the columns of $X$ are the corresponding sparse codes $X_i$'s. Problem $\mathrm{(P4)}$ has a simple solution involving a singular value decomposition (SVD), which is similar to the transform update step in SST \cite{sabres3}. We first decompose the positive-definite matrix $\tilde{Y}\tilde{Y}^{T} + \lambda I_n = UU^{T}$ (e.g., using Cholesky decomposition). 
Then, denoting the full singular value decomposition (SVD) of the matrix $U^{-1}\tilde{Y} X^{T} = S \Sigma V^{T}$, where $S,\Sigma,V \in \mathbb{R}^{n \times n}$, an optimal transform $\hat{W}$ in $\mathrm{(P4)}$ is
\begin{align} \label{closedform}  
 \;\; \;\; \hat{W} = 0.5 V\left ( \Sigma + (\Sigma^{2}+2\lambda I_{n})^{\frac{1}{2}} \right )S^{T}U^{-1}
\end{align} 
where $(\cdot)^{\frac{1}{2}}$ above denotes the positive definite square root, and $I_n$ is the $n \times n$ identity.

\textbf{Initialization Insensitivity and Cluster Elimination}. Unlike the previously proposed OCTOBOS learning algorithm \cite{wensabres}, which requires initialization of the clusters using heuristic methods such as K-means, the FRIST learning algorithm only needs initialization of the parent transform $W$.  In Section \ref{convergeSupp}, numerical results demonstrate the fast convergence of the proposed FRIST learning algorithm, which can be insensitive to the parent transform initialization. In practice, we apply a heuristic cluster elimination strategy in the FRIST learning algorithm, to select the desired $K$ FR operators. In the first iteration, all possible FR operators $\Phi_{k}$'s \cite{le2005bandelet, zhan2015fast} (i.e., all possible child transforms $W_{k}$'s) are considered for sparse coding and clustering. After each iteration, the learning algorithm eliminates half of the operators with smallest cluster sizes, until the number of selected FR operators drops to K, which only takes a few iterations. For the rest of the iterations, the algorithm only considers the selected K $\Phi_{k}$'s in the sparse coding and clustering steps.

\textbf{Computational Cost Analysis}. The \emph{sparse coding and clustering} step computes the optimal sparse codes and clusters, with $O(Kn^{2}N)$ cost.  
In the transform update step, we compute the optimal solution for the square parent transform in $\mathrm{(P4)}$. The cost of computing this solution scales as $O(n^{2}N)$ (dominated by matrix-matrix products), assuming $N \gg n$, which is cheaper than the sparse coding and clustering step. Thus, the overall computational cost per iteration of FRIST learning using the proposed alternating algorithm scales as $O(Kn^{2}N)$, which is typically lower than the cost per iteration of the overcomplete K-SVD learning algorithm \cite{elad}, with the number of dictionary atoms $m = Kn$, and synthesis sparsity $s \propto n$. 
FRIST learning also converges quickly in practice as illustrated later in Section \ref{convergeSupp}.
The computational costs per-iteration of SST, OCTOBOS, FRIST, and KSVD learning are summarized in Table \ref{comptable}.

\begin{table}[t]
\caption{Computational cost (per iteration) comparison between SST ($W \in \mathbb{R}^{n \times n}$), OCTOBOS ($K$ clusters, each $W_k \in \mathbb{R}^{n \times n}$), FRIST and KSVD ($D \in \mathbb{R}^{n \times m}$) learning. $N$ is the number of training signals.}
\label{comptable}
\begin{center}
\fontsize{12}{18pt}\selectfont
\begin{tabular}{|c|c|c|c|c|}
\hline
 & SST. & OCTOBOS & FRIST & KSVD  \\
\hline
Cost  & $O(n^{2}N)$ & $O(Kn^{2}N)$ & $O(Kn^{2}N)$ & $O(mn^{2}N)$ \\
\hline
\end{tabular}
\end{center}
\end{table}

\subsection{Convergence Analysis} \label{convthm}
\newtheorem{proposition}{Proposition}
\newtheorem{lemma}{Lemma}
\newtheorem{theorem}{Theorem}
\newtheorem{corollary}{Corollary}
\newtheorem{conjecture}{Conjecture}

We analyze the convergence behavior of the proposed FRIST learning algorithm for $(\mathrm{P2})$, assuming that every step in the algorithms (such as SVD) is computed exactly. 

\textbf{Notation}. Problem $(\mathrm{P2})$ is formulated with sparsity constraints (for each $X_{i}$), which is equivalent to an unconstrained formulation with sparsity barrier penalties $\phi (X_i)$ (which equals $+ \infty$ when the constraint is violated, and is zero otherwise). Thus, the objective function of Problem $(\mathrm{P2})$ can be rewritten as
\begin{equation}
f\left ( W, X, \Lambda \right ) = \sum _{k=1}^K \sum_{i \in C_{k}} \left \{ \left \| W \Phi_{k}Y_{i}-X_{i} \right \|_{2}^{2}  + \phi(X_i) \right \} + \lambda Q(W)
\end{equation}\label{not1}
where $\Lambda \in \mathbb{R}^{1 \times N}$ is the vector whose $\mathrm{i}$th element $\Lambda_{i} \in \left \{ 1, .., K \right \}$ denotes the cluster label corresponding to the signal $Y_{i}$, i.e., $C_{\Lambda_i}$. We use $\left \{ W^{t}, X^{t}, \Lambda^{t} \right \}$ to denote the output in each iteration (consisting of the sparse coding and clustering, and transform update steps) $t$, generated by the proposed FRIST learning algorithm.

\textbf{Main Results}. Since FRIST can be interpreted as a structured OCTOBOS, the convergence results for the FRIST learning algorithm take a form similar to those obtained for the OCTOBOS learning algorithm \cite{wensabres} in our recent work. The convergence result for the FRIST learning algorithm for $(\mathrm{P2})$, is summarized in the following theorem and corollaries.

\begin{theorem}\label{theorem1}
For each initialization $(W^{0}, X^{0}, \Lambda^{0})$, the following results hold:
\begin{enumerate}[(i)]
\item The objective sequence $\left \{f^t = f(W^t, X^t, \Lambda^t) \right \}$ in the FRIST learning algorithm is monotone decreasing, and converges to a finite value, $f^{*} = f^{*}(W^{0}, X^{0}, \Lambda^{0})$. 
\item The iterate sequence $\left \{ W^{t}, X^{t}, \Lambda^{t} \right \}$ is bounded, with all of its accumulation points equivalent, i.e., achieving the exact same value $f^{*}$.
\item Every accumulation point $( W, X, \Lambda )$ of the iterate sequence satisfies the following partial global optimality conditions:
\begin{equation} \label{drto1bbaa}
(X, \Lambda) \in \underset{\tilde{X}, \tilde{\Lambda}}{\arg\min} \; \,  f\left ( W, \tilde{X},\tilde{\Lambda} \right )
\end{equation} 
\begin{equation} \label{tropi3fazzbaa2}
W \in \underset{\tilde{W}}{\arg\min} \; \,  f\left ( \tilde{W}, X, \Lambda \right )
\end{equation}
\item For each accumulation point $( W, X, \Lambda )$, there exists $\epsilon = \epsilon (W) > 0$ such that
\begin{equation} \label{dde22}
f\left (W + dW, X + \Delta X, \Lambda \right ) \geq f\left ( W, X, \Lambda \right ) = f^{*},
\end{equation}
which holds for all $dW \in \mathbb{R}^{n \times n}$  satisfying $\left \| dW \right \|_{F} \leq \epsilon$, and all $\Delta X \in \mathbb{R}^{n \times N}$ satisfying $\left \| \Delta X \right \|_{\infty} < \min_{k} \min_{i \in C_{k}}\left \{ \psi _{s}(W \Phi_k Y_{i}) : \left \| W \Phi_k Y_{i} \right \|_{0}>s \right \}$. Here, we define the infinity norm of matrix $\Delta X$ as $\left \| \Delta X \right \|_{\infty} \triangleq \max_{i, j} |\Delta X_{i,j}|$, and the operator $\psi_{s} ( \cdot )$ returns the $\mathrm{s}$th largest magnitude in a vector.
\end{enumerate}
\end{theorem}
\vspace{0.05in}

Conclusion $(iv)$ provides a partial local optimality condition for each accumulation point with respect to $(W, X)$, where the local perturbation $dW$ in Equation (\ref{dde22}) is sufficiently small, and $\Delta X$ is specified by a finite region, which is determined by a scalar $\kappa$ that limits the amplitudes of entries in $\Delta X$ (i.e., $\left \| \Delta X \right \|_{\infty} < \kappa$). Here $\kappa = \min_{k} \kappa_k$, and each $\kappa_k =  \min_{i \in C_{k}}\left \{ \psi _{s}(W \Phi_k Y_{i}) : \left \| W \Phi_k Y_{i} \right \|_{0}>s \right \}$ is computed by $(i)$ choosing the vectors with sparsity $>s$ from $\left \{ W \Phi_k Y_{i} \right \}$ where $i \in C_{k}$, $(ii)$ selecting the $\mathrm{s}$th largest magnitude in each of those vectors, and $(iii)$ returning the smallest of those values. If there are no vectors with sparsity of $W\Phi_{k}Y_{i}$ greater than $s$ in a class $k$, then that $\kappa_{k}$ is set as $\infty$.

\begin{corollary}\label{corollary1}
For a particular initial $(W^{0}, X^{0}, \Lambda^{0})$, the iterate sequence in the FRIST learning algorithm converges to an equivalence class of accumulation points, which are also partial minimizers satisfying \eqref{drto1bbaa}, \eqref{tropi3fazzbaa2}, and \eqref{dde22}. 
\end{corollary}
\vspace{0.04in}

\begin{corollary}\label{corollary2}
The iterate sequence $\left \{ W^{t}, X^{t}, \Lambda^{t} \right \}$ in the FRIST learning algorithm is globally convergent (i.e., it converges from any initialization) to the set of partial minimizers of the objective $f\left (W, X, \Lambda  \right )$.
\end{corollary}
\vspace{0.04in}

For reasons of space, 
we only provide an outline of proofs. The conclusion $(i)$ in Theorem \ref{theorem1} is obvious, as the proposed alternating algorithm solves the sub-problem in each step exactly. The proof of Conclusion $(ii)$ follows the same arguments as in the proofs in Lemma $3$ and Lemma $5$ in \cite{wensabres}. In Conclusion $(iii)$, Condition (\ref{drto1bbaa}) can be proved using the arguments for Lemma $7$ from \cite{wensabres}, while Condition (\ref{tropi3fazzbaa2}) can be proved with the arguments for Lemma $6$ from \cite{sabres3}. The last conclusion in Theorem \ref{theorem1} can be shown using similar arguments as from the proof of Lemma $9$ in \cite{sabres3}.

Theorem \ref{theorem1} and Corollaries \ref{corollary1} and \ref{corollary2} establish that with any initialization $(W^{0}, X^{0}, \Lambda^{0})$, the iterate sequence $\left \{ W^{t}, X^{t}, \Lambda^{t} \right \}$ generated by the FRIST learning algorithm converges to an equivalence class (corresponding to a common objective value $f^{*}$ -- that may depend on initialization) of partial minimizers of the objective. No assumptions are made about the initialization to establish these results. We leave the investigation of stronger convergence results (e.g., convergence to global minima) with additional assumptions including on the algorithm initialization or using a probabilistic analysis framework, to future work.

\section{Applications} \label{app}

Natural, or biomedical images typically contain a variety of directional features and edges. Thus the FRIST model is particularly appealing for applications in image processing and inverse problems. In this section, we consider three such applications, namely image denoising, image inpainting, and blind compressed sensing (BCS)-based magnetic resonance imaging (MRI).

\subsection{Image Denoising} \label{denoisingForm}
Image denoising is one of the most fundamental inverse problems in image processing. The goal is to reconstruct a 2D image represented as a vector $y \in \mathbb{R}^{P}$, from its measurement $z = y + h$, corrupted by a noise vector $h$. Various denoising algorithms have been proposed recently, with state-of-the-art performance \cite{DCTart, dbov}. Similar to previous dictionary and transform learning based image denoising methods \cite{elad2, wensabres}, we propose the following patch-based image denoising formulation using FRIST learning:
\begin{align}
 \nonumber (\mathrm{P5})\;\;\; \min_{W, \left \{y_{i}, x_{i}, C_{k} \right \}} \sum _{k=1}^K \sum_{i \in C_{k}} &  \left \{ \left \| W \Phi_k y_{i}-x_{i} \right \|_{2}^{2} + \tau \left \| R_{i}\:z - y_{i}\right \|_{2}^{2} \right \} + \lambda  \,  Q(W) \\
\nonumber &  \;\;\;\;\;\;\; s.t.\; \:  \left \| x_{i} \right \|_{0}\leq s_{i}\; \: \forall \,\,  i, \,\,\, \left\{C_{k}\right\} \in \Gamma \;\;\; 
\end{align}  
where $R_{i} \in \mathbb{R}^{n \times P}$ denotes the patch extraction operator, i.e., $R_{i}z \in \mathbb{R}^{n }$ represents the $i$th overlapping patch of the corrupted image $z$ as a vector. We assume $N$ overlapping patches in total. The data fidelity term $\tau \left \| R_{i}\:z - y_{i}\right \|_{2}^{2}$ measures the discrepancy between the observed patch $R_i z$ and the (unknown) noiseless patch $y_i$, and uses a weight $\tau = \tau_{0}/\sigma$ that is inversely proportional to the given noise standard deviation $\sigma$ \cite{elad2, sabres3}, and $\tau_{0}>0$. The vector $x_{i} \in \mathbb{R}^{n}$ represents the sparse code of $y_{i}$ in the FRIST domain, with an a priori unknown sparsity level $s_{i}$. We follow the previous SST-based and OCTOBOS-based denoising methods \cite{doubsp2l, wensabres}, and impose a sparsity constraint on each $y_{i}$.

We propose a simple iterative denoising algorithm based on $(\mathrm{P5})$. Each iteration involves the following steps: $(i)$ sparse coding and clustering, $(ii)$ sparsity level update, and $(iii)$ transform update. Once the iterations complete, we have a denoised image reconstruction step. 
We initialize the $\left \{ y_i \right \}$ in $(\mathrm{P5})$ using the noisy image patches $\left \{ R_i z \right \}$. Step $(i)$ is the same as described in Section \ref{algorithm}. We then update the sparsity levels $s_{i}$ for all $i$, similar to the SST learning-based denoising algorithm \cite{doubsp2l}. With fixed $W$ and clusters $\left\{ C_k \right\}$, we observe the solution for $y_{i}$ ($i \in C_k$) in $(\mathrm{P5})$ in the least squares sense,
\begin{equation} \label{spup}
y_{i}= \Phi_{k}^{T}  \begin{bmatrix}
\sqrt{\tau} \, I \\ W

\end{bmatrix}^{\dagger}\begin{bmatrix}
\sqrt{\tau}\:v_i\\ H_{s_i}(W v_i)

\end{bmatrix}=G_{1}v_i+G_{2} H_{s_i}(W v_i)
\end{equation}
where $G_1$ and $G_2$ are appropriate matrices in the above decomposition, and $v_i \triangleq \Phi_{k} R_{i}\: z$ are the rotated/flipped noisy patches, which can be pre-computed in each iteration. We choose the optimal $s_i$ to be the smallest integer that makes the reconstructed $y_i$ in $(\ref{spup})$ satisfy the error condition
$\left \| R_i z - y_{i}\right \|_{2}^{2} \leq nC^{2}\sigma^{2}$, 
where $C$ is a constant parameter \cite{doubsp2l}. Once the sparsity level update (step $(ii)$) is completed, we proceed to the transform update based on the method in Section \ref{algorithm}. 
The algorithm alternates between steps $(i)$-$(iii)$ for a fixed number of iterations, and eventually the denoised image patches $\left\{y_{i}\right\}$ are obtained using (\ref{spup}).
Each pixel in the reconstructed patch is projected onto the underlying intensity range (image pixel is typically stored as 8-bit integer, which corresponds to the intensity range $[0, 255]$).
The denoised image is reconstructed by averaging the overlapping denoised patches at their respective image locations. 

For improved denoising, the algorithm for (P5) is repeated for several passes by replacing $z$ with the most recent denoised image estimate in each pass. The noise standard deviation $\sigma$ decreases gradually in each such pass, and is found (tuned) empirically \cite{wensabres}.

\subsection{Image Inpainting} \label{inpaintingForm}
The goal of image inpainting is to recover missing pixels in an image.
The given image measurement, with missing pixel intensities set to zero, is denoted as $z = \Xi y + \varepsilon$, where $\varepsilon$ is the additive noise on the available pixels, and $\Xi \in\mathbb{R}^{P \times P}$ is a diagonal binary matrix with zeros at locations corresponding to missing pixels. We propose the following patch-based image inpainting formulation using FRIST learning:
\begin{align}
 \nonumber (\mathrm{P6})\;\;\; \min_{W, \left \{y_{i}, x_{i}, C_{k} \right \}} \sum _{k=1}^K \sum_{i \in C_{k}} \left\{ \left \| W \Phi_k y_{i}-x_{i} \right \|_{2}^{2} + \tau^2 \left \| x_{i} \right \|_{0}   + \gamma \left \| P_i y_{i} - z_{i} \right \|_{2}^{2} \right\} + \lambda  \,  Q(W)  
\end{align} 
where $z_i = R_i z$ and $y_i = R_i y$. The diagonal binary matrix $P_i \in\mathbb{R}^{n \times n}$ captures the available (non-missing) pixels in $z_i$. The sparsity penalty $\tau^2 \left \| x_{i} \right \|_{0}$ is used, which leads to an efficient algorithm. The fidelity term $\gamma \left \| P_i y_{i} - z_{i} \right \|_{2}^{2}$ for the $i$th patch has the coefficient $\gamma$ that is chosen inversely proportional to the noise standard deviation $\sigma$ (in the measured pixels). The parameter $\tau$ is chosen proportional to the number of pixels that are missing in $z$.

Our proposed iterative algorithm for solving $(\mathrm{P6})$ involves the following steps: $(i)$ sparse coding and clustering, and $(ii)$ transform update. Once the iterations complete, we have a $(iii)$ patch reconstruction step. The sparse coding problem with a sparsity penalty has a closed-form solution \cite{sawenbres1}, and thus Step $(i)$ is equivalent to solving the following clustering problem:
\begin{align} \label{penaltyclustering}
\min_{1 \leq k \leq K}\: \left \| W \Phi_k y_{i} - T_{\tau} (W \Phi_k y_i) \right \|_{2}^{2} + \tau^2 \left \| T_\tau (W \Phi_k y_i) \right \|_0  \;\,\;  \forall \,\,  i 
\end{align} 
where the hard thresholding operator $T_{\tau}(\cdot)$ is defined as
\begin{equation} \label{equ88}
 \left ( T_{\tau} (b) \right )_{j}=\left\{\begin{matrix}
 0&, \;\;\left | b_{j} \right | < \tau \\
b_{j}  & ,\;\;\left | b_{j} \right | \geq \tau 
\end{matrix}\right.
\end{equation}
where the vector $b \in \mathbb{R}^{n}$, and the subscript $j$ indexes its entries. The optimal sparse codes are obtained by hard thresholding $y_i$ in the optimal transform domain. Step $(ii)$ is similar to that in the denoising algorithm in Section \ref{denoisingForm}. In the following, we discuss Step $(iii)$ by considering two cases.

\textbf{Ideal image inpainting without noise}. In the ideal case when the noise $\varepsilon$ is absent, i.e., $\sigma = 0$, the coefficient of the fidelity term $\gamma \rightarrow \infty$. Thus the fidelity term can be replaced with hard constraints $P_i\:y_i = z_i\; \forall \, i$. In the noiseless reconstruction step, with fixed $\left \{x_{i}, C_{k} \right \}$ and $W$, we first reconstruct each image patch $y_{i}$ by solving the following linearly constrained least squares problem:
\begin{equation}
\min_{y_{i}} \left \| W \Phi_{k_i} y_{i}-x_{i} \right \|_{2}^{2}\;\; s.t.\; \: P_i\:y_i = z_i\;.
\end{equation}
We define $z_i = P_i y_i \triangleq y_i - e_i$, where $e_i =(I_{n} - P_i) y_i$. 
$\Omega_i = \mathrm{supp}(\Phi_{k_i} \mathrm{diag}(I_n - P_i))$, which is complementary to $\mathrm{supp}(\Phi_{k_i} \mathrm{diag}(P_i))$, where ``diag" denotes the diagonal of a matrix extracted as a vector.
Since the constraint leads to the relationship $y_i = z_i + e_i$ with $z_i$ given, we solve the equivalent minimization problem over $e_i$ as follows:
\begin{equation} \label{inpaintingRecon}
\min_{e_{i}} \left \| W \Phi_{k_i} e_{i}- (x_{i} - W\Phi_{k_i}\:z_i) \right \|_{2}^{2}\;\;s.t.\; \: \text{supp}(\Phi_{k_i} e_i) = \Omega_i \;.
\end{equation}
Here, we define $W_{\Omega_i}$ to be the submatrix of $W$ formed by columns indexed in $\Omega_i$, and $(\Phi_{k_i} e_{i})_{\Omega_i}$ to be the vector containing the entries at locations $\Omega_i$ of $\Phi_{k_i} e_{i}$. Thus, $W \Phi_k e_{i} = W_{\Omega_i} (\Phi_{k_i} e_{i})_{\Omega_i}$, and we define $\xi^{i} \triangleq \Phi_{k_i} e_{i}$. The reconstruction problem is then re-written as the following unconstrained problem:
\begin{equation} 
\min_{\xi_{\Omega_i}^{i}} \left \| W_{\Omega_i} \xi_{\Omega_i}^{i}- (x_{i} - W\Phi_k\:z_i) \right \|_{2}^{2}\;\; \forall \, i \; .
\end{equation}
The above least squares problem has a simple solution given as $\hat{\xi}_{\Omega_i}^{i} = W_{\Omega_i}^{\dagger}(x_{i} - W\Phi_k\:z_i)$. Accordingly, we can calculate $\hat{e}_i = \Phi_{k_i}^{T}\hat{\xi}^{i}$, and thus the reconstructed patches $\hat{y}_i = \hat{e}_i + z_i$.

\textbf{Robust image inpainting}. We now consider the case of noisy $z$, and propose a robust inpainting algorithm (i.e., for the aforementioned Step $(iii)$). This is useful because real image measurements are inevitably corrupted with noise \cite{elad3}. The robust reconstruction step for each patch is to solve the following problem:
\begin{equation} \label{robustRecon}
\min_{y_{i}} \left \| W \Phi_{k_i} y_{i}-x_{i} \right \|_{2}^{2}  + \gamma \left \| P_i y_{i} - z_{i} \right \|_{2}^{2}
\end{equation}
Let $\tilde{z}_i \triangleq \Phi_{k_i} z_i$, $u_i \triangleq \Phi_{k_i} y_i$, and $\tilde{P}_i \triangleq \Phi_{k_i} P_i \Phi_{k_i}^{T}$, where $\Phi_{k_i}$ is a permutation matrix. The rotated solution $\hat{u}_i$ in optimization problem (\ref{robustRecon}) is equivalent to
\begin{equation} \label{rotateRecon}
\hat{u}_i = \underset{u_i}{\arg\min} \left \| W u_i-x_{i} \right \|_{2}^{2}  + \gamma \left \| \tilde{P}_i u_i - \tilde{z}_i \right \|_{2}^{2}
\end{equation}
which has a least squares solution $\hat{u}_i = (W^{T}W + \gamma\tilde{P}_i)^{-1}(W^{T}x_i + \gamma\tilde{P}_i\tilde{z}_i)$.  As the matrix inversion $(W^{T}W + \gamma\tilde{P}_i)^{-1}$ is expensive with a cost of $O(n^3)$ for each patch reconstruction, we apply the Woodbury Matrix Identity and rewrite the solution to (\ref{rotateRecon}) as
\begin{equation} \label{rotateRecon2}
 \hat{u}_i =  [B - B_{\Upsilon_i}^{T}(\frac{1}{\gamma} I_{q^{i}} + \Psi_i)^{-1} B_{\Upsilon_i}](W^{T}x_i + \gamma\tilde{P}_i\tilde{z}_i)  
\end{equation}
where $B \triangleq (W^T W)^{-1}$ can be pre-computed, and the support of $\tilde{z}_i$ is denoted as $\Upsilon_i \triangleq \text{supp}(\mathrm{diag}(\tilde{P_i})
)$. The scalar $q^{i} = |\Upsilon_i | $ counts the number of measured pixels in $z_i$. Here, $B_{\Upsilon_i}$ is the submatrix of $B$ formed by $\Upsilon_i$-indexed rows, while $\Psi_i$ is the submatrix of $B_{\Upsilon_i}$ formed by $\Upsilon_i$-indexed columns. Thus, the matrix inversion $(\frac{1}{\gamma} I_{q^{i}} + \Psi_i)^{-1}$ has cost of $O((q^{i})^{3})$ and the other matrix-matrix products in (15) have cost $O(n (q^{i})^{2})$, compared to computing  $(W^{T}W + \gamma\tilde{P}_i)^{-1}$ with cost of $O(n^{3})$ for the reconstruction of each patch. 
For an inpainting problem with most pixels missing ($q^{i} \ll n$), this represents significant savings. On the other hand, with few pixels missing ($q^{i} \approx n$), a similar procedure can be used with $I - P_i$ replacing $P_i$.
Once $\hat{u}_i$ is computed, the patch in (\ref{robustRecon}) is recovered as $\hat{y}_{i} = \Phi_{k_i}^{T}\hat{u}_i$.

Similar to Section \ref{denoisingForm}, each pixel in the reconstructed patch is projected onto the (underlying) intensity range (e.g., $[0, 255]$ for image pixel stored using 8-bit integer).
Eventually, we output the inpainted image by averaging the reconstructed patches at their respective image locations. We perform multiple passes in the inpainting algorithm for $(\mathrm{P6})$ for improved inpainting. In each pass, we initialize $\left\{y_{i}\right\}$ using patches extracted from the most recent inpainted image. By doing so, we  indirectly reinforce the dependency between overlapping patches in each pass.

\subsection{BCS-based MRI} \label{MRIform}

Compressed Sensing (CS) exploits sparsity and enables accurate MRI reconstruction from limited k-space or Fourier measurements \cite{CSMRI, syber, sparseMRI}. However, CS-based MRI may suffer from artifacts at high undersampling factors, when using non-adaptive or analytical sparsifying transforms \cite{dlmri}. Recent works \cite{syber} proposed Blind Compressed Sensing (BCS)-based MR image reconstruction methods using learned signal models, and achieved superior reconstruction results.
The terminology blind compressed sensing (or image model-blind compressed sensing) is used because the dictionary or sparsifying transform for the underlying image patches is assumed unknown a priori, and is learned simultaneously with the image from the undersampled (compressive) measurements themselves.
MR image patches typically contain various oriented features \cite{zhan2015fast}, which have recently been shown to be well sparsifiable by directional wavelets \cite{qu2012undersampled}. As an alternative to approaches involving directional analytical transforms, here, we propose an adaptive FRIST-based approach that can adapt a parent transform $W$ while clustering image patches simultaneously based on their geometric orientations. This leads to more accurate modeling of MR image features.

Similar to the previous TL-MRI work \cite{syber}, we propose a BCS-based MR image reconstruction scheme using the (adaptive) FRIST model, dubbed FRIST-MRI. 
For computational efficiency, we restrict the parent $W$ to be a unitary transform in the following.
The FRIST-blind image recovery problem with a sparsity constraint is formulated as
\begin{align}
 \nonumber \;\;\;\; (\mathrm{P7})\;\;\;\;\; \min_{W, y, \left \{x_{i}, C_{k} \right \}} &  \mu \left \| F_u y - z \right \|_{2}^{2} + \sum _{k=1}^K \sum_{i \in C_{k}} \left \| W \Phi_k R_i y - x_{i} \right \|_{2}^{2} \\
 \nonumber   \;\;\;\;\;\;\; & s.t.\; \: W^{H}W=I,\,\,  \left \| X \right \|_{0}\leq s,\; \, \left \| y \right \|_2 \leq L, \,\,\, \left\{C_{k}\right\} \in \Gamma \; ,
\end{align} 
where $W^{H}W=I$ is the unitary constraint, $y \in \mathbb{C}^{P}$ is the MR image to be reconstructed, and $z \in \mathbb{C}^{M}$ denotes the measurements with the sensing matrix $F_u \in \mathbb{C}^{M \times P}$, which is the undersampled (single coil, Cartesian) Fourier encoding matrix. Here $M \ll P$, so Problem $(\mathrm{P7})$ is aimed to reconstruct an MR image $y$ from highly undersampled measurements $z$. The constraint $\left \| y \right \|_2 \leq L$ with $L > 0$, represents prior knowledge of the signal energy/range. The sparsity term $\left \| X \right \|_{0}$ counts the number of non-zeros in the entire sparse code matrix $X$, whose columns are the sparse codes $\left\{ x_{i} \right\}$. This sparsity constraint enables variable sparsity levels for individual patches \cite{syber}. 


We use a block coordinate descent approach \cite{syber} to solve Problem $(\mathrm{P7})$. The proposed algorithm alternates between $(i)$ sparse coding and clustering, $(ii)$ parent transform update, and $(iii)$ MR image update. We initialize this algorithm, dubbed FRIST-MRI, with an image (estimate for $y$) such as the zero-filled Fourier reconstruction $F_u^{H} z$. Step $(i)$ solves Problem $(\mathrm{P7})$ for $\left\{ x_{i}, C_{k} \right\}$ with fixed $W$ and $y$ as 
\begin{align}\label{MRIspar}
\min_{\left \{x_{i}, C_{k} \right \}} \sum _{k=1}^K \sum_{i \in C_{k}} \left \| W \Phi_k R_i y - x_{i} \right \|_{2}^{2}\;\;\;\;\;s.t.\; \:  \left \| X \right \|_{0}\leq s, \,\,\, \left\{C_{k}\right\} \in \Gamma.
\end{align} 
The exact solution to Problem (\ref{MRIspar}) requires calculating the sparsification error (objective) for each possible clustering. The cost for this scales as $O(P n^2 K^P)$ for $P$ patches \footnote{The number of patches is $P$ when we use a patch overlap stride of $1$ and include patches at image boundaries by allowing them to `wrap-around' on the opposite side of the image [35].}, which is computationally infeasible. 
Instead, we present an approximate solution here, which we observed to work well in our experiments. 
In this method, we first compute the total sparsification error $SE_k$, associated with each $\Phi_k$, by solving the following problem:
\begin{align}\label{approxMRIspar}
SE_k = \sum_{i = 1}^{P} SE_{k}^{i} \triangleq \min_{\left \{\beta_{i}^{k} \right \}} \sum_{i = 1}^{P} \left \| W \Phi_k R_i y - \beta_{i}^{k} \right \|_{2}^{2} \;\;\;\;s.t.\; \: \;\;\; \left \| B^{k} \right \|_{0}\leq s
\end{align} 
where the columns of $B^{k}$ are $\left \{ \beta_{i}^{k} \right \}$. The optimal $B^{k}$ above is obtained by thresholding the matrix with columns
$\begin{Bmatrix}
W\Phi_{k}R_{i}y
\end{Bmatrix}_{i=1}^{P}$
and retaining the $s$ largest magnitude elements. The clusters $\left\{ C_{k} \right\}$ in (\ref{MRIspar}) are approximately computed by assigning $i \in C_{\hat{k}_{i}}$ where $\hat{k}_{i} = \underset{k}{\arg\min}\, SE_k^{i}$. Once the clusters are computed, the corresponding sparse codes $\hat{X}$ in (\ref{MRIspar}) (for fixed clusters) are easily found by thresholding the matrix $\begin{bmatrix}W \Phi_{\hat{k}_1} R_{1} y \mid\, ... \,\mid W \Phi_{\hat{k}_P} R_{P} y\end{bmatrix}$ and retaining the $s$ largest magnitude elements \cite{syber}.

Step $(ii)$ updates the parent transform $W$ with the unitary constraint (and with other variables fixed). The optimal solution, which is similar to previous work \cite{sabres3}, is computed as follows. First, we calculate the full SVD $A X^{H} = \tilde{S}\tilde{\Sigma}\tilde{V}^{H}$, where the columns of $A$ are $\left\{ \Phi_k R_i y \right\}_{i=1}^{P}$. The optimal unitary parent transform is then $\hat{W} = \tilde{V} \tilde{S}^{H}$.

Step $(iii)$ solves for $y$ with fixed $W$ and $\left\{ x_{i}, C_{k} \right\}$ as
\begin{align}\label{MRIrec}
\min_{y } \sum _{k=1}^K \sum_{i \in C_{k}} \left \| W \Phi_k R_i y - x_{i} \right \|_{2}^{2} +  \mu \left \| F_u y - z \right \|_{2}^{2} \;\;\;\;\;s.t.\; \:  \left \| y \right \|_2 \leq L.
\end{align} 
As Problem (\ref{MRIrec}) is a least squares problem with an $\ell_{2}$ constraint, it can be solved exactly using the Lagrange multiplier method \cite{bcs}. Thus (\ref{MRIrec}) is equivalent to
\begin{align}\label{MRIlag}
\min_{y } \sum _{k=1}^K \sum_{i \in C_{k}} \left \| W \Phi_k R_i y - x_{i} \right \|_{2}^{2} +  \mu \left \| F_u y - z \right \|_{2}^{2} + \rho (\left \| y \right \|_2^2 - L)
\end{align} 
where $\rho \geq 0$ is the optimally chosen Lagrange multiplier. An alternative approach to solving Problem (\ref{MRIrec}) is by employing the iterative projected gradient method. However, because of the specific structure of the matrices (e.g., partial Fourier sensing matrix in MRI) involved, (\ref{MRIrec}) can be solved much more easily and efficiently with the Lagrange multiplier method as discussed next.

Similar to the previous TL-MRI work \cite{syber}, the normal equation for Problem (\ref{MRIlag}) (for known multiplier $\rho$) can be simplified as follows, where $F$ denotes the full Fourier encoding matrix assumed normalized ($F^{H}F = I$):
\begin{align}\label{normal}
(F E F^{H} + \mu F F_{u}^{H}F_u F^{H} + \rho I) F y = F \sum _{k=1}^K \sum_{i \in C_{k}}R_{i}^{T} \Phi_{k}^{H}W^{H} x_i + \mu F F_{u}^{H}z
\end{align} 
where $E \triangleq \sum _{k=1}^K \sum_{i \in C_{k}} R_{i}^{T} \Phi_{k}^{H}W^{H} W \Phi_k R_i = \sum _{i=1}^P R_{i}^{T}R_i$. When the patch stride is $1$ and all wrap-around patches are included, $E = n I$, with $I$ the $P \times P$ identity. 
Since $ F E F^{H}$, $\mu F F_{u}^{H}F_u F^{H}$ \cite{syber}, and $\rho I$ are all diagonal matrices, the matrix pre-multiplying $Fy$ in (\ref{normal}) is diagonal and invertible. 
Hence, (\ref{normal}) can be solved cheaply. Importantly, using a unitary constraint for $W$ leads to an efficient update in (\ref{normal}). In particular, the matrix $E$ is not easily diagonalizable when $W$ is not unitary. The problem of finding the optimal Lagrange multiplier reduces to solving a simple scalar equation (see for example, equation (3.17) in \cite{syber}) that can be solved using Newton's method. Thus, the approach based on the Lagrange multiplier method is much simpler compared to an iterative projected gradient scheme to estimate the (typically large) vector-valued image in Problem (\ref{MRIrec}). 

\section{Experiments} \label{exp}

\begin{figure}
\begin{center}
\begin{tabular}{ccccccc}
\includegraphics[width=0.63in]{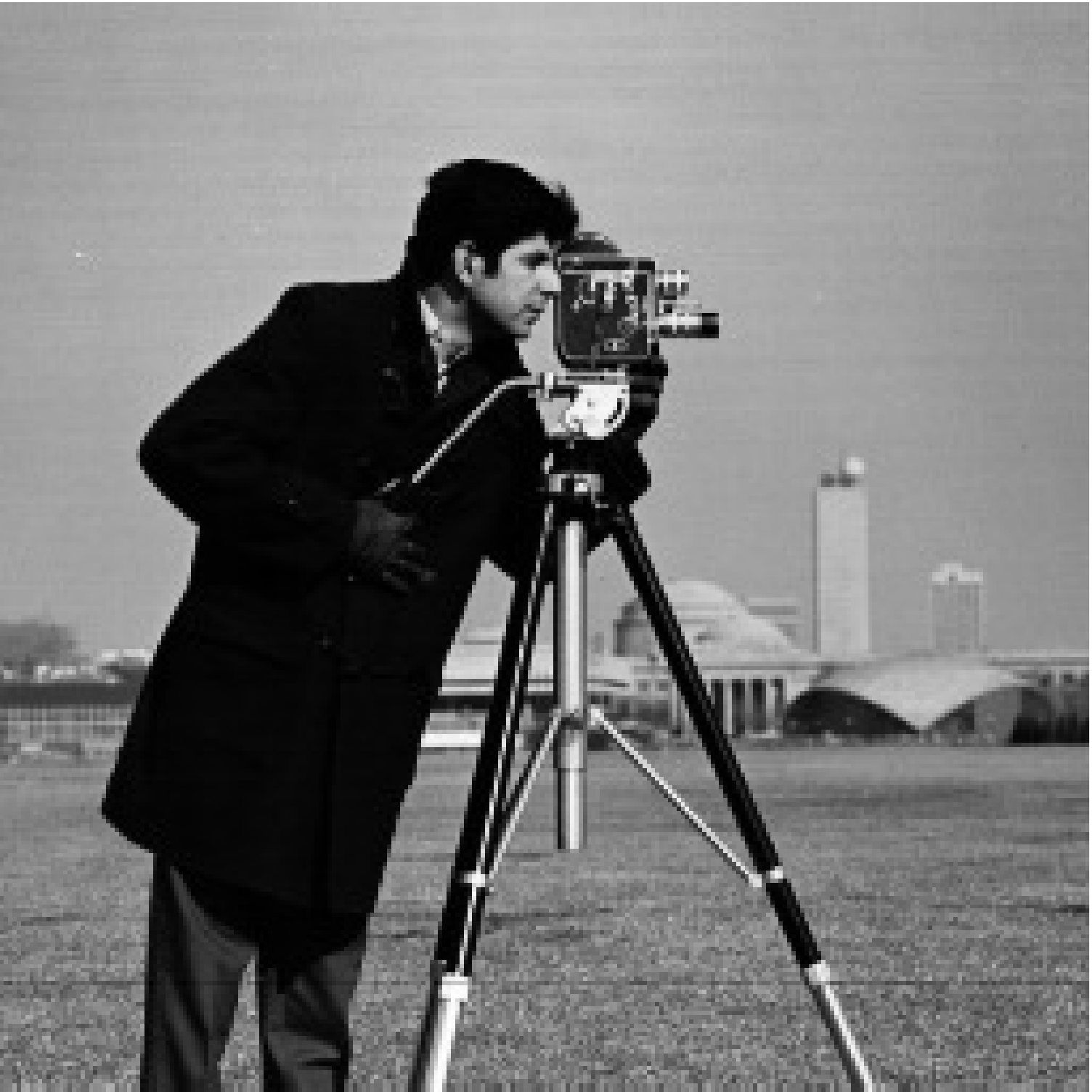}&
\includegraphics[width=0.63in]{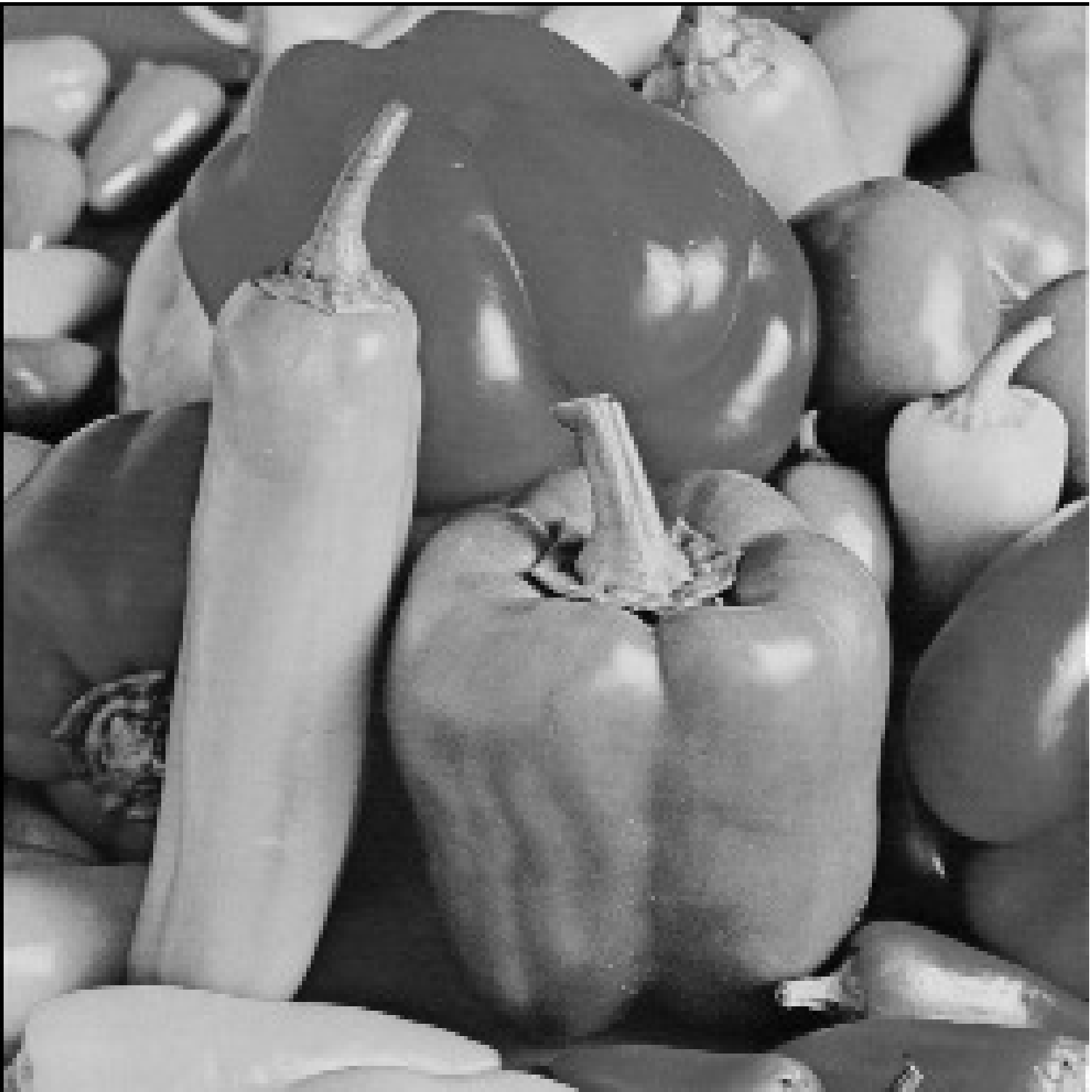}&
\includegraphics[width=0.63in]{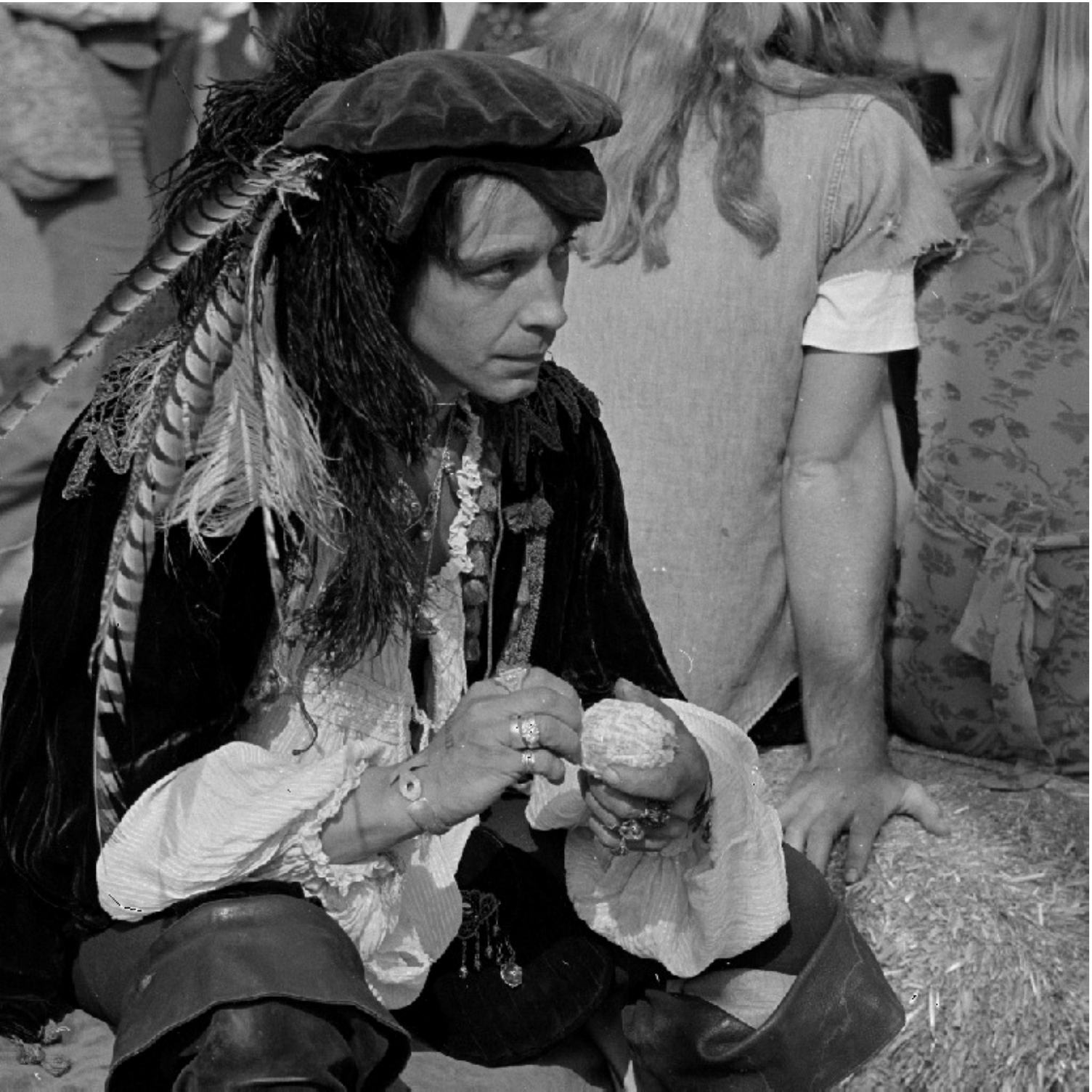}&
\includegraphics[width=0.63in]{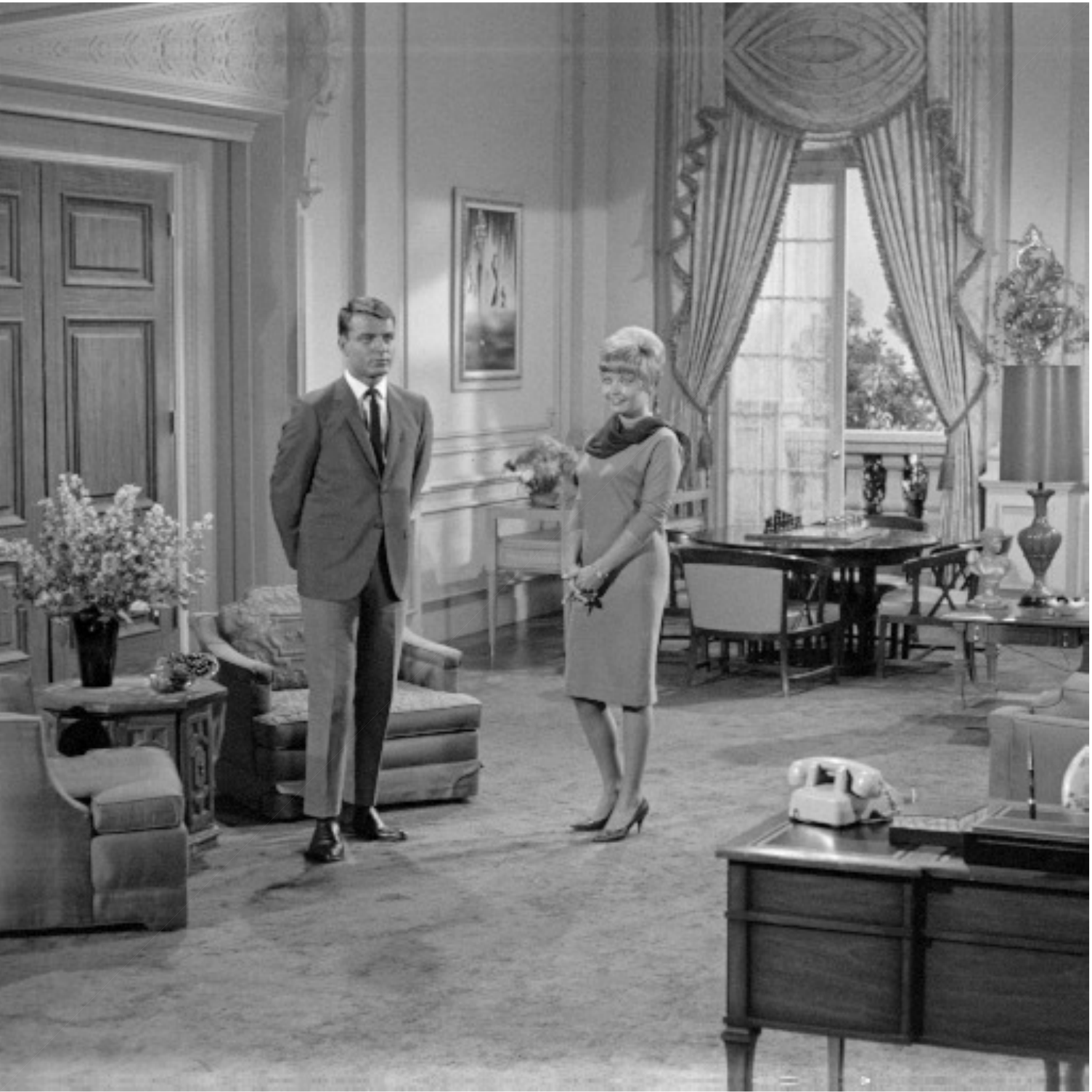}&
\includegraphics[width=0.63in]{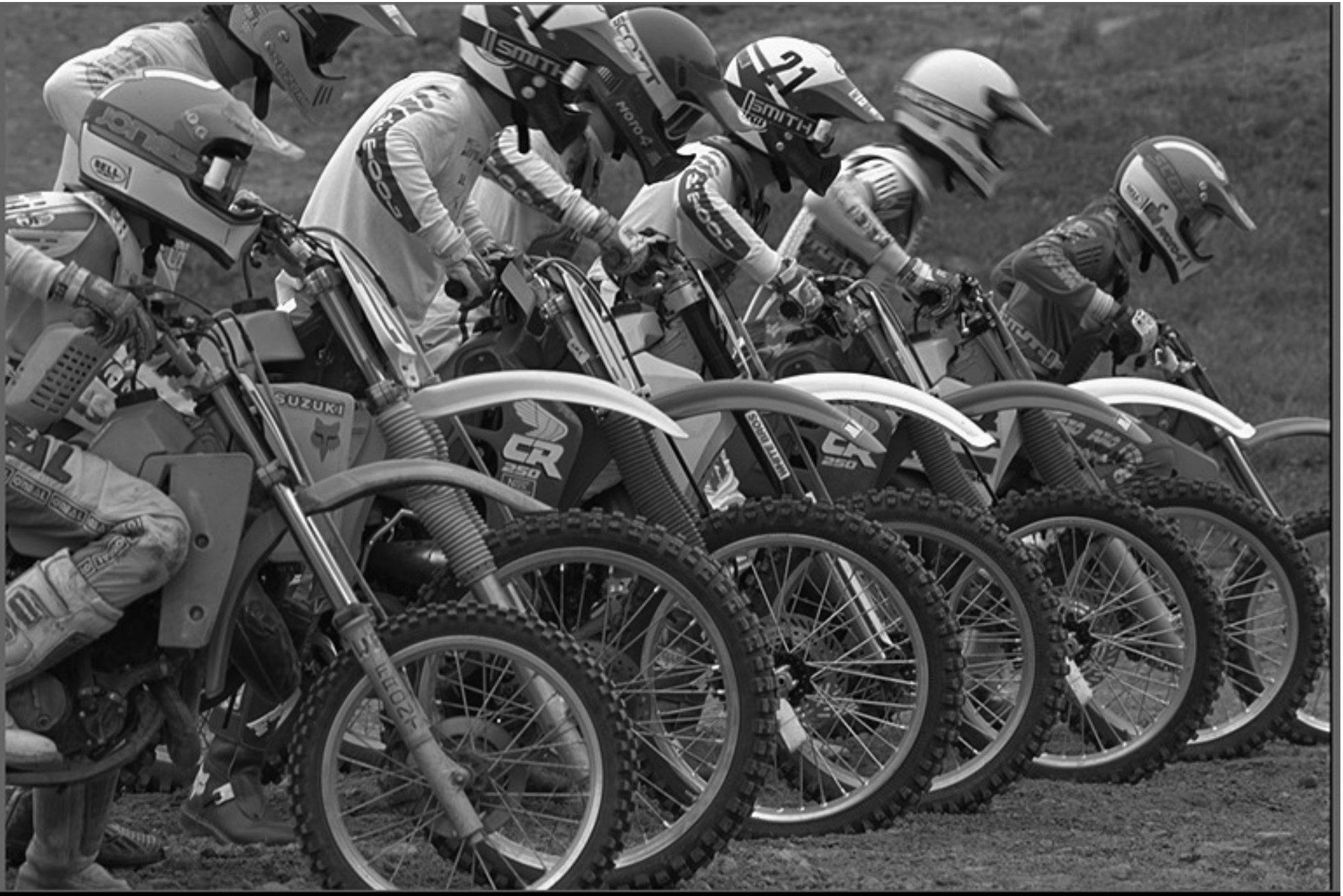}&
\includegraphics[width=0.63in]{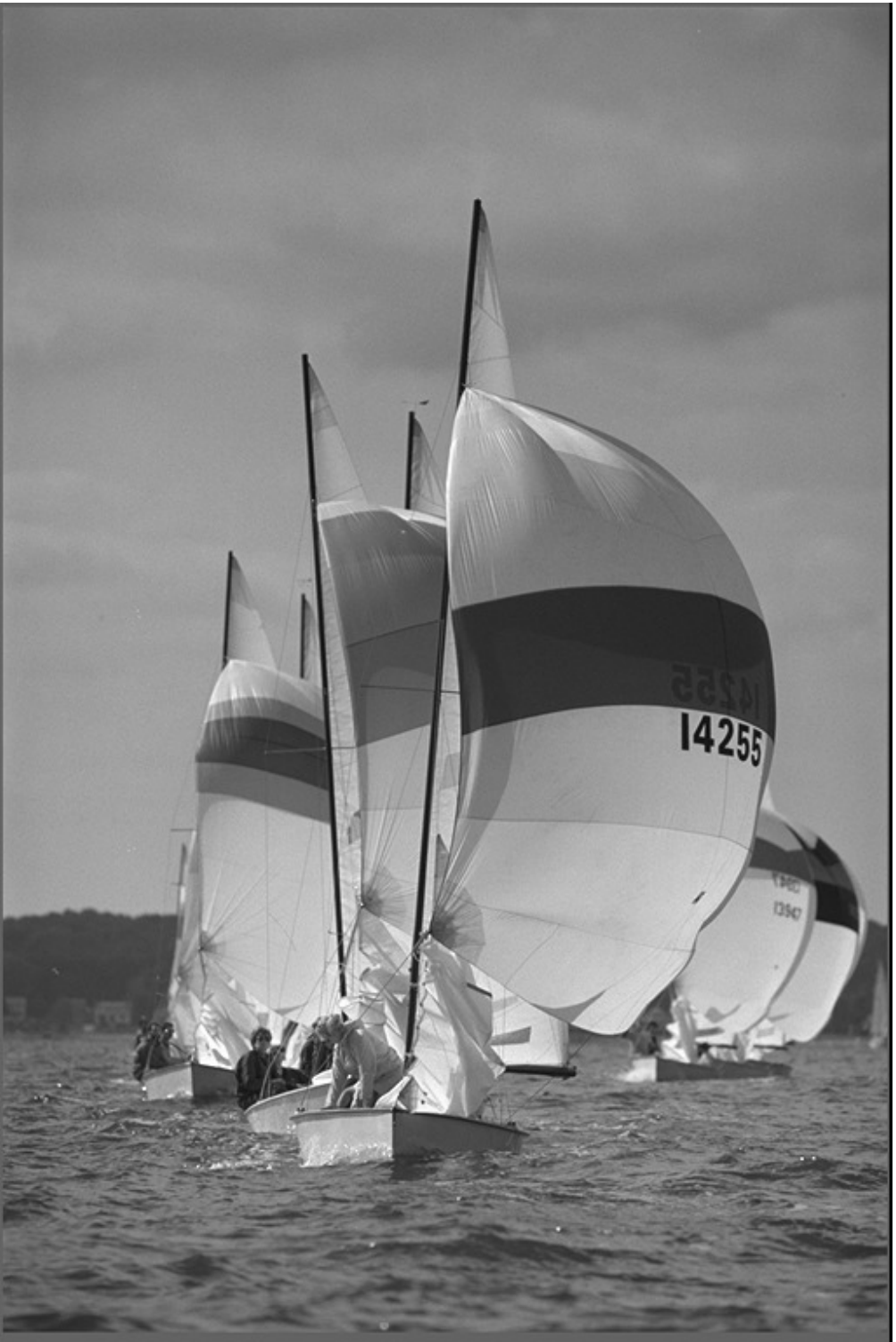}&
\includegraphics[width=0.63in]{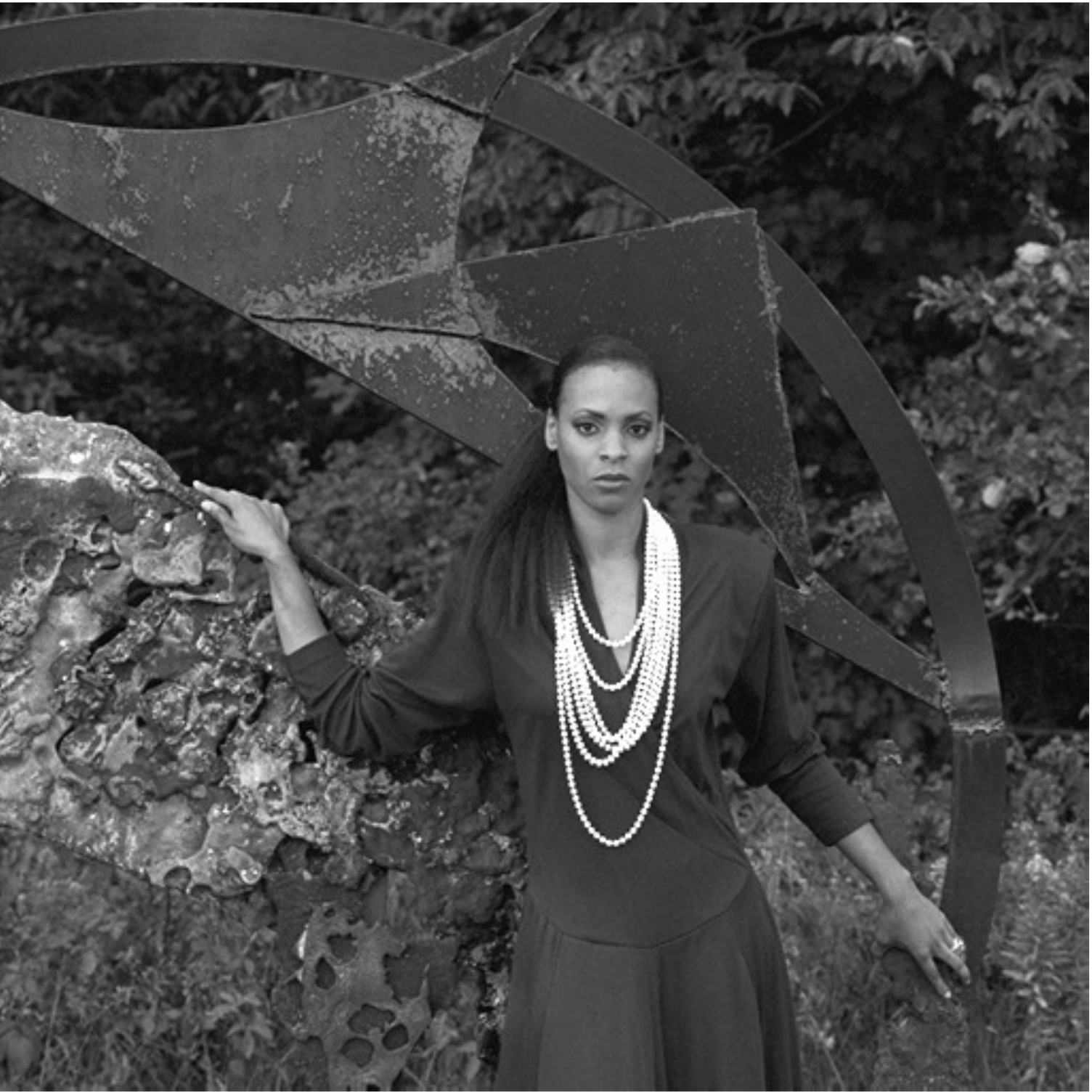}\\
Cameraman & Peppers & Man & Couple & kodak 5 & kodak 9 & kodak 18\\
\end{tabular}
\end{center}
\caption{Testing images used in the image denoising and image inpainting experiments.}
\label{testingIm}
\end{figure}

We present numerical convergence results for the FRIST learning algorithm along with image segmentation examples, as well as some preliminary results demonstrating the promise of FRIST learning in applications including image sparse representation, denoising, robust inpainting, and MRI reconstruction. We work with $8 \times 8$ non-overlapping patches for the study of convergence and sparse representation, $8 \times 8$ overlapping patches for image segmentation, denoising, and robust inpainting, and $6 \times 6$ overlapping patches (including patches at image boundaries that `wrap around' on the opposite side of the image) for the MRI experiments. Figure.\ref{testingIm} lists the testing images that are used in the image denoising and inpainting experiments.

\subsection{Empirical convergence results} \label{convergeSupp}

We first illustrate the convergence behavior of FRIST learning. We randomly extract $10^{4}$ non-overlapping patches from the $44$ images in the USC-SIPI database \cite{sipi} (the color images are converted to gray-scale images), and learn a FRIST model, with a $64 \times 64$ parent transform $W$, from the randomly selected patches using fixed sparsity level $s = 10$ per patch. We set $K = 2$, and $\lambda_0 = 3.1 \times 10^{-3}$ for visualization simplicity. In the experiment, we initialize the learning algorithm with different square $64 \times 64$ parent transforms $W$'s, including the ($i$) Karhunen-Lo\`eve Transform (KLT), ($ii$) 2D DCT, ($iii$) random matrix with i.i.d. Gaussian entries (zero mean and standard deviation $0.2$), and ($iv$) identity matrix.

\begin{figure}
\begin{center}
\begin{tabular}{ccc}
\includegraphics[height=1.35in]{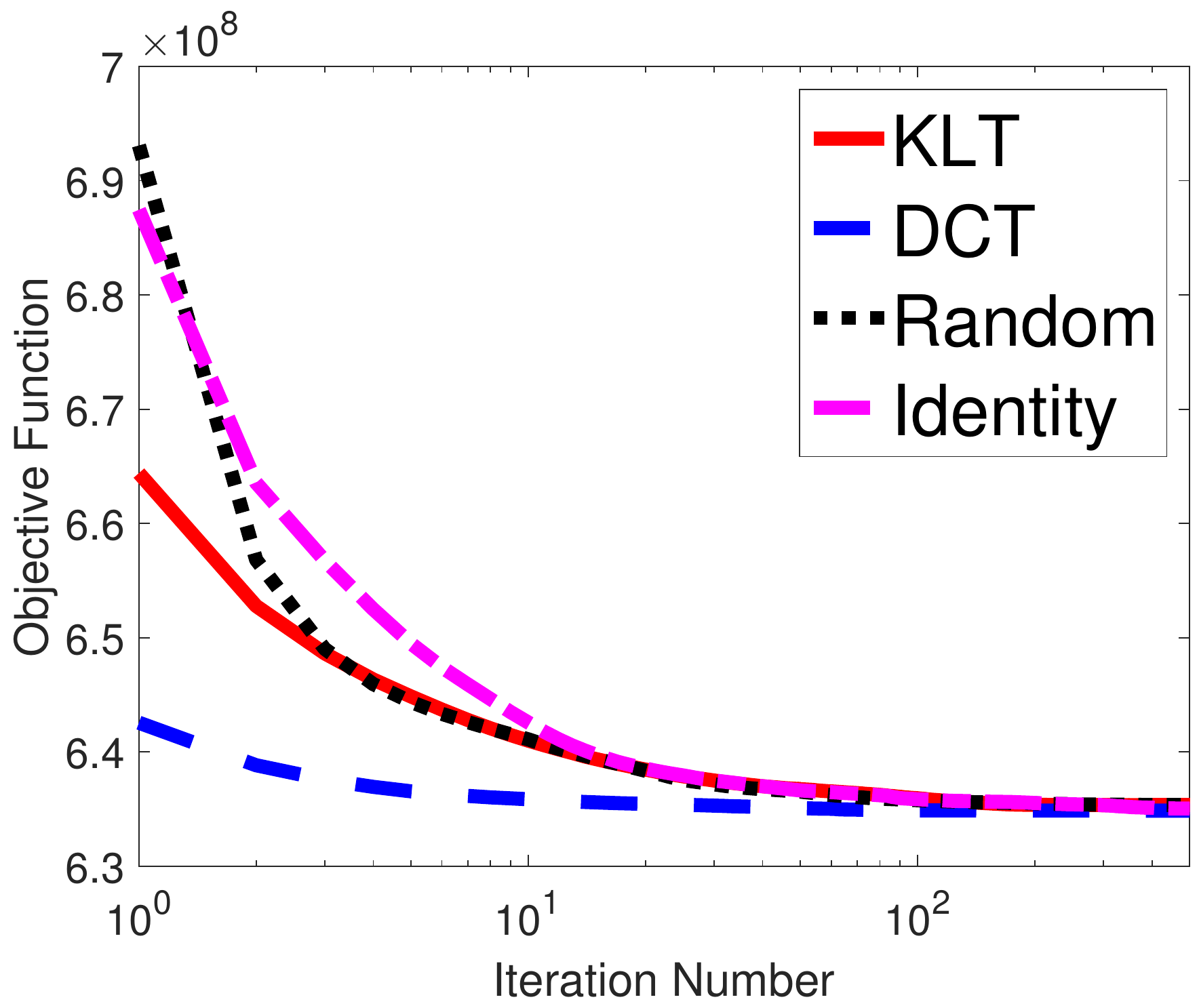}&
\includegraphics[height=1.35in]{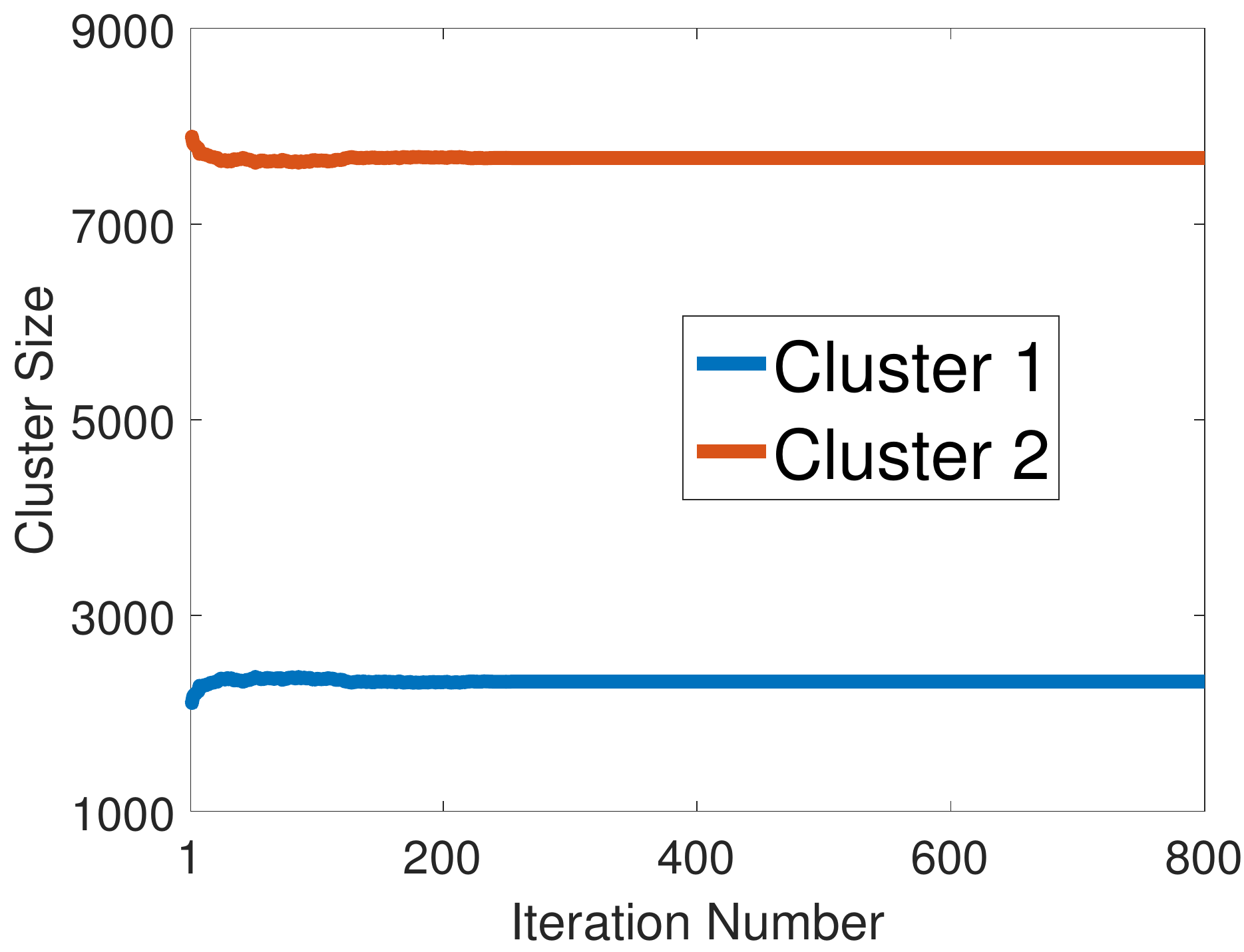}&
\includegraphics[height=1.35in]{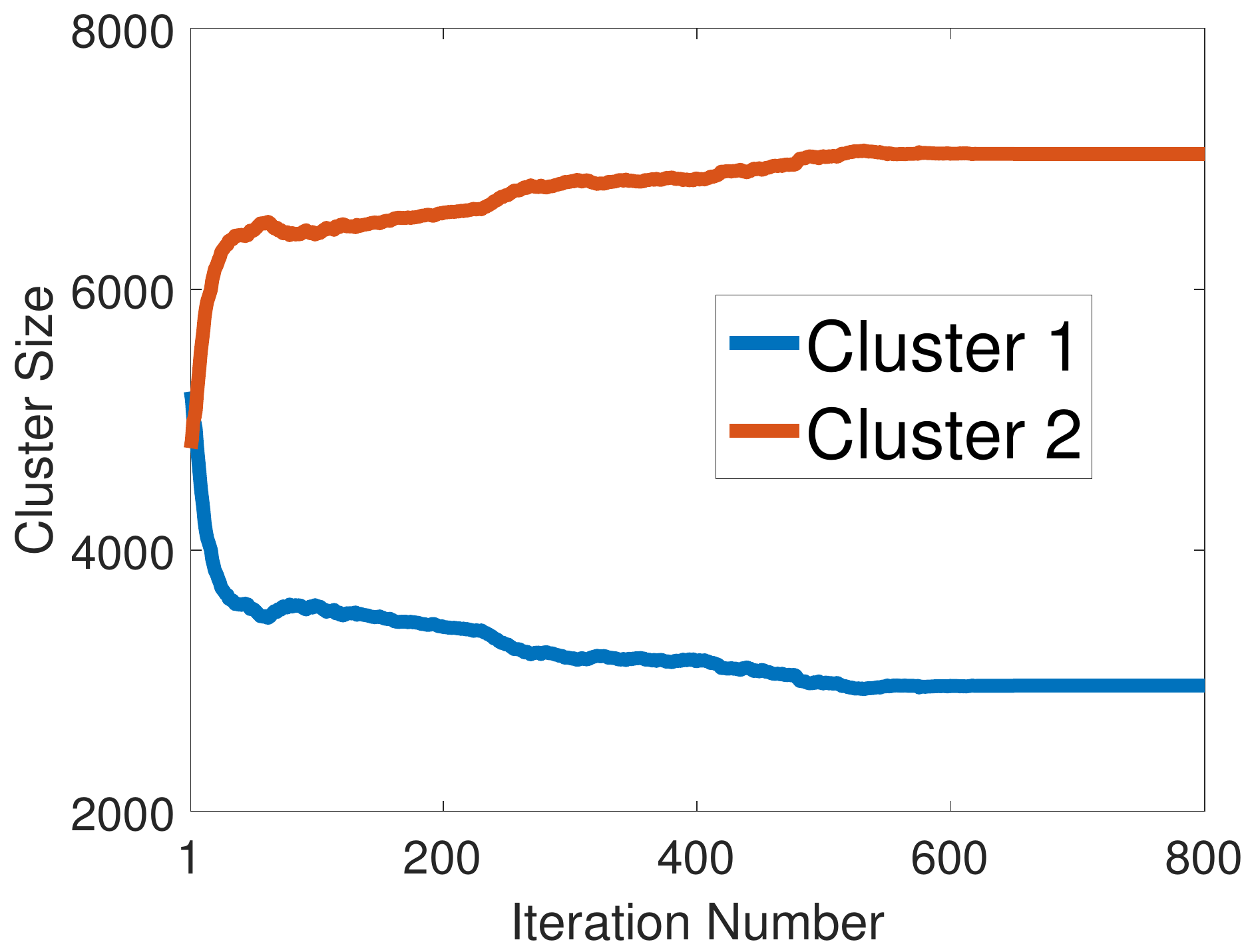}\\
(a) FRIST Objective & (b) DCT initialization & (c) KLT initialization \\
\\
\includegraphics[height=1.35in]{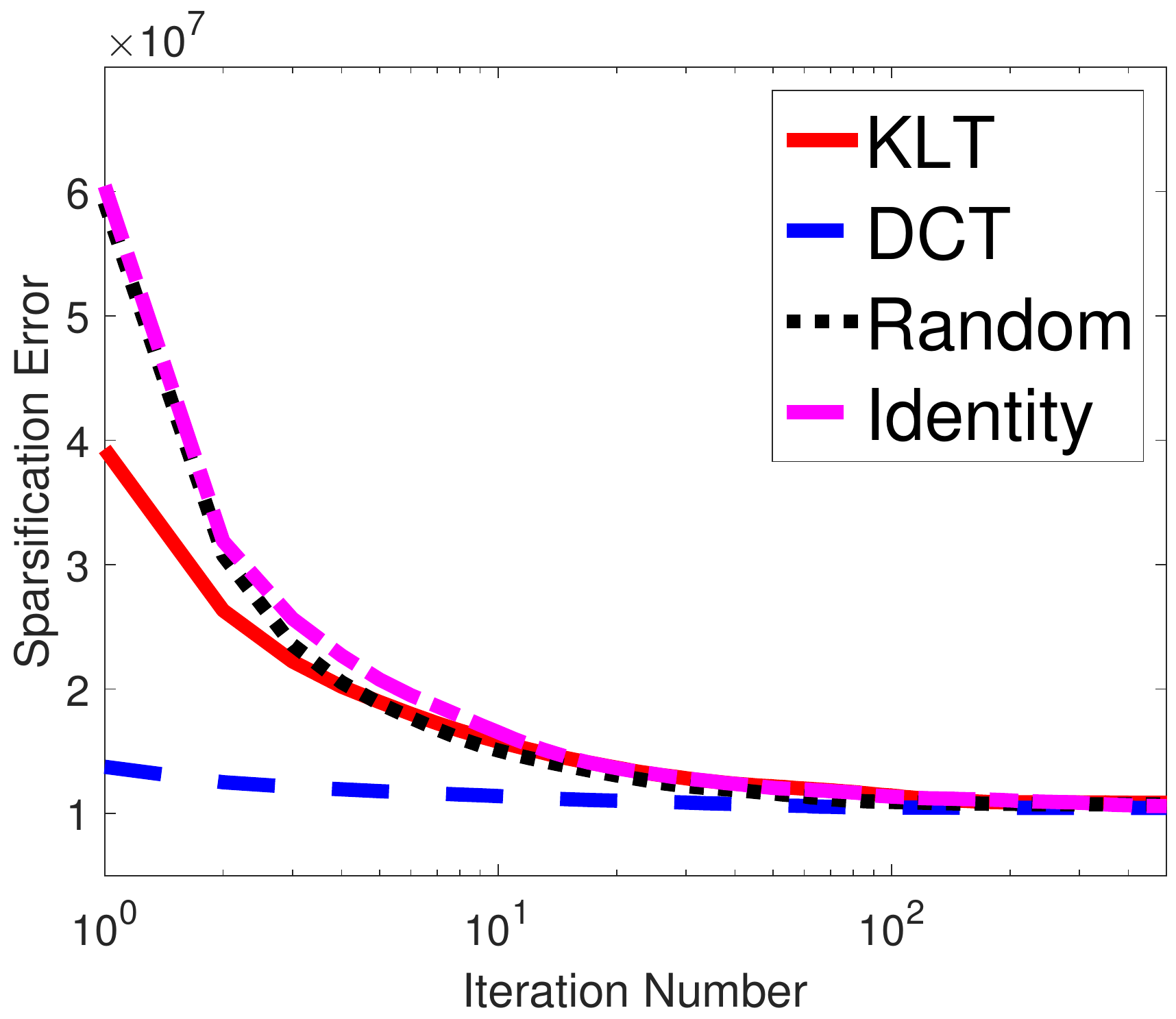}&
\includegraphics[height=1.35in]{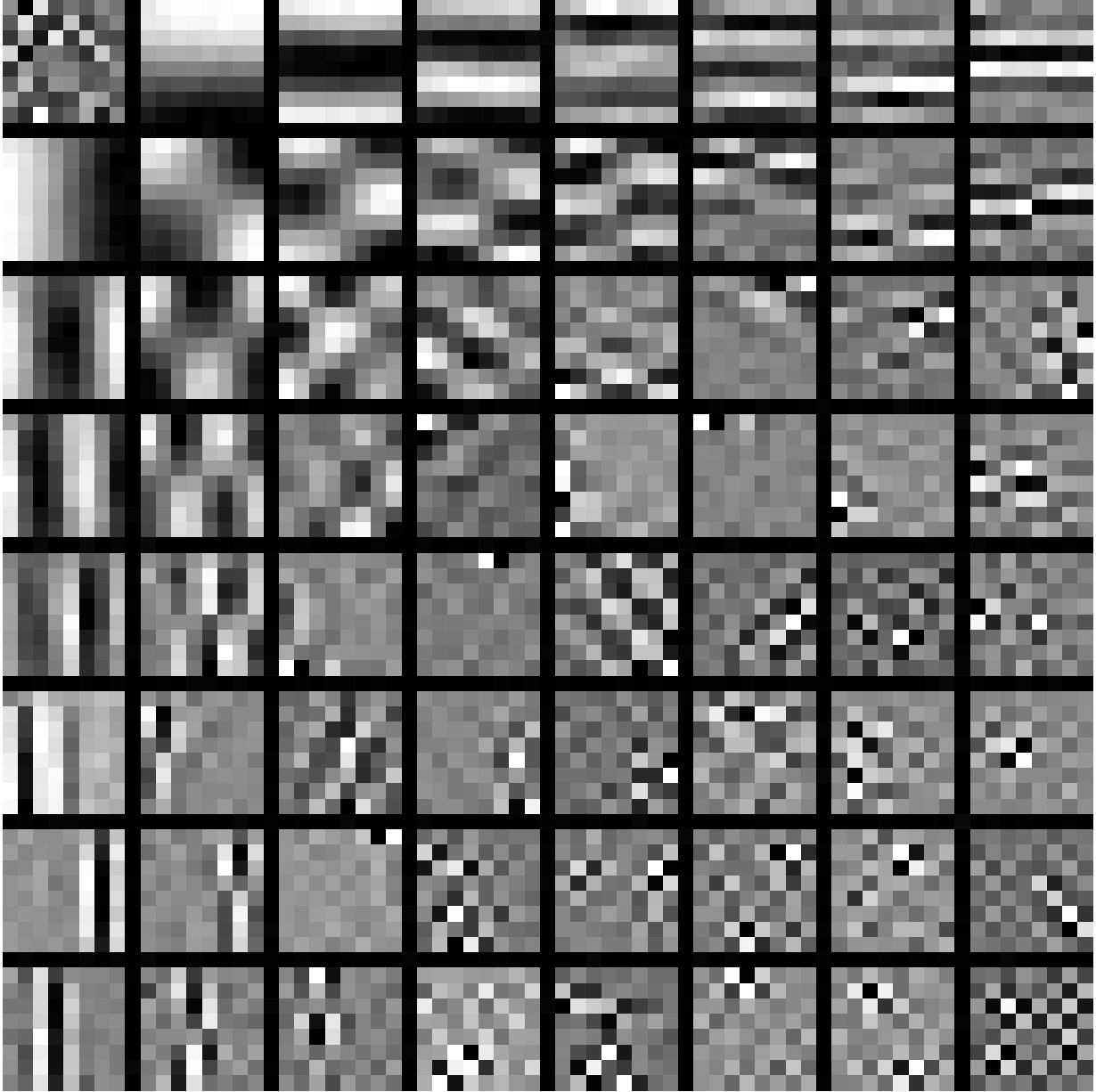}&
\includegraphics[height=1.35in]{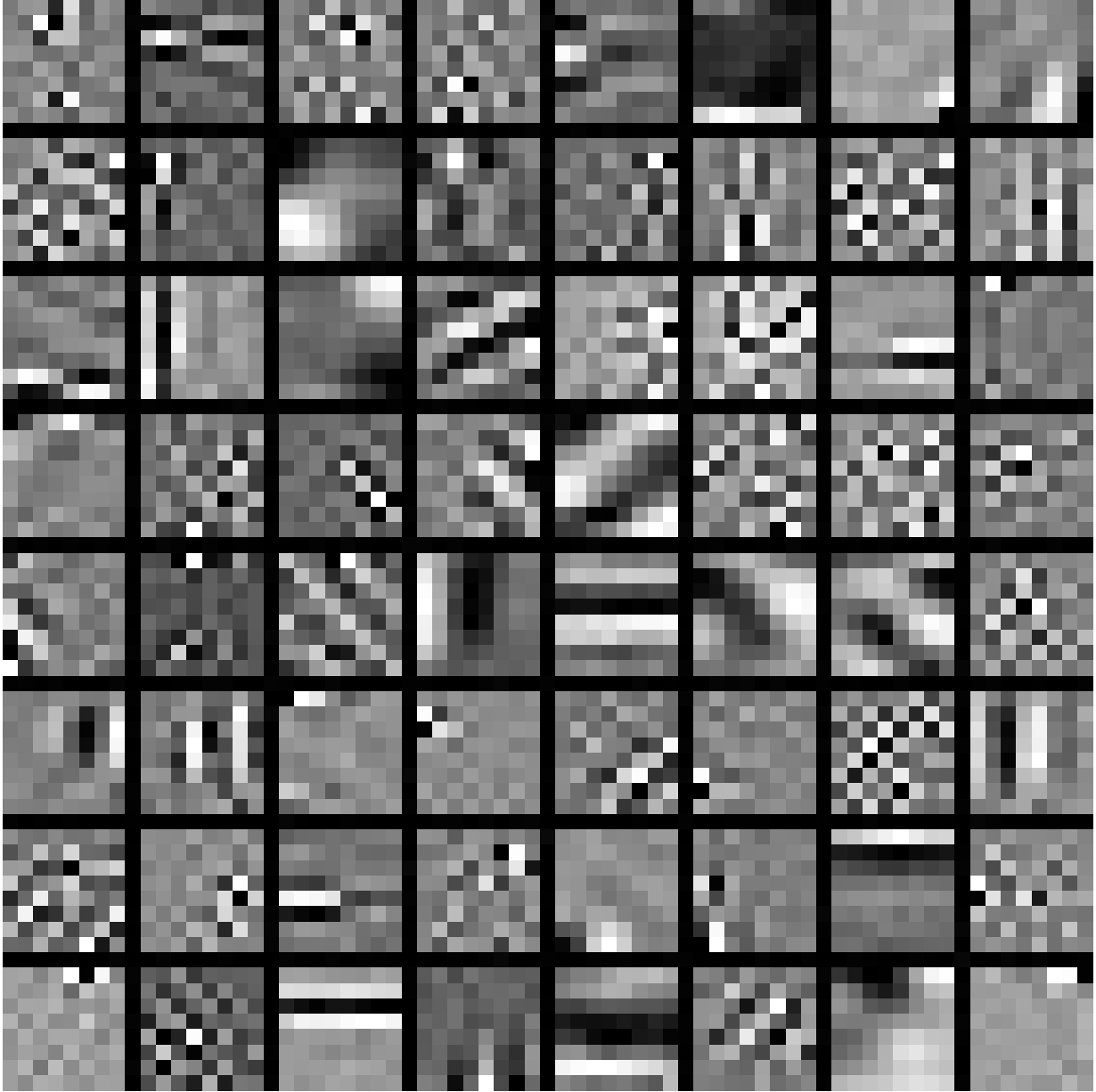}\\
(d) FRIST   & (e) FRIST parent $W$  & (f) FRIST parent $W$ \\
 Sparsification Error & with DCT initialization &  with random initialization \\
\end{tabular}
\end{center}
\caption{Convergence of the FRIST objective, sparsification error, and cluster sizes with various parent transform initializations, as well as the visualizations of the learned FRIST parent transforms with DCT and random initializations.}
\label{convFig}
\end{figure}

Figures \ref{convFig}(a) and \ref{convFig}(d) illustrate that the objective function and sparsification error in (P2) converge to similar values from various initializations of $W$, indicating that the algorithm is reasonably insensitive or robust to initializations in practice. Figures \ref{convFig}(b) and \ref{convFig}(c) show the changes in cluster sizes over iterations for the 2D DCT and KLT initializations. The final values of the corresponding cluster sizes are also similar (although, not necessarily identical) for various initializations. Figure \ref{convFig}(e) and \ref{convFig}(f) display the learned FRIST parent $W$'s with DCT and random matrix initializations. They are non-identical, and capture features that sparsify the image patches equally well. Thus we consider such learned transforms to be essentially equivalent as they achieve similar objective values and sparsification errors for the training data and are similarly conditioned (The learned parent $W$'s with the DCT and random matrix initializations have condition numbers $1.04$ and $1.06$, respectively).

The numerical results demonstrate that our FRIST learning algorithm is reasonably robust, or insensitive to initialization. Good initialization for the parent transform $W$, such as the DCT, leads to faster convergence during learning. Thus, we initialize the parent transform $W$ using the 2D DCT in the rest of the experiments.

\subsection{Image Segmentation and Clustering Behavior} \label{segment}

\begin{figure}
\begin{center}
\begin{tabular}{ccc}
\includegraphics[height=1.5in]{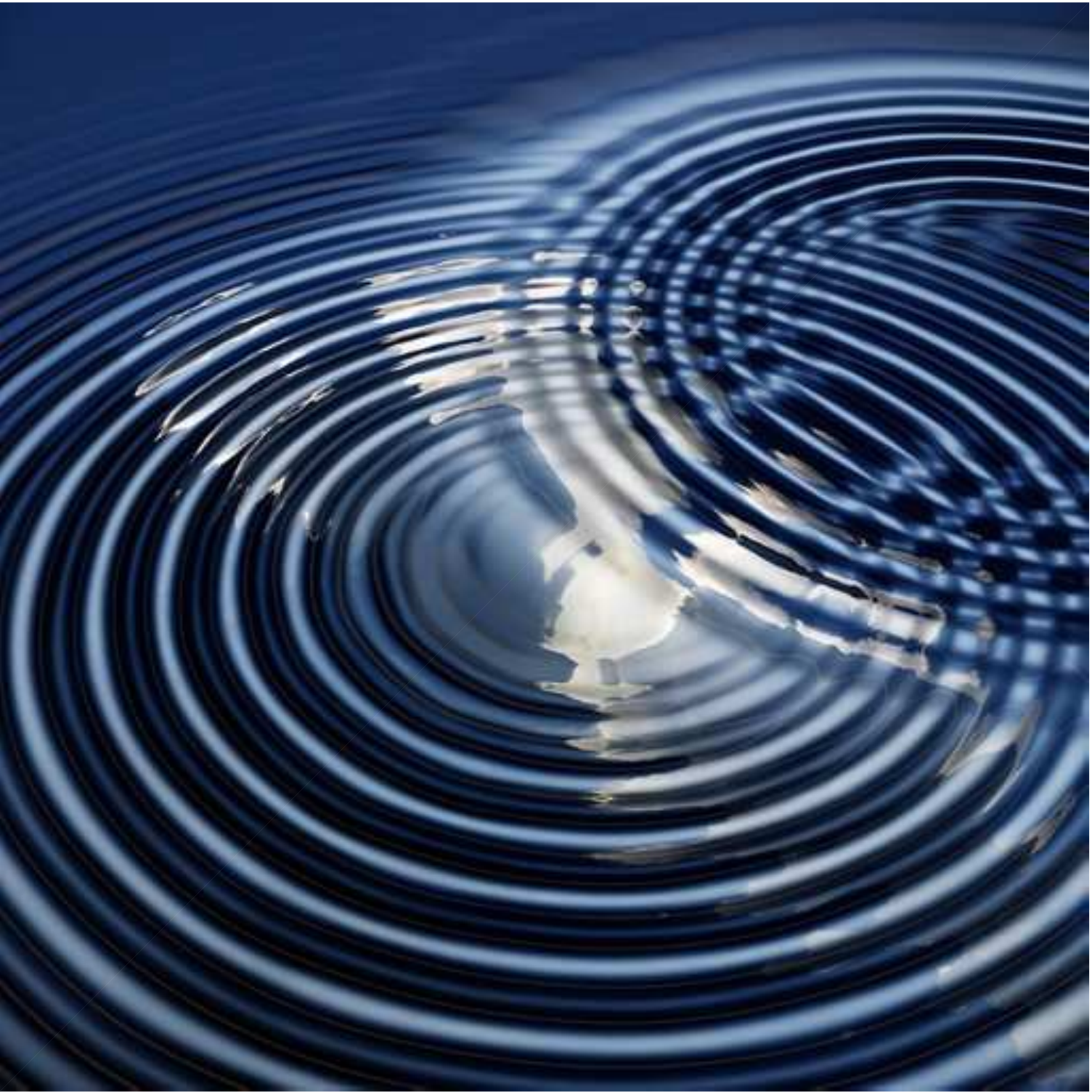}&
\includegraphics[height=1.5in]{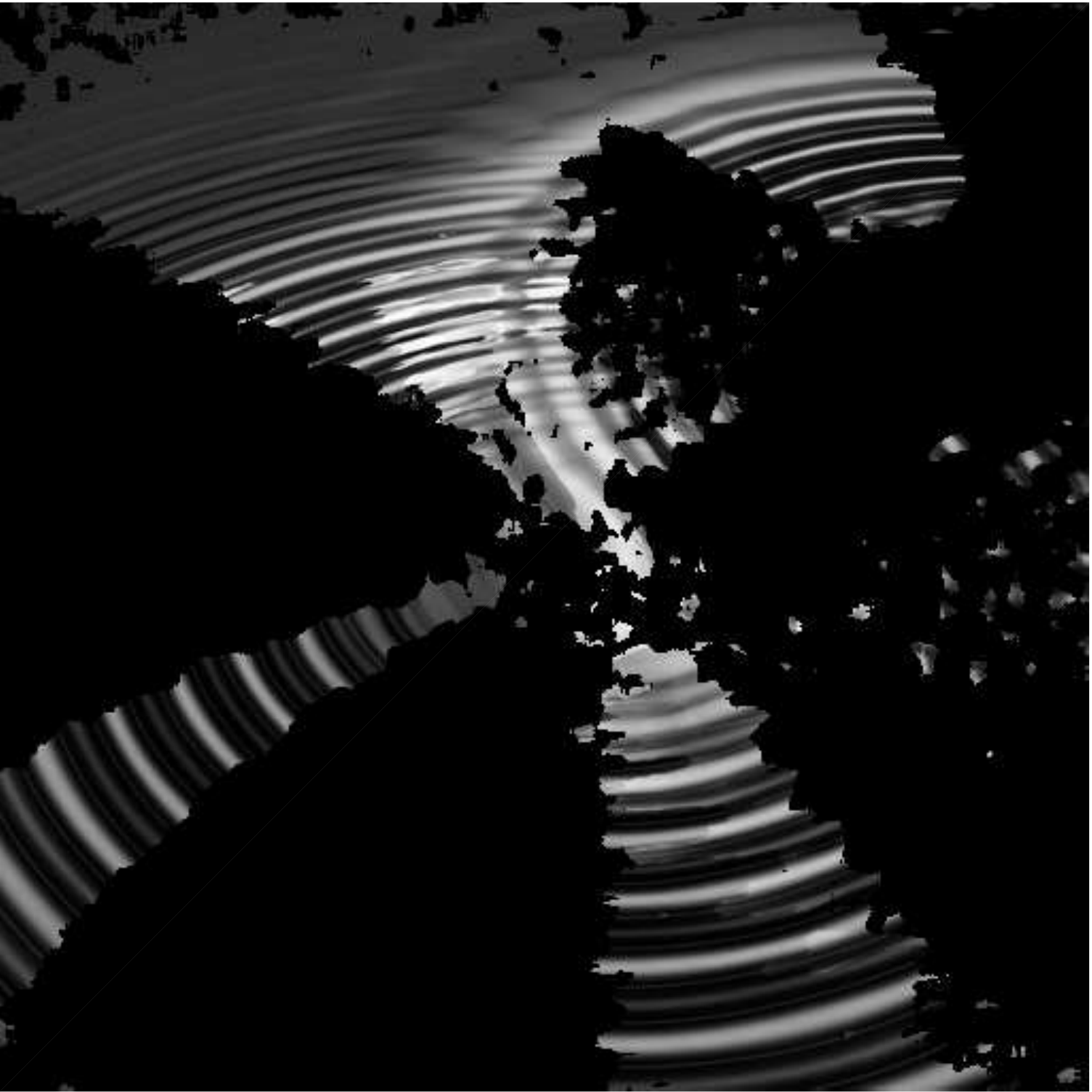}&
\includegraphics[height=1.5in]{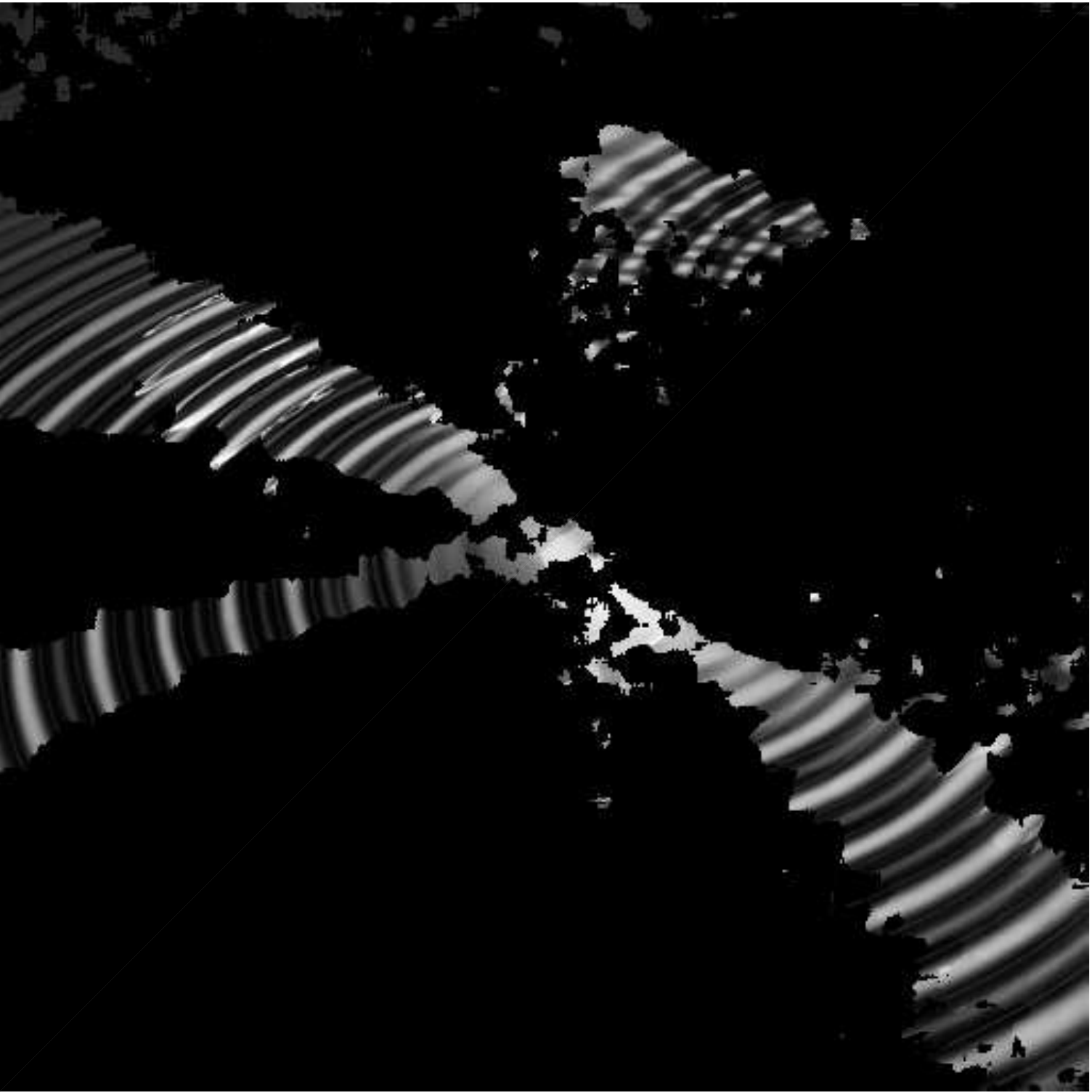}\\
(a) \textit{Wave} & (c) Class 1 & (e) Class 3 \\
 \\
\includegraphics[height=1.5in]{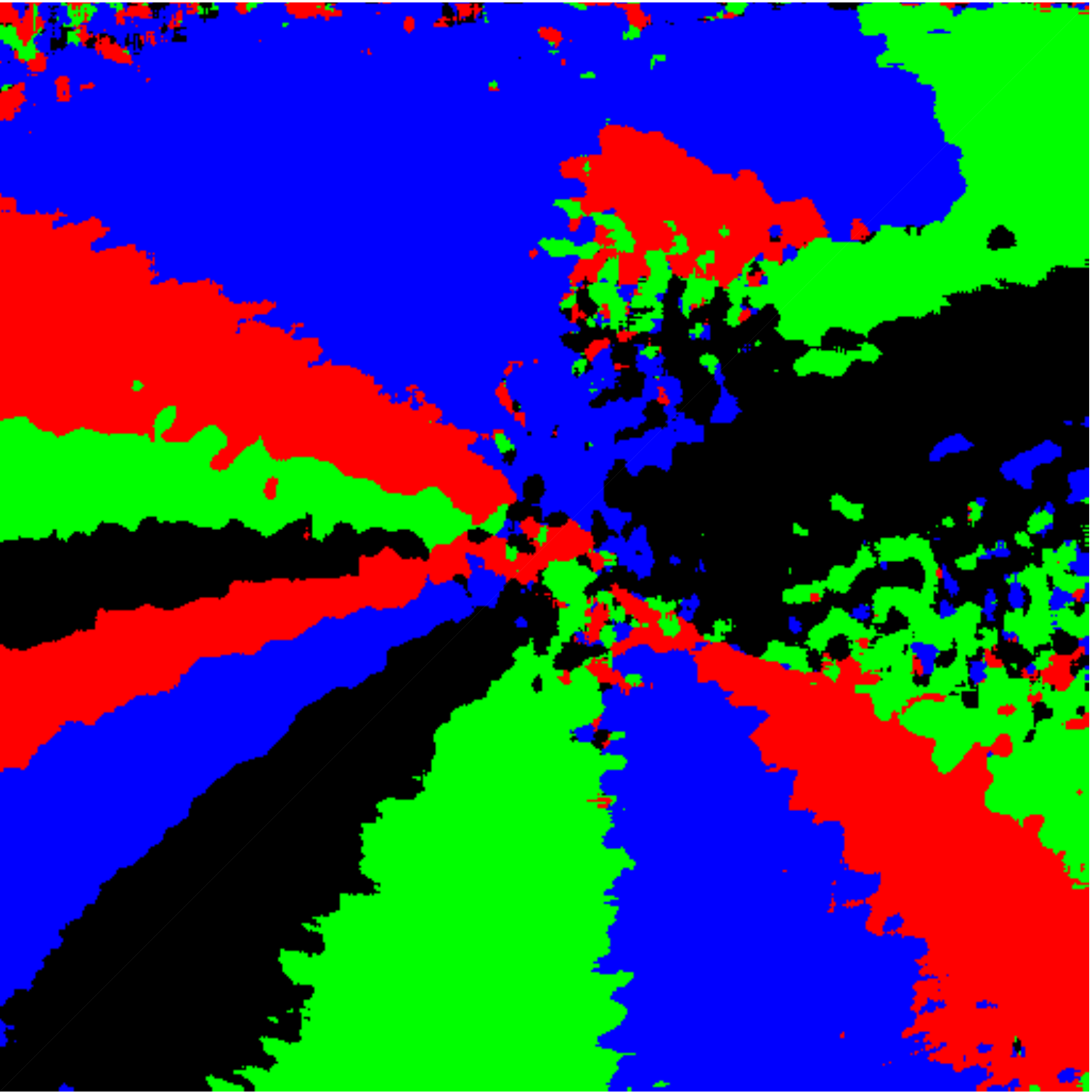}&
\includegraphics[height=1.5in]{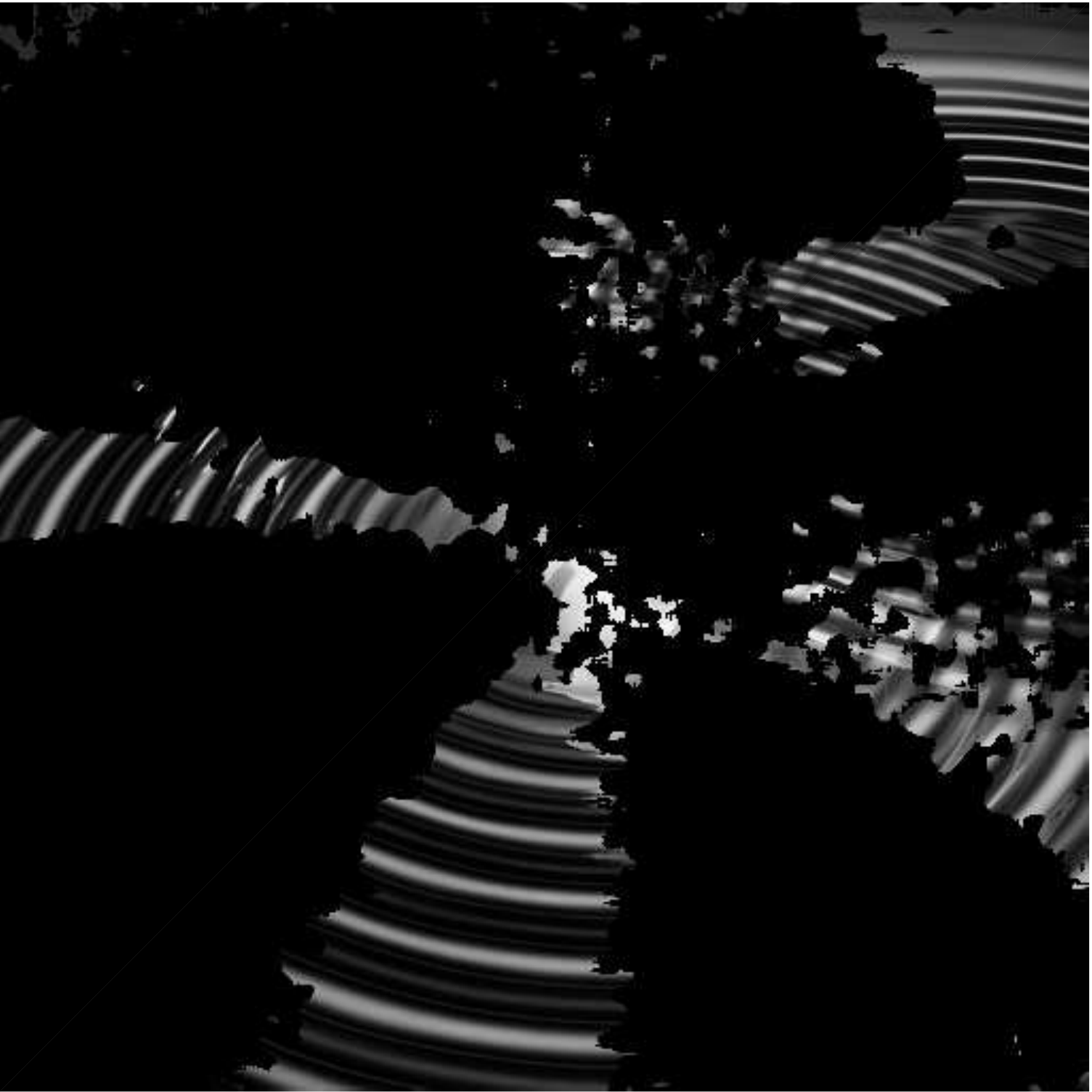}&
\includegraphics[height=1.5in]{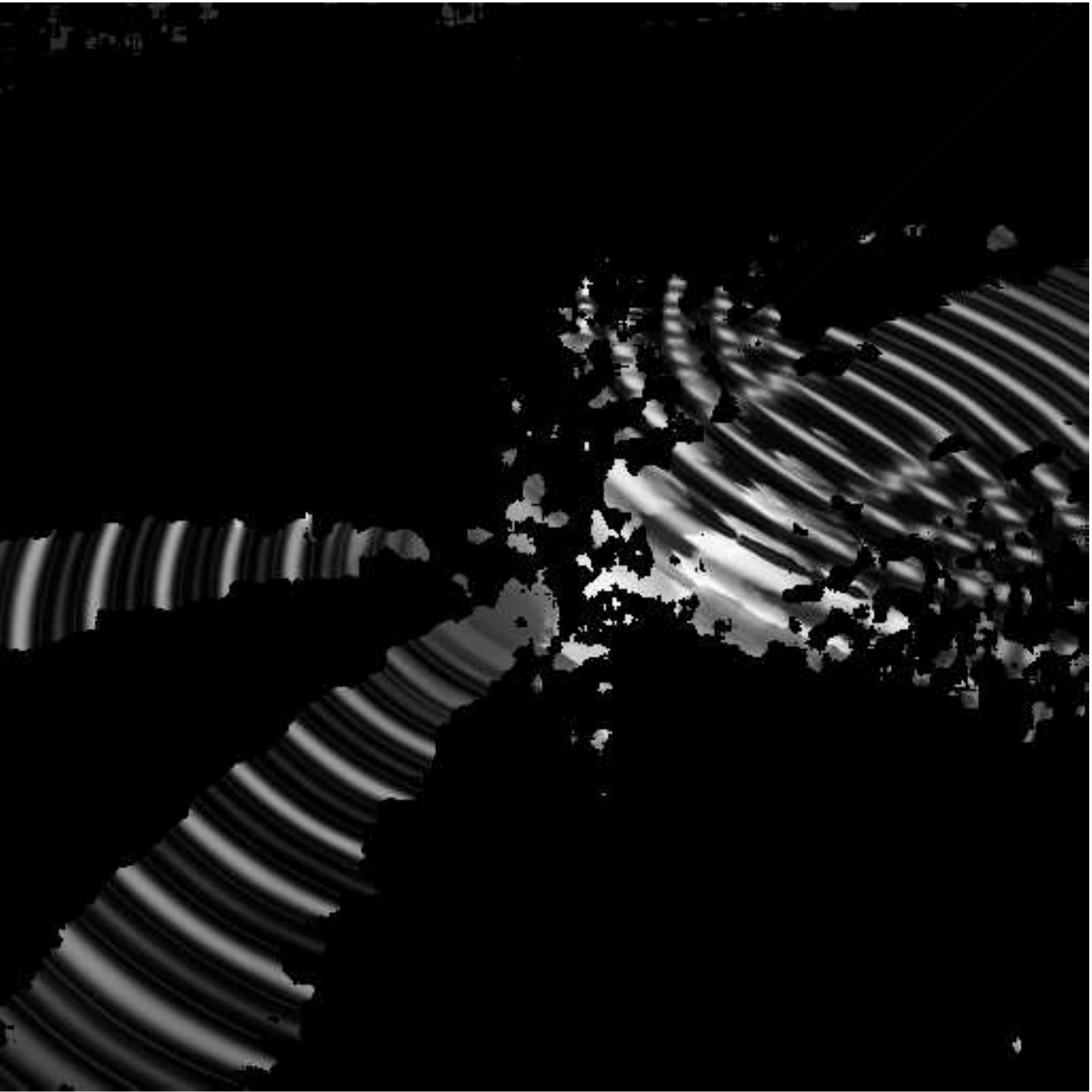}\\
(b) Pixel memberships & (d) Class 2 & (f) Class 4 \\
\end{tabular}
\end{center}
\caption{Image segmentation result of \textit{Wave} ($512 \times 512$) using FRIST learning on the gray-scale version of the image. The colors Red, Green, Blue, and Black in (b), present pixels that belong to the four classes. Pixels that are clustered into a specific class are shown in gray-scale (using intensities in the original gray-scale image), while pixels that are not clustered into that class are shown in black for (c)-(f).}
\label{clusteringWave}
\end{figure}

\begin{figure}
\begin{center}
\begin{tabular}{ccc}
\includegraphics[height=0.85in]{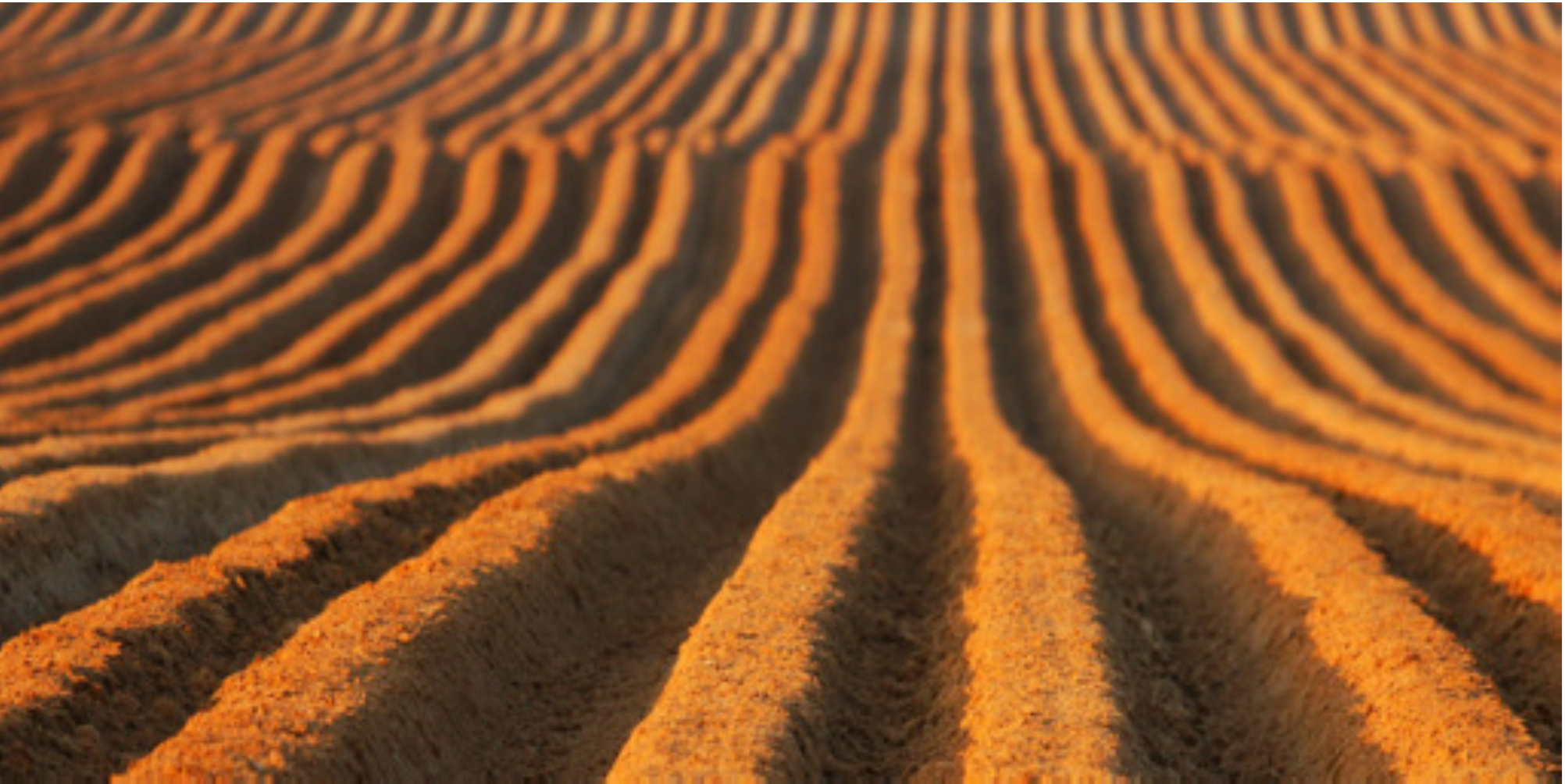}&
\includegraphics[height=0.85in]{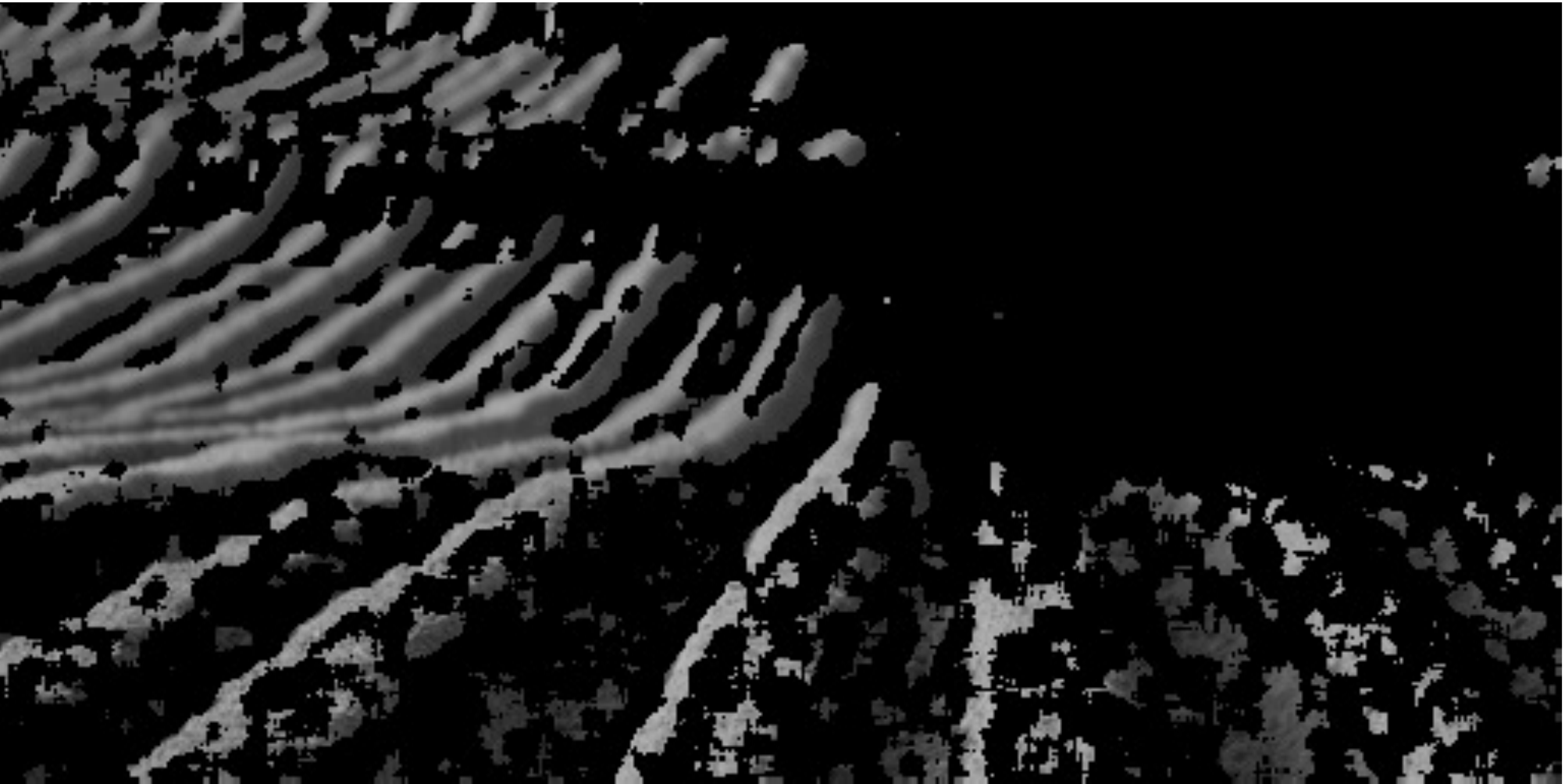}&
\includegraphics[height=0.85in]{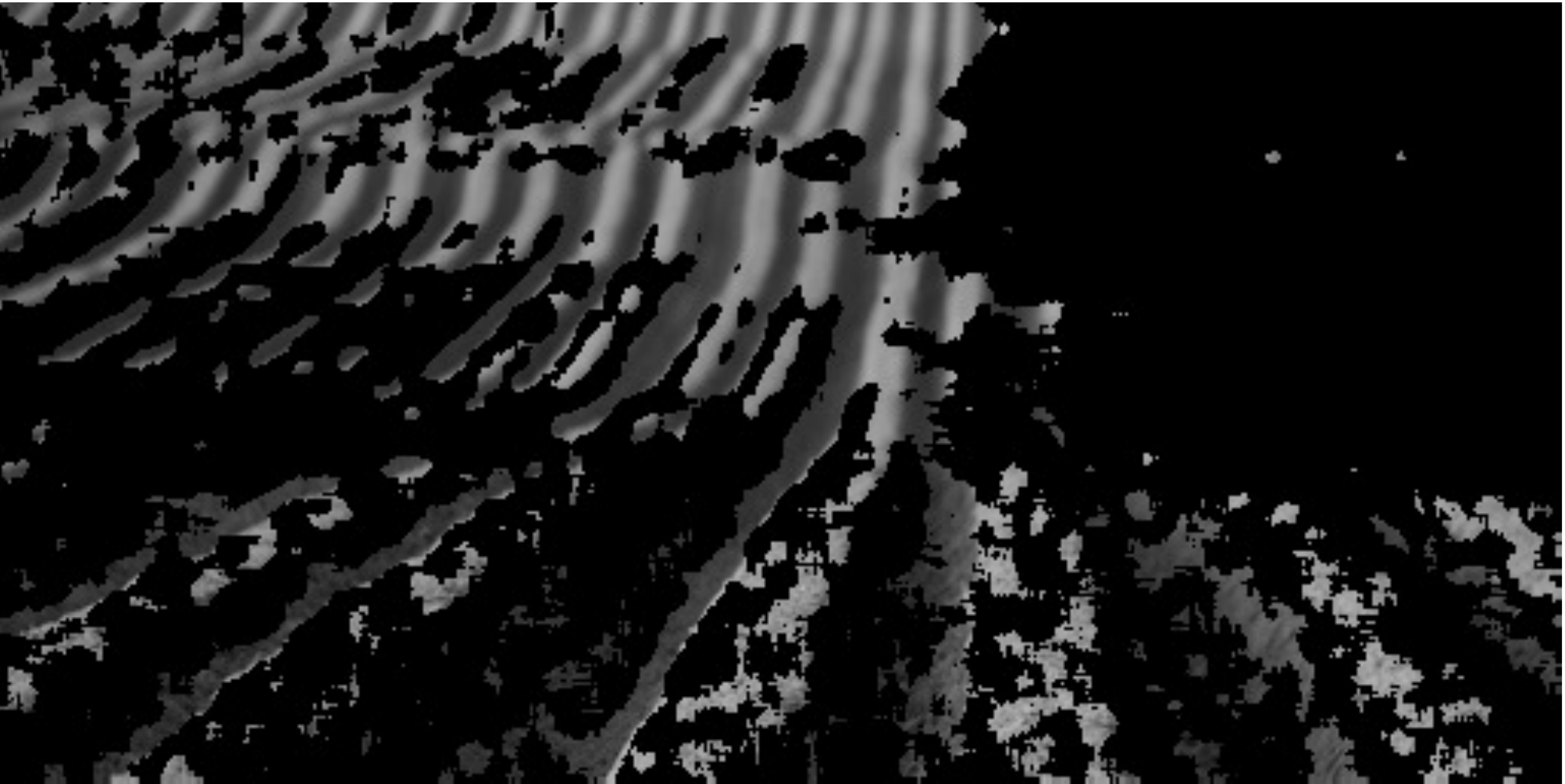}\\
(a) \textit{Field} & (c) Class 1 & (e) Class 3 \\
 \\
\includegraphics[height=0.85in]{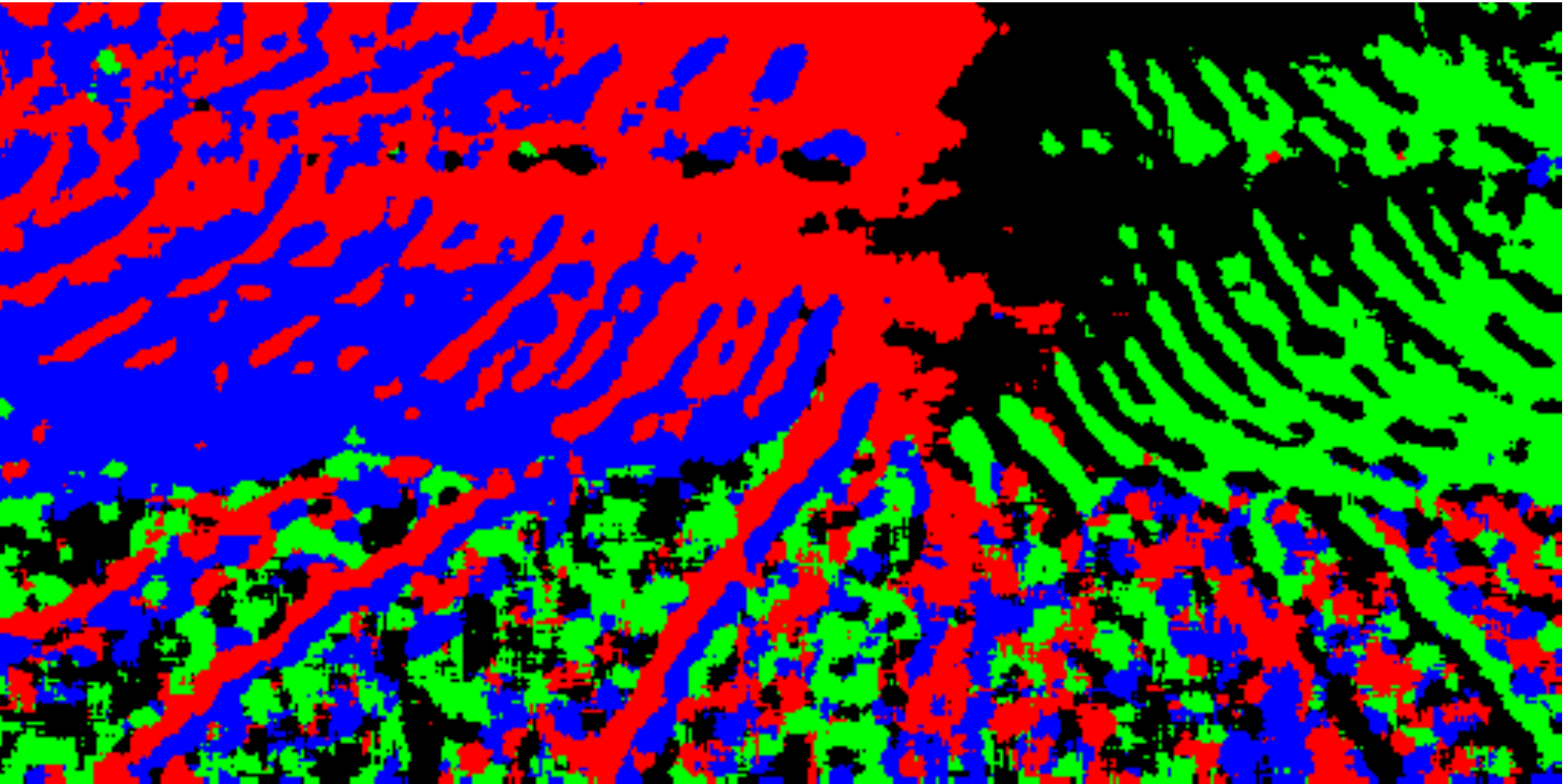}&
\includegraphics[height=0.85in]{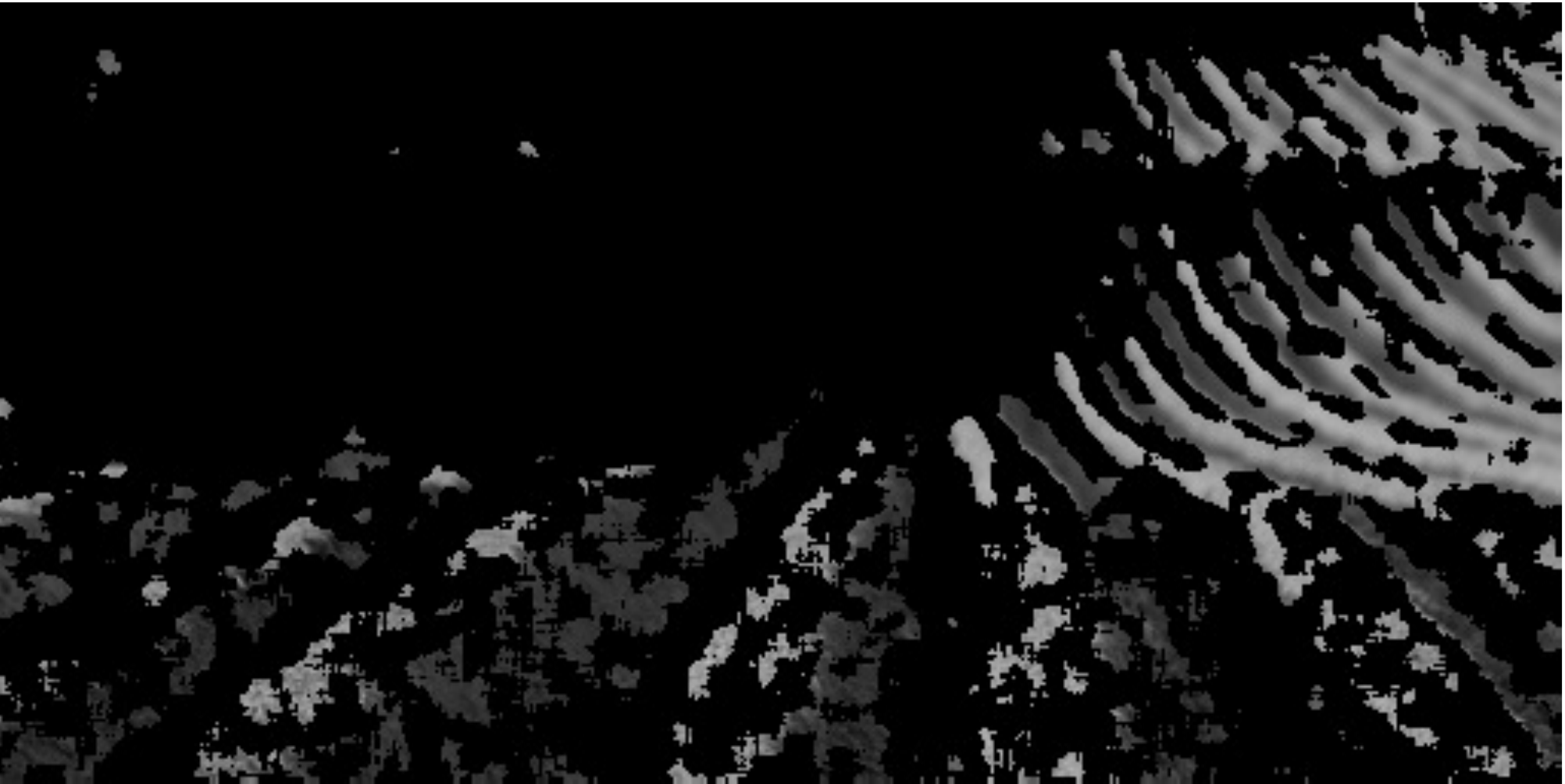}&
\includegraphics[height=0.85in]{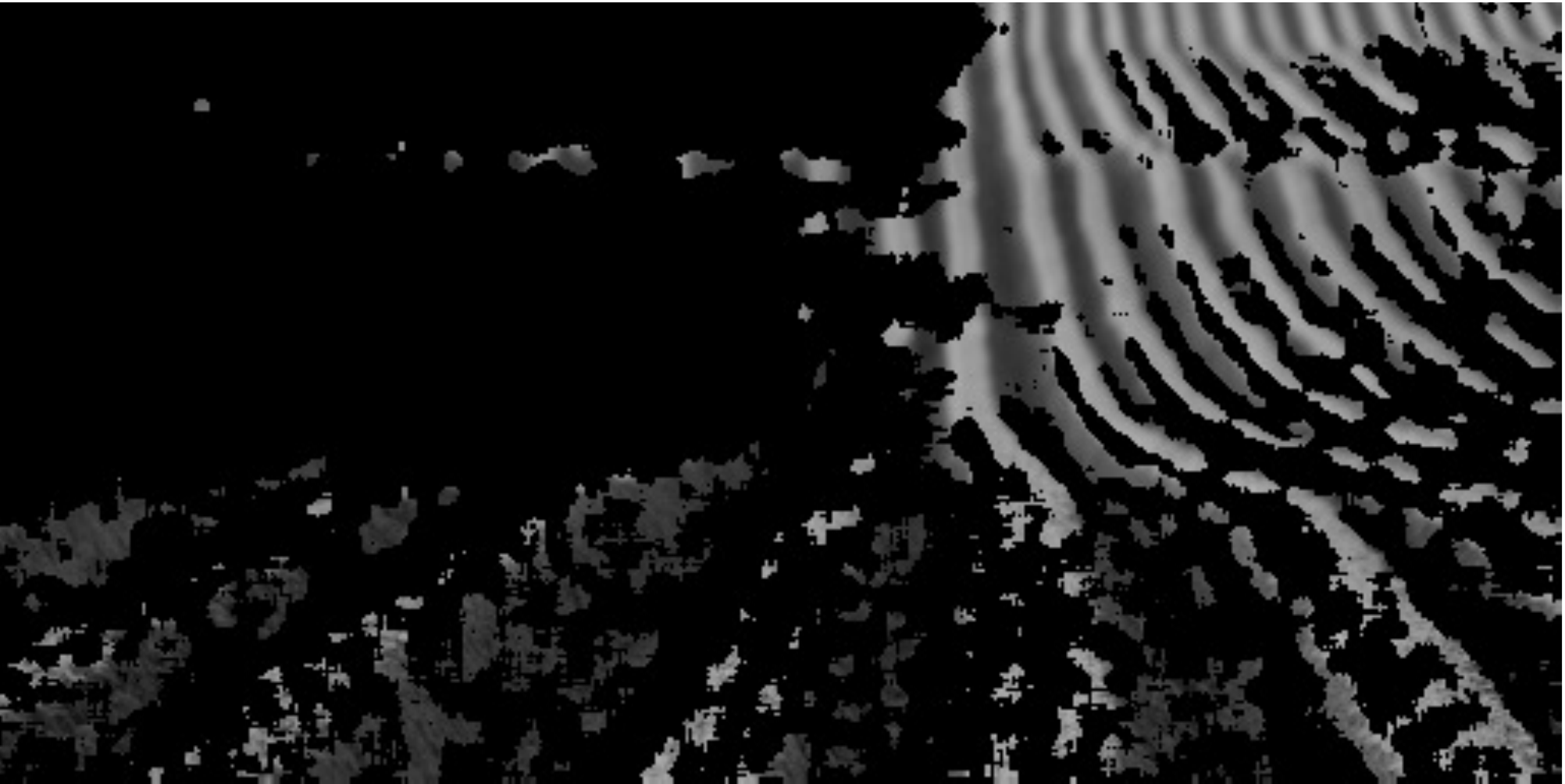}\\
(b) Pixel memberships & (d) Class 2 & (f) Class 4 \\
\end{tabular}
\end{center}
\caption{Image segmentation result of \textit{Field} ($256 \times 512$) using FRIST learning on the gray-scale version of the image. The colors Red, Green, Blue, and Black in (b), present pixels that belong to the four classes. Pixels that are clustered into a specific class are shown in gray-scale (using intensities in the original gray-scale image), while pixels that are not clustered into that class are shown in black for (c)-(f).}
\label{clusteringField}
\end{figure}

\begin{figure}
\begin{center}
\begin{tabular}{c}
\includegraphics[height=1.4in]{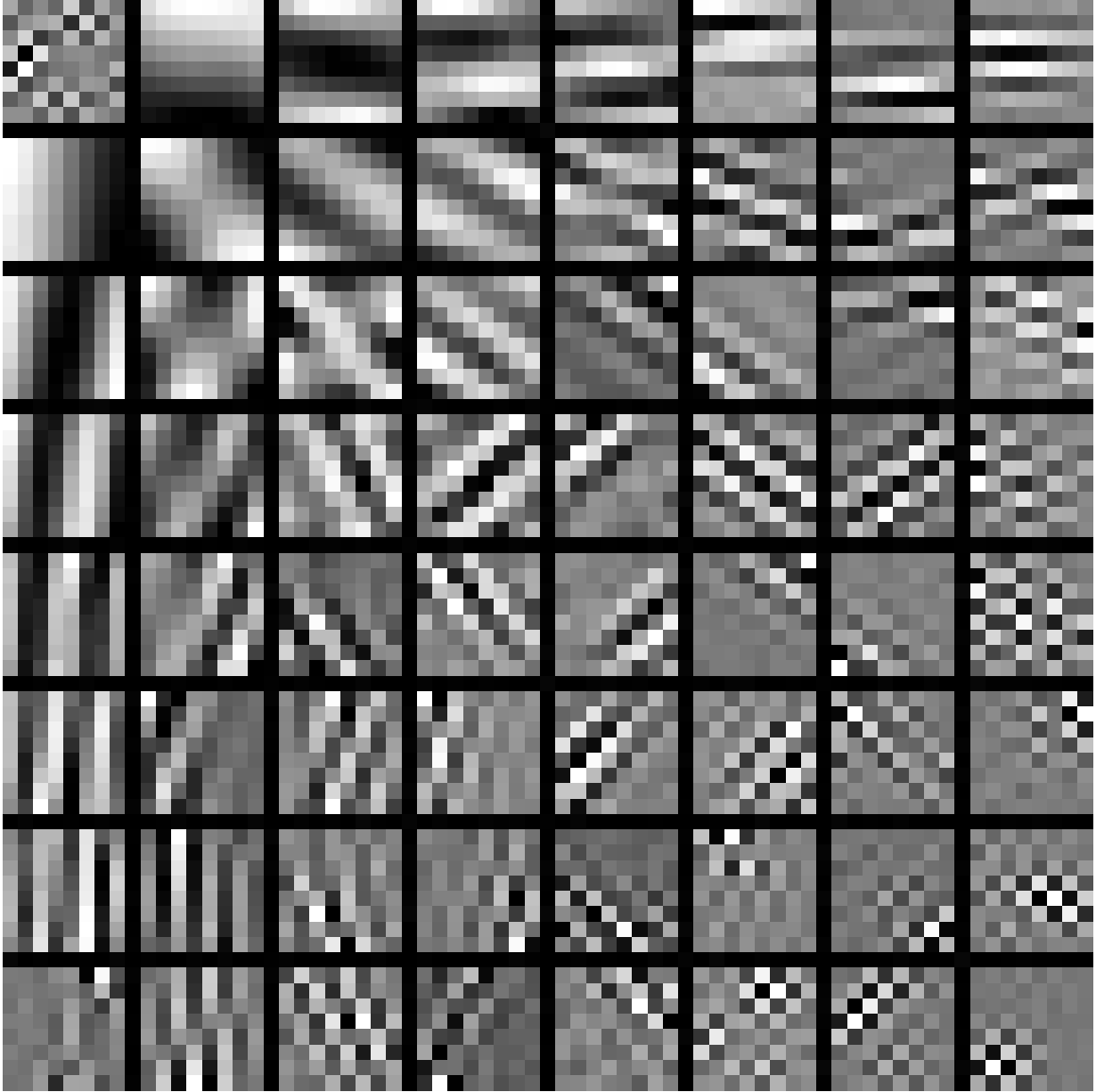}\\
(a) Parent transform\\
  \\
\end{tabular}
\begin{tabular}{cccc}
\includegraphics[height=1.3in]{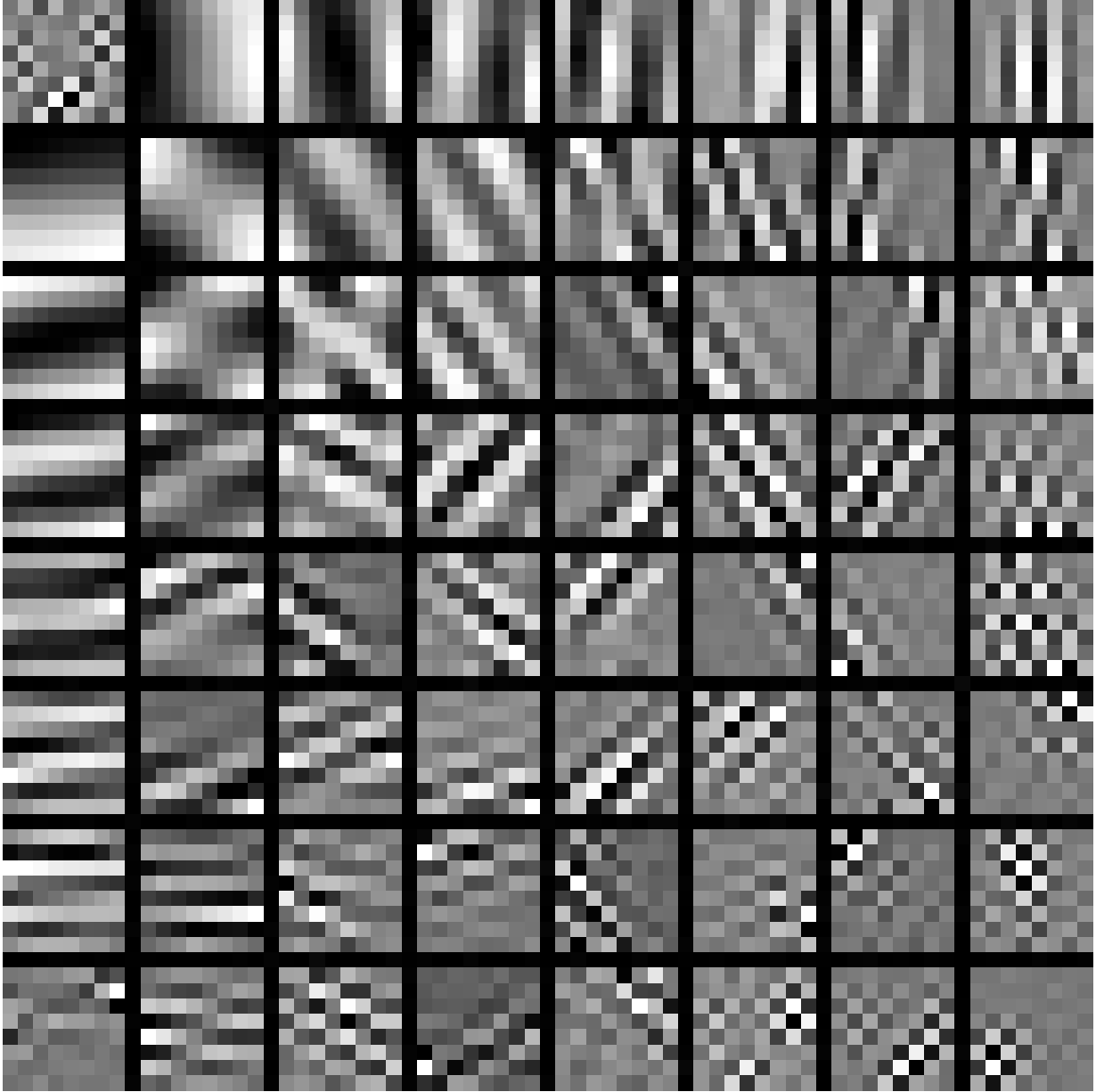}&
\includegraphics[height=1.3in]{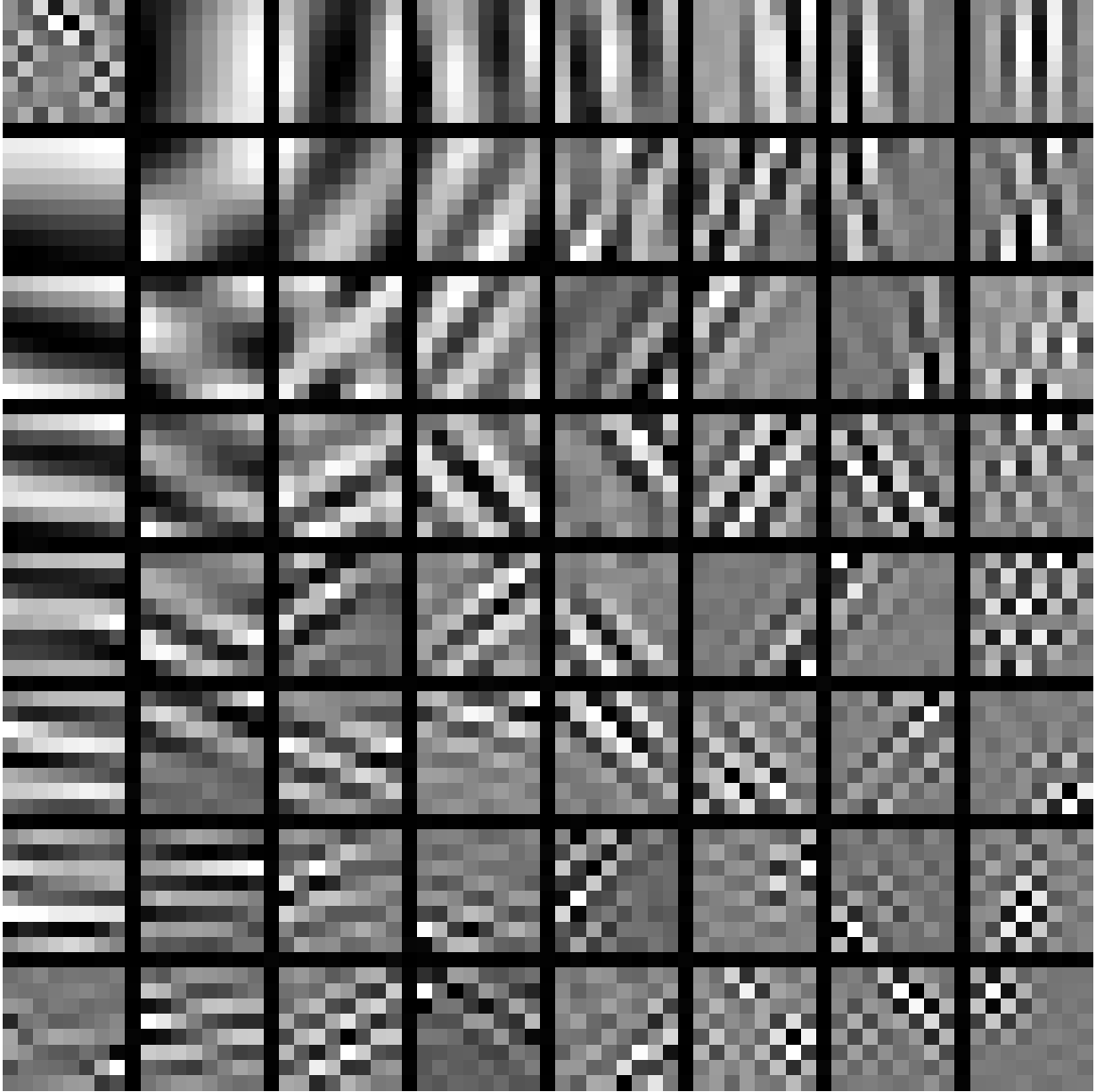}&
\includegraphics[height=1.3in]{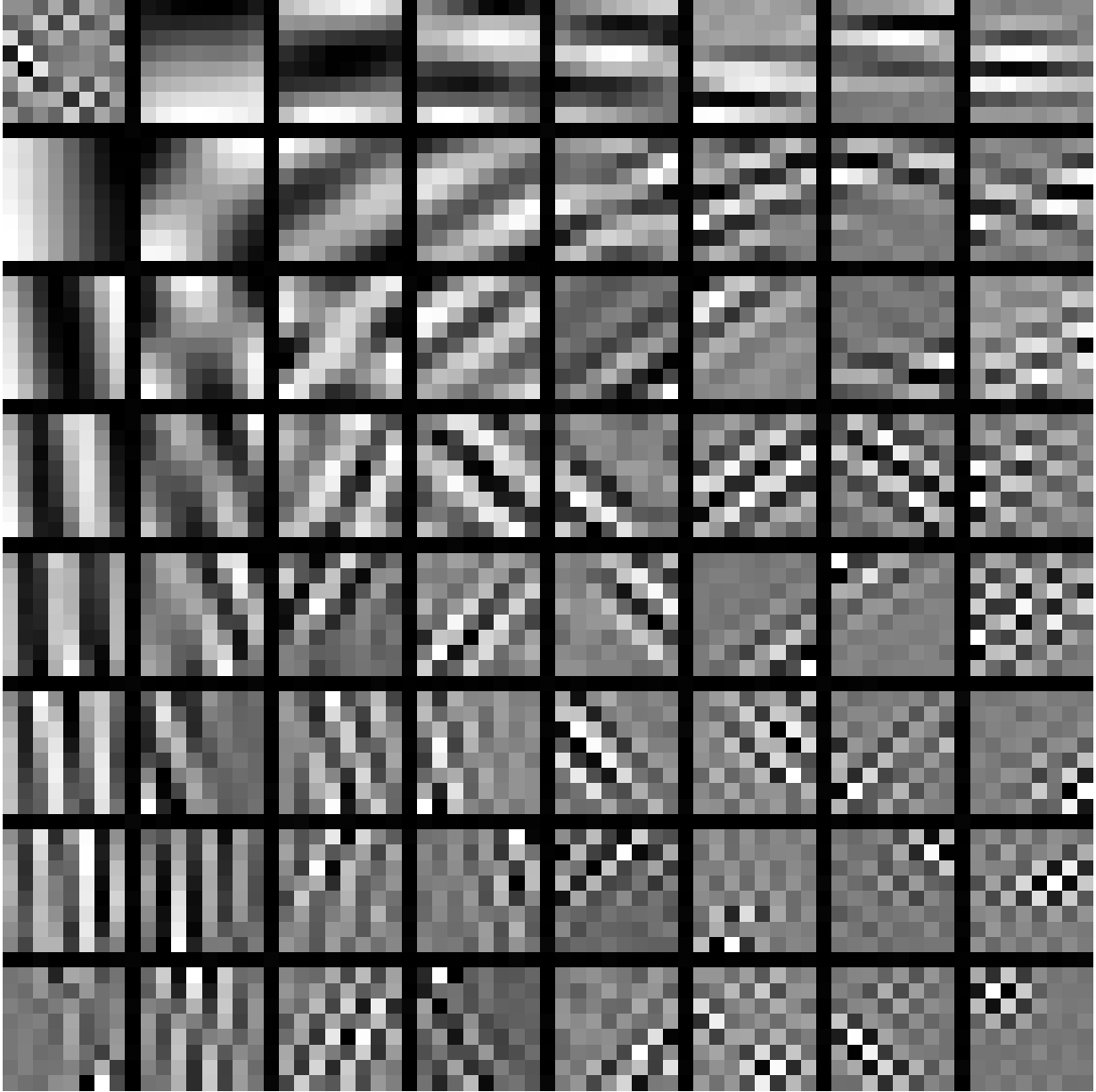}&
\includegraphics[height=1.3in]{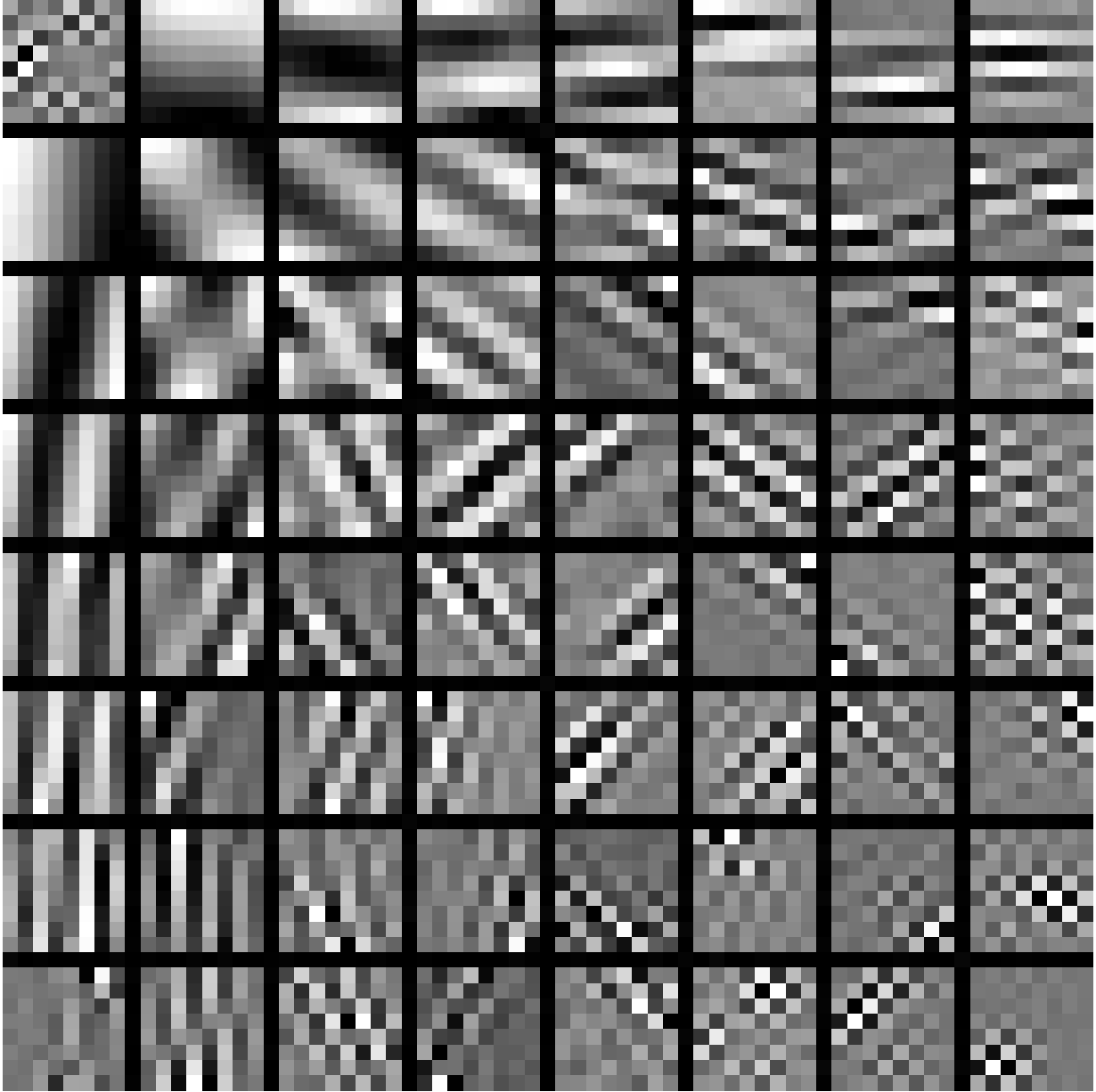}\\
(b) & (c) & (d) & (e)
\end{tabular}
\end{center}
\caption{Visualization of the learned (a) parent transform, and (b)-(e) children transforms in FRIST for the image \textit{Wave}. The rows of each child transform are displayed as $8 \times 8$ patches.}
\label{transformVisual}
\end{figure}

The FRIST learning algorithm is capable of clustering image patches according to their orientations. In this subsection, we illustrate the FRIST clustering behavior by image segmentation experiments. We consider the images \textit{Wave} ($512 \times 512$) and \textit{Field} ($512 \times 512$) shown in Fig. \ref{clusteringWave}(a) and Fig. \ref{clusteringField}(a) as inputs. Both images contains directional textures, and we aim to cluster the pixels of the images into one of four classes, which represent different orientations or flips. For each input image, we convert it into gray-scale, extract the overlapping mean-subtracted patches, and learn a FRIST while clustering the patches using the algorithm in Section \ref{algorithm}. As overlapping patches are used, each pixel in the image belongs to several overlapping patches. We cluster a pixel into a particular class by majority voting among the patches that contain it.

We set $s = 10$, and $K=4$ in the clustering experiments. Figure \ref{clusteringWave} and Figure \ref{clusteringField} illustrate the segmentation results of images \textit{Wave} and \textit{Field}, respectively. Figure \ref{clusteringWave}(b) and Figure \ref{clusteringField}(b) illustrate the pixel memberships with four different colors (blue, red, green, and black, for classes 1 to 4, respectively). Figures \ref{clusteringWave}(c)-(f) and Figures \ref{clusteringField}(c)-(f) each visualize the image pixels clustered into a specific class in gray-scale, and the pixels that are not clustered into that class are shown in black. Each class captures edges at specific orientations.

The parent transform $W$ and its children transforms $W_k$'s in the learned FRIST for the \textit{Wave} image are visualized in Fig. \ref{transformVisual} with the rows of each $W_k$ displayed as $8 \times 8$ patches. We observe that each child transform contains distinct directional features that were adaptively learned to sparsify edges with specific orientations better. The parent $W$ turns out to be identical to the child transform shown in Fig. \ref{clusteringField}(e), implying that the corresponding FR operator is the identity matrix.

The preliminary image segmentation results here demonstrate some potential for the FRIST scheme for directional classification or segmentation. More importantly, we wish to illustrate why FRIST can provide improvements over SST or OCTOBOS in various inverse problems. As natural images usually contain a variety of directional features and edges, FRIST is capable of grouping those patches with similar orientations/flips, and thus provides better sparsification in each cluster using directional children transforms, even while learning only a single small parent transform $W$ (which could be learned even in cases of very limited or corrupted data).

\subsection{Sparse Image Representation} \label{sr}

\begin{table}[t]
\caption{PSNR values for reconstruction of images from sparse representations obtained using the 2D DCT, learned SST and OCTOBOS, square and overcomplete K-SVD, and learned FRIST. The first row of the table provides average PSNR values computed over the $44$ images from the USC-SIPI database. The best PSNR values are marked in bold.}
\label{compression}
\begin{center}
\fontsize{10}{16pt}\selectfont
\begin{tabular}{|c|c|c|c|c|c|c|}
\hline
\multirow{1}{*}{Methods} & \multirow{2}{*}{2D DCT} & \multirow{1}{*}{SST} & \multirow{1}{*}{OCTOBOS}   & \multicolumn{2}{c|}{K-SVD} & \multirow{1}{*}{\textbf{FRIST}}\\ 
\cline{3-7}  
    Model Size    &       &      $64 \times 64$    &     $128 \times 64$ &             $64 \times 64$   	 	&    $64 \times 128$  & $64 \times 64$  \\
\hline
\textit{USC-SIPI} & 34.36 & 34.20 & 33.62 & 34.11 & 35.08 & \textbf{35.14}\\
\hline
\textit{Cameraman} & 29.49 & 29.43 & 29.03 & 29.09 & 30.16 & \textbf{30.63}\\
\hline
\textit{House} & 36.89 & 36.36 & 35.38 & 36.31 & 37.41 & \textbf{37.71}\\
\hline
\end{tabular}
\end{center}
\end{table}

Most of the popular image compression methods make use of analytical sparsifying transforms. In particular, the commonly used JPEG uses the 2D DCT to sparsify image patches.
Data-driven adaptation of dictionaries using the K-SVD scheme has also been shown to be beneficial for image compression, compared to fixed analytical transforms \cite{ksvd2008compression}. 
In this section, we show that the proposed FRIST learning scheme provides improved sparse representations of images compared to related adaptive sparse modeling methods. While we focus here on a simple study of the sparse representation abilities of adaptive FRIST, the investigation of a complete adaptive image compression framework based on FRIST and its comparison to benchmarks is left for future work.

We learn a FRIST, with a $64 \times 64$ parent transform $W$, from the $10^{4}$ randomly selected patches (from USC-SIPI images) used in Section \ref{convergeSupp}. We set $K = 32$, $s = 10$ and $\lambda_0 = 3.1 \times 10^{-3}$. We compare the learned FRIST with other popular adaptive sparse signals models. In particular, we train a $64 \times 64$ SST \cite{sabres3}, a $128 \times 64$ OCTOBOS \cite{wensabres}, as well as a $64 \times 64$ square (synthesis) dictionary and a $64 \times 128$ overcomplete dictionary using KSVD \cite{elad2}, using the same training patches and sparsity level as for FRIST.

With the learned models, we represent each image from the USC-SIPI database as well as some other standard images. Each image is represented compactly by storing its sparse representation including $(i)$ non-zero coefficient values in the sparse codes of the $8 \times 8$ non-overlapping patches, $(ii)$ locations of the non-zeros (plus the cluster membership if necessary) in the sparse codes for each patch and $(iii)$ the adaptive sparse signal model (e.g., the dictionary or transform matrix -- this would typically involve negligible overhead). For each method, the patch sparsity (or equivalently, the number of non-zero coefficients per patch) is set to $s=10$ (same as during training). The adaptive SST, square KSVD, and adaptive FRIST methods store only a $64 \times 64$ square matrix in $(iii)$ above, whereas the overcomplete KSVD and OCTOBOS methods store a $128 \times 64$ matrix.

The images (i.e., their non-overlapping patches) are reconstructed from their sparse representations in a least squares sense, and the reconstruction quality for each image is evaluated using the Peak-Signal-to-Noise Ratio (PSNR), expressed in decibels (dB). We use the average of the PSNR values over all 44 USC-SIPI images as the indicator of the quality of sparse representation of the USC-SIPI database.

Table \ref{compression} lists the sparse representation reconstruction results for the USC-SIPI database and the images \textit{Cameraman} ($256 \times 256$) and \textit{House} ($256 \times 256$). We observe that the learned FRIST model provides the best reconstruction quality compared to other adaptive sparse signal models or the analytical 2D DCT, for both the USC-SIPI images and the external images. 
Compared to unstructured overcomplete models such as KSVD and OCTOBOS, the proposed FRIST provides improved PSNRs, while achieving potentially fewer bits for sparse representation \footnote{Assuming for simplicity that $L$ bits are used to describe each non-zero coefficient value in each model, the total number of bits for storing the sparse code (non-zero locations, non-zero coefficient values, cluster membership) of a patch is $6s+Ls+5$ in $64 \times 64$ FRIST ($K = 32$), and $7s+Ls$ in $64 \times 128$ KSVD. For the setting $s = 10$, FRIST requires 5 fewer bits per patch compared to KSVD.}.
Additionally, dictionary learning based representation requires synthesis sparse coding, which is more expensive compared to the cheap and exact sparse coding in the transform model-based methods \cite{sabres}. As mentioned before, the investigation of an image compression system based on learned FRIST models, and its analysis as well as quantitative comparison to other compression benchmarks is left for future work.

\subsection{Image Denoising} \label{denoising}

\begin{table}[t]
\caption{PSNR values (in $dB$) for denoising with $64 \times 64$ adaptive FRIST along with the corresponding PSNR values for denoising using the $64 \times 64$ 2D DCT, the $64 \times 64$ adaptive SST, the $64 \times 256$ overcomplete K-SVD, and the $256 \times 64$ learned OCTOBOS. The best PSNR values are marked in bold.}
\label{denoisingTable}
\begin{center}
\fontsize{10}{18pt}\selectfont
\begin{tabular}{|c|c|c|c|c|c|c|c|} 
\hline
\multirow{2}{*}{Image} & \multirow{2}{*}{$\sigma$} & \multicolumn{1}{c|}{Noisy} & \multirow{2}{*}{DCT} & \multirow{2}{*}{SST \cite{sabres3}}  & \multirow{2}{*}{K-SVD \cite{elad2}}   & \multirow{2}{*}{OCTOBOS \cite{wensabres}}  & \multirow{2}{*}{FRIST} \\ 
           &               &  PSNR   & &  & &   &                 \\
\hline
\multirow{2}{*}{\textit{Peppers}} & 5 & 34.14 & 37.70 & 37.95 & 37.78 & 38.09 & \textbf{38.16} \\ \cline{2-8}		
 & 10 & 28.10 & 34.00 & 34.37 & 34.24 & 34.57 & \textbf{34.68} \\  \cline{2-8}				
($256 \times 256$) & 15 & 24.58 & 31.83 & 32.14 &  32.18 & 32.43 & \textbf{32.54}\\ \cline{2-8}
  & 20 & 22.12 & 30.06 & 30.62 & 30.80 & 30.97 & \textbf{31.02}\\ 
\hline
\multirow{2}{*}{\textit{Cameraman}} & 5 & 34.12 & 37.77 & 38.01 & 37.82 & \textbf{38.16} & \textbf{38.16} 	\\ \cline{2-8}	
 & 10 & 28.14 & 33.63 & 33.90 & 33.72 & 34.13 & \textbf{34.16} \\ \cline{2-8}	
($256 \times 256$) & 15 & 24.61 & 31.33 & 31.65 & 31.51 & 31.95 & \textbf{31.97} \\ \cline{2-8}	
& 20 & 22.10 & 29.81 & 29.91 & 29.82 & 30.24 & \textbf{30.33} \\
\hline
\multirow{2}{*}{\textit{Man}} & 5 & 34.15 & 36.59 & 36.64  & 36.47 & 36.73 & \textbf{36.82}\\ \cline{2-8}		
 & 10 & 28.13 & 32.86 & 32.95 & 32.71 & 32.98 & \textbf{33.06} \\  \cline{2-8}			
 ($768 \times 768$) & 15  & 24.63 & 30.88 & 30.96 & 30.78 & 31.07 & \textbf{31.10} \\  \cline{2-8}	
& 20  & 22.11 & 29.42 & 29.58 & 29.40 & 29.74 & \textbf{29.76} \\
\hline
\multirow{2}{*}{\textit{Couple}} & 5 & 34.16 & 37.25 & 37.32 & 37.29 & 37.40 & \textbf{37.43} \\ \cline{2-8}		
 & 10 & 28.11 & 33.48 & 33.60 & 33.50 & 33.73 & \textbf{33.78} \\  \cline{2-8}			
 ($512 \times 512$) & 15 & 24.59 & 31.35 & 31.47 & 31.44 & \textbf{31.71} & \textbf{31.71}\\ \cline{2-8}	
& 20 & 22.11 & 29.82 & 30.01 & 30.02 & 30.34 & \textbf{30.36}\\
\hline

\multirow{2}{*}{\textit{Kodak 5}} & 5 & 34.17 & 36.72 & 36.96 & 36.32 & 37.10 & \textbf{37.17} \\ \cline{2-8}		
 & 10 & 28.12 & 32.03 & 32.33 & 31.86 & 32.57 & \textbf{32.62} \\  \cline{2-8}			
 ($768 \times 512$) & 15 & 24.60 & 29.51 & 29.84 & 29.49 & 30.13 & \textbf{30.16}\\ \cline{2-8}	
& 20 & 22.13 & 27.79 & 28.09 & 27.89 & 28.40 & \textbf{28.47}\\
\hline
\multirow{2}{*}{\textit{Kodak 9}} & 5 & 34.14 & 39.35 & 39.45 & 38.85 & \textbf{39.53} & \textbf{39.53} \\ \cline{2-8}		
 & 10 & 28.15 & 35.66 & 35.98 & 35.39 & 36.23 & \textbf{36.26} \\  \cline{2-8}			
 ($512 \times 768$) & 15 & 24.60 & 33.36 & 33.89 & 33.39 & 34.27 & \textbf{34.28}\\ \cline{2-8}	
& 20 & 22.11 & 31.66 & 32.30 & 31.90 & 32.73 & \textbf{32.76}\\
\hline
\multirow{2}{*}{\textit{Kodak 18}} & 5 & 34.17 & 36.75 & 36.72 & 36.50 & \textbf{36.83} & \textbf{36.83} \\ \cline{2-8}		
 & 10 & 28.12 & 32.40 & 32.44 & 32.20 & \textbf{32.59} & \textbf{32.59} \\  \cline{2-8}			
 ($512 \times 768$) & 15 & 24.62 & 30.02 & 30.06 & 29.88 & 30.27 & \textbf{30.31}\\ \cline{2-8}	
& 20 & 22.12 & 28.42 & 28.49 & 28.35 & 28.72 & \textbf{28.77}\\
\hline
\end{tabular}
\end{center}
\end{table}

\begin{table}[t]
\caption{The denoising PSNR improvements by FRIST relative to the noisy image PSNR and denoising PSNR values using the competing methods, which are averaged over seven testing images (with the standard deviation provided after / separation) for each $\sigma$ value in Table \ref{denoisingTable}. The largest PSNR improvement using FRIST is marked in bold.}
\label{denoisingTableAver}
\begin{center}
\fontsize{10}{18pt}\selectfont
\begin{tabular}{|c|c|c|c|c|c|} 
\hline
 \multirow{2}{*}{$\sigma$} & Noisy & \multirow{2}{*}{ DCT } & \multirow{2}{*}{SST \cite{sabres3}}  & \multirow{2}{*}{K-SVD \cite{elad2}}   & \multirow{2}{*}{OCTOBOS \cite{wensabres}} \\ 
  & PSNR & &  & &   \\
\hline

 5 & $3.58\; /\; 0.98$ & $ 0.28\; / \;0.15 $ & $ 0.15\; /\;0.05 $ & $ 0.44\; /\;0.24 $ & $ 0.03\; /\;0.03 $  \\
\hline		
 10   & $ 5.75\; /\;1.30 $ & $ 0.44\; /\;0.21 $ & $ 0.23\; /\;0.08 $ & $ 0.47\; /\;0.22 $ & $ 0.04\; /\;0.03 $ \\
\hline				
 15 & $ 7.12\; /\;1.43 $ & $ 0.55\; /\;0.26 $ & $ 0.30\; /\;0.09 $ & $ 0.48\; /\;0.22 $ & $ 0.04\; /\;0.04 $ \\
\hline
 20 & $\mathbf{ 8.10}\; /\;1.45 $ & $\mathbf{ 0.64}\; /\;0.29 $ & $\mathbf{ 0.35}\; /\;0.10 $ & $\mathbf{ 0.50}\; /\;0.21 $ & $\mathbf{ 0.05}\; /\;0.03 $ \\
\hline
\end{tabular}
\end{center}
\end{table}

We present denoising results using our FRIST-based framework in Section \ref{denoisingForm}. We simulate i.i.d. Gaussian noise at 4 different noise levels ($\sigma = 5$, $10$, $15$, $20$) for seven standard images in Fig. \ref{testingIm}. Denoising results obtained by our proposed algorithm in Section \ref{denoisingForm} are compared with those obtained by the adaptive overcomplete K-SVD denoising scheme \cite{elad2}, adaptive SST denoising scheme \cite{sabres3} and the adaptive OCTOBOS denoising scheme \cite{wensabres}. We also compare to the denoising result using the SST method, but with fixed 2D DCT (i.e., no learning).

We set $K = 64$, $n = 64$, $C= 1.04$ for the FRIST denoising method. For the adaptive SST and OCTOBOS denoising methods, we follow the same parameter settings as used in the previous works \cite{sabres3, wensabres}. The same parameter settings as for the SST method is used for the DCT-based denoising algorithm. A $64 \times 256$ learned synthesis dictionary is used in the synthesis K-SVD denoising method and for the OCTOBOS denoising scheme, we use a corresponding $256 \times 64$ learned OCTOBOS. For the K-SVD, adaptive SST, and adaptive OCTOBOS denoising methods, we used the publicly available implementations \cite{ksvdweb, tlweb} in this experiment.

Table \ref{denoisingTable} lists the denoised image PSNR values for the various methods for the seven tested images at several noise levels. The proposed FRIST scheme provides consistently better PSNRs compared to the other fixed or adaptive sparse modeling methods including DCT, SST, K-SVD, and OCTOBOS. 
The average denoising PSNR improvements provided by adaptive FRIST over DCT, adaptive SST, K-SVD, and adaptive OCTOBOS are $0.48$ dB, $0.26$ dB, $0.47$ dB, and $0.04$ dB respectively, and the standard deviation in these improvements was $0.26$ dB, $0.11$ dB, $0.22$ dB, and $0.03$ dB, respectively.

\textbf{Performance vs. Noise Level}. Table \ref{denoisingTableAver} lists the noisy PSNR and denoising PSNR values averaged over $7$ testing images, using the competing methods relative to those using FRIST. It is clear for larger $\sigma$ values, the proposed FRIST provides larger average denoising PSNR improvements, compared to the competing methods.
Because the FRIST model is more constrained, thus more robust to measurement corruption (e.g., additive noise), comparing to other sparse models. Sections \ref{inpainting} and \ref{MRIexp} will provide more evidence demonstrating the robustness of the proposed FRIST method for other types of measurement corruptions, including missing pixel in image inpainting and Fourier-domain undersampling in MRI.

\begin{figure}
\begin{center}
\begin{tabular}{cc}
\includegraphics[height=1.7in]{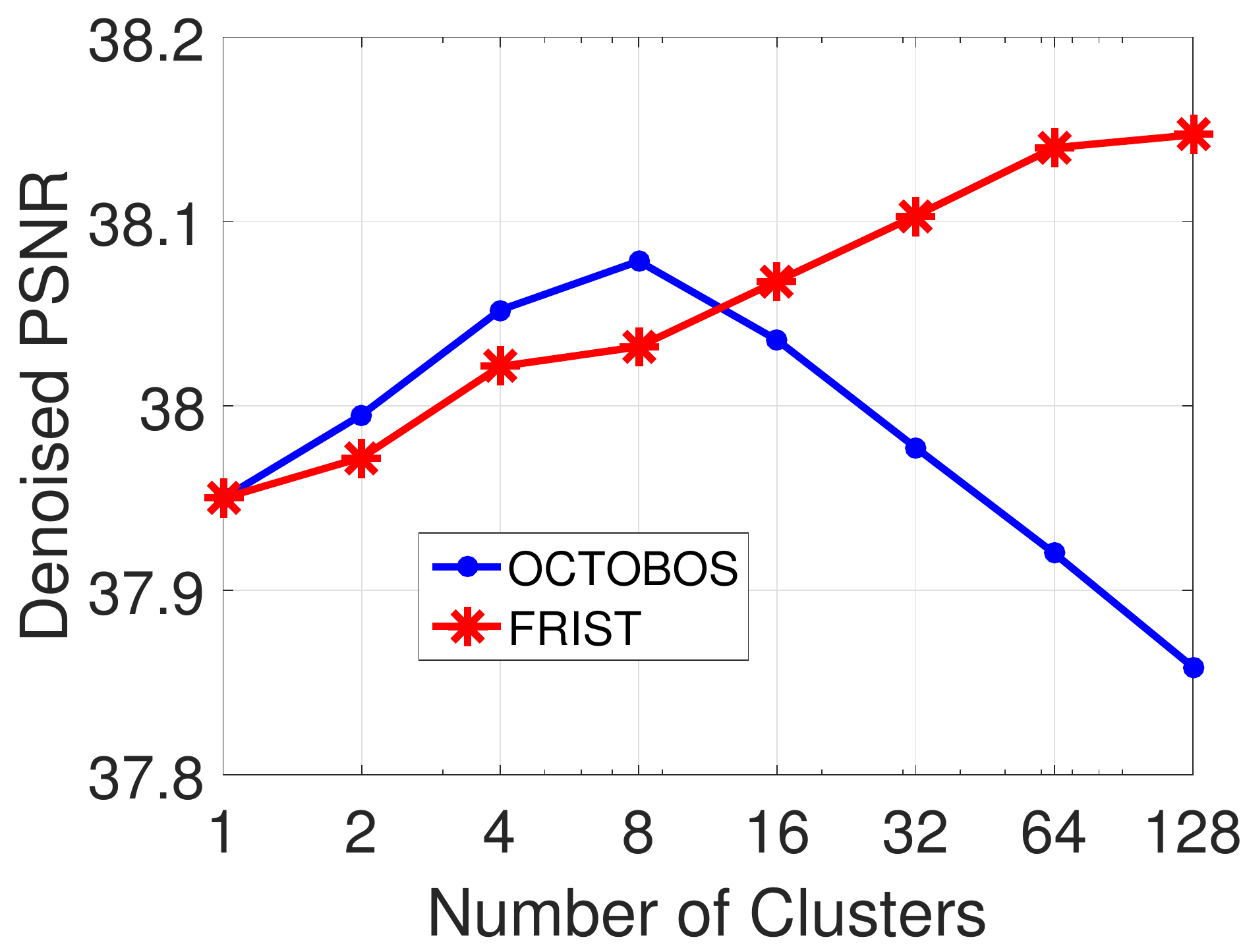}&
\includegraphics[height=1.7in]{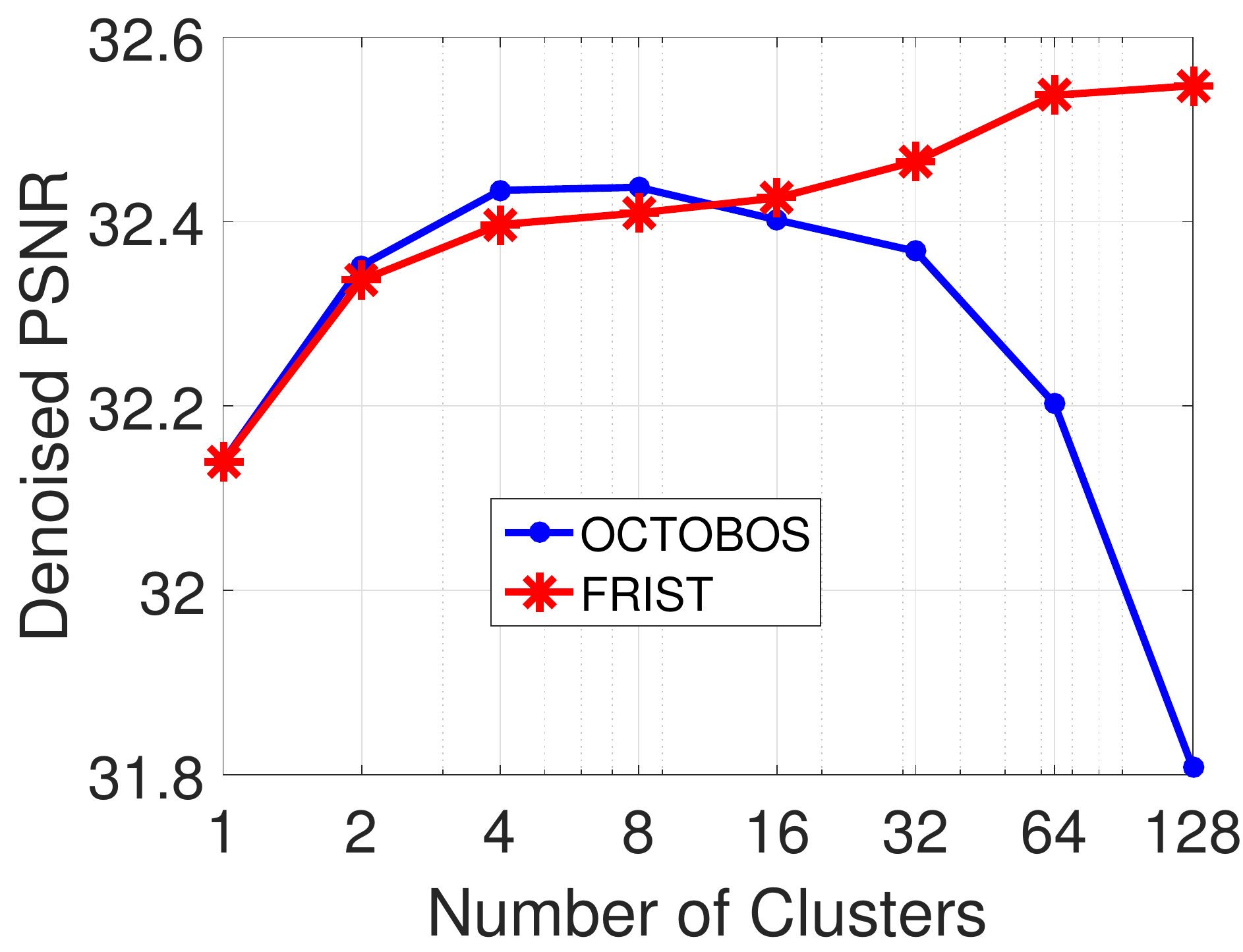}\\
(a) $\sigma = 5$ & (b) $\sigma = 15$ \\
\end{tabular}
\end{center}
\caption{Denoising PSNR for \textit{Peppers} as a function of the number of clusters (including flipping and rotations) $K$.}
\label{increaseK}
\end{figure}

\textbf{Performance vs. Number of Clusters}. In applications such as image denoising, when OCTOBOS or FRIST are learned from limited noisy patches, OCTOBOS with many more degrees of freedom is more likely to overfit the data and learn noisy features, which can degrade the denoising performance.
Figure \ref{increaseK} provides an empirical illustration of this behavior, and plots the denoising PSNRs for \textit{Peppers} as a function of the number of child transforms or number of clusters $K$ for $\sigma = 5$ and $\sigma = 15$. 
In both cases, the denoising PSNRs of the OCTOBOS and FRIST schemes increase with $K$ initially.
However, beyond an optimal value of $K$, the OCTOBOS denoising scheme suffers from overfitting the noise. 
Thus the OCTOBOS performance in Fig. \ref{increaseK} quickly degrades as the number of transforms (in the collection/union) or clusters to be learned from a set of noisy image patches is increased \cite{wensabres}. 
In contrast, the structured FRIST-based denoising scheme (involving much fewer degrees of freedom) is more robust or resilient to noise. 
As $K$ increases, adaptive FRIST denoising provides continually monotonically increasing denoising PSNR in Fig. \ref{increaseK}. 
For example, while the FRIST PSNR achieves a peak value for $K=128$, the PSNR for adaptive OCTOBOS denoising is significantly lower at such a large $K$. 


Although we focused our comparisons here on related adaptive sparse modeling methods, a very recent work \cite{wen2017strollr} shows that combining transform learning based denoising with non-local similarity models leads to better denoising performance, and outperforms the state-of-the-art BM3D denoising method \cite{dbov}. A further extension of the work in \cite{wen2017strollr} to include FRIST learning is of interest and could potentially provide even greater advantages, but we leave this detailed investigation to future work.

\subsection{Image Inpainting} \label{inpainting}

\begin{table}[t]
\caption{PSNR values for image inpainting, averaged over six images, using the proposed adaptive FRIST based method, along with the corresponding values obtained using cubic interpolation (Cubic), patch smooth ordering (Smooth), patch-based DCT, adaptive SST, and adaptive OCTOBOS based methods, for various fractions of available pixels and noise levels. The denoising PSNR improvements using FRIST relative to those using competing methods (the second row in each cell), as well as the standard deviation of these PSNR improvements (the third row, in parenthesis) are listed below. The best denoising PSNR value in each row is marked in bold.}
\label{inpaintingTable}
\begin{center}
\fontsize{11}{18pt}\selectfont
\begin{tabular}{|c|c|c|c|c|c|c|c|c|}		
\hline
\multicolumn{1}{|c|}{Avail.} & \multirow{2}{*}{$\sigma$} & Corrupted & \multirow{2}{*}{Cubic} & \multirow{2}{*}{Smooth} & \multirow{2}{*}{DCT} & \multirow{2}{*}{SST} & \multirow{2}{*}{OCTOBOS} & \multirow{2}{*}{FRIST} \\ 
    pixels  &     & PSNR &  &   &   &    &   &          \\   
\hline 

\multirow{12}{*}{20$\%$} & \multirow{3}{*}{0} & \multirow{3}{*}{7.89} & 20.35 & 27.99 & 28.32 & 28.49 & 28.60 & \multirow{3}{*}{\textbf{28.65}} \\ 
\cline{4-8}
	&  &  &  8.30  &  0.66 &  0.33 &  0.16 &  0.05 &  \\ 
		&  &  & (1.34) & (0.30) & (0.12) & (0.06) & (0.03) &  \\
\cline{2-9}	
 & \multirow{3}{*}{5} & \multirow{3}{*}{7.89} & 20.02 & 27.86 & 28.26 & 28.44 & 28.53 & \multirow{3}{*}{\textbf{28.61}} \\ 
\cline{4-8}
	&  &  &  8.59  &  0.75 &  0.35 &  0.17 &  0.08 &  \\ 
		&  &  & (1.37) & (0.38) & (0.13) & (0.06) & (0.04) &  \\ 
\cline{2-9}	
 & \multirow{3}{*}{10} & \multirow{3}{*}{7.88} & 19.37 & 26.46 & 27.46 & 27.98 & 28.25 & \multirow{3}{*}{\textbf{28.41}} \\ 
\cline{4-8}
	&  &  &  9.04 &  1.95 &  0.95 &  0.43 &  0.16 &  \\ 
		&  &  & (1.41) & (0.53) & (0.42) & (0.15) & (0.09) &  \\ 
\cline{2-9}	
  & \multirow{3}{*}{15} & \multirow{3}{*}{7.87} & 18.54 & 25.02 & 26.60 & 27.38 & 27.71 & \multirow{3}{*}{\textbf{27.92}} \\
\cline{4-8}
	&  &  &   9.38 &  2.90 &  1.32 &  0.54 &  0.21 &  \\ 
		&  &  & (1.42) & (0.66) & (0.51) & (0.18) & (0.10) &  \\ 
\hline
\multirow{12}{*}{10$\%$} & \multirow{3}{*}{0} & \multirow{3}{*}{7.37} & 19.19 & 24.87 & 25.21 & 25.25 & 25.25 & \multirow{3}{*}{\textbf{25.31}} \\ 
\cline{4-8}
	&  &  &  6.12 &  0.44 &  0.10 &  0.06 &  0.06 &  \\ 
		&  &  & (1.02) & (0.20) & (0.04) & (0.03) & (0.03) &  \\ 
\cline{2-9}	
 & \multirow{3}{*}{5} & \multirow{3}{*}{7.37} & 18.90 & 24.81 & 24.98 & 25.19 & 25.30 & \multirow{3}{*}{\textbf{25.38}} \\
\cline{4-8}
	&  &  &  6.48 &  0.57 &  0.40 &  0.19 &  0.08 &  \\ 
		&  &  & (1.09) & (0.24) & (0.19) & (0.09) & (0.04) &  \\  
\cline{2-9}	
 & \multirow{3}{*}{10} & \multirow{3}{*}{7.37} & 18.46 & 24.10 & 24.40 & 24.44 & 24.69 & \multirow{3}{*}{\textbf{24.80}} \\ 
\cline{4-8}
	&  &  &  6.34 &  0.70 &  0.40 &  0.36 &  0.11 &  \\ 
		&  &  & (1.10) & (0.29) & (0.18) & (0.15) & (0.05) &  \\ 
\cline{2-9}	
  & \multirow{3}{*}{15} & \multirow{3}{*}{7.36} & 17.88 & 23.28 & 23.62 & 23.87 & 24.11 & \multirow{3}{*}{\textbf{24.22}} \\ 
	\cline{4-8}
	&  &  &   6.34 &  0.94 &  0.60 &  0.35 &  0.11 &  \\ 
		&  &  &  (1.13) & (0.35) & (0.27) & (0.19) & (0.07) &  \\ 
\hline
\end{tabular}
\end{center}
\end{table}

\begin{figure}
\begin{center}
\begin{tabular}{ccc}
\includegraphics[height=1.9in]{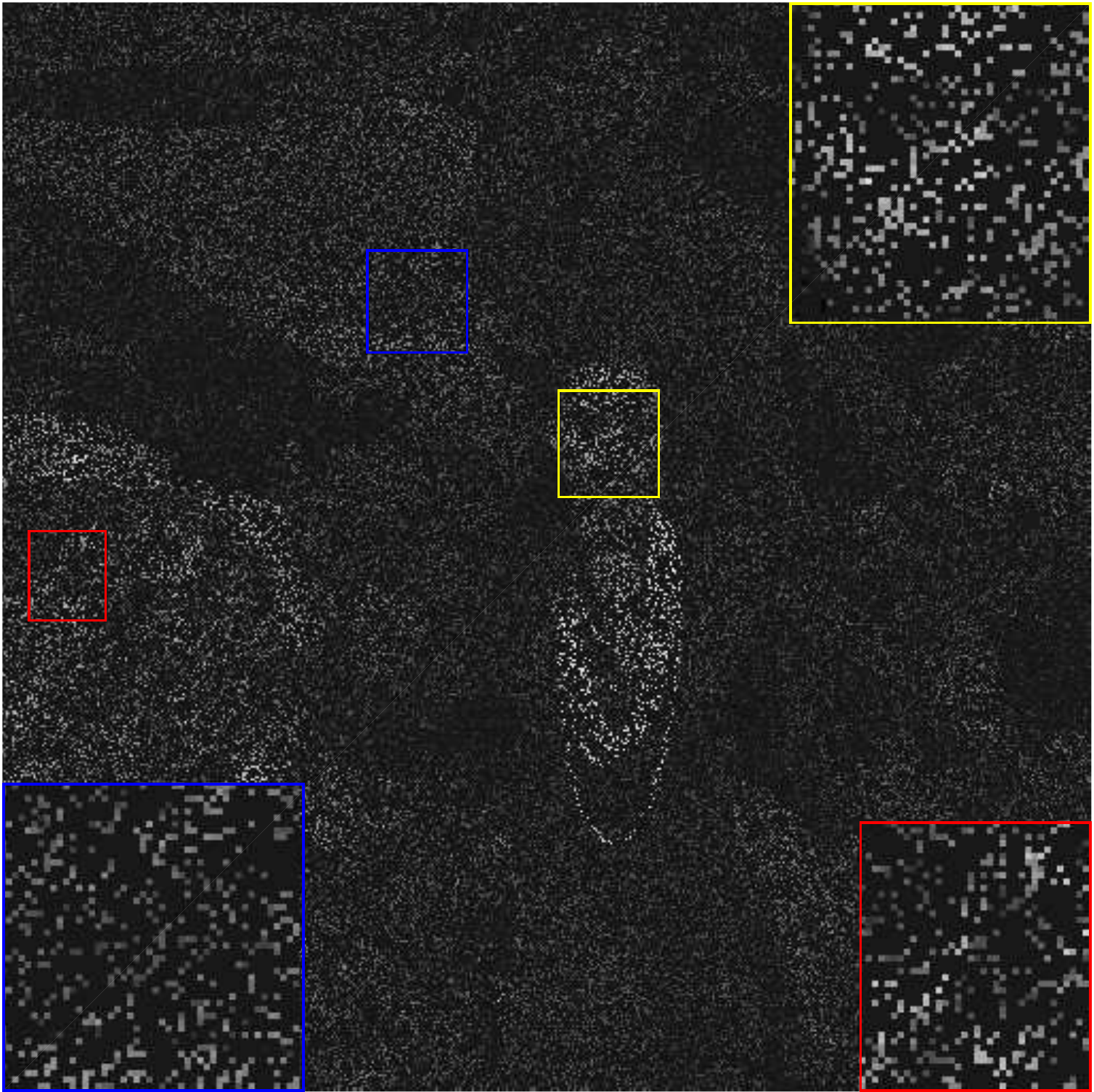}&
\includegraphics[height=1.9in]{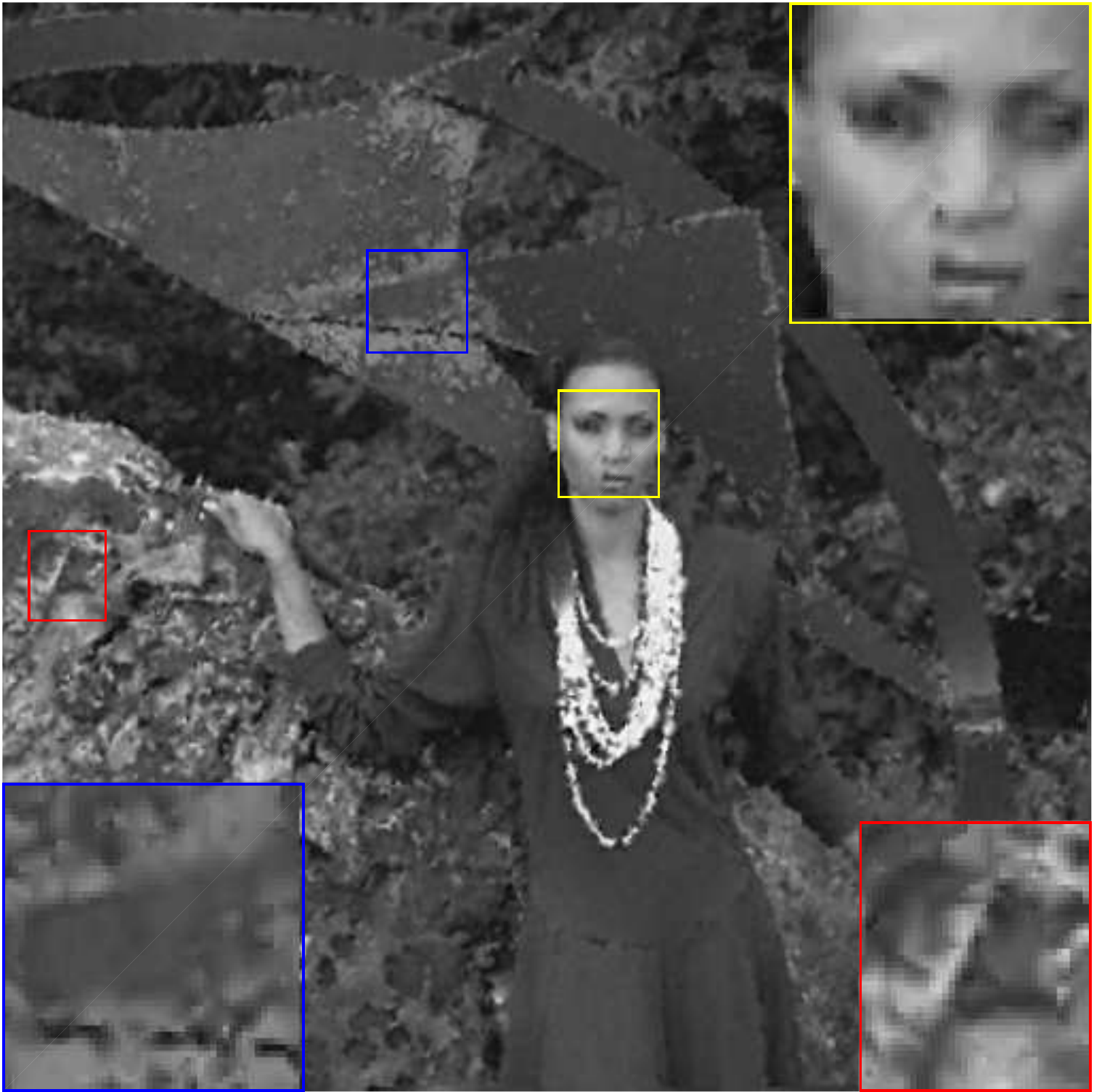}&
\includegraphics[height=1.9in]{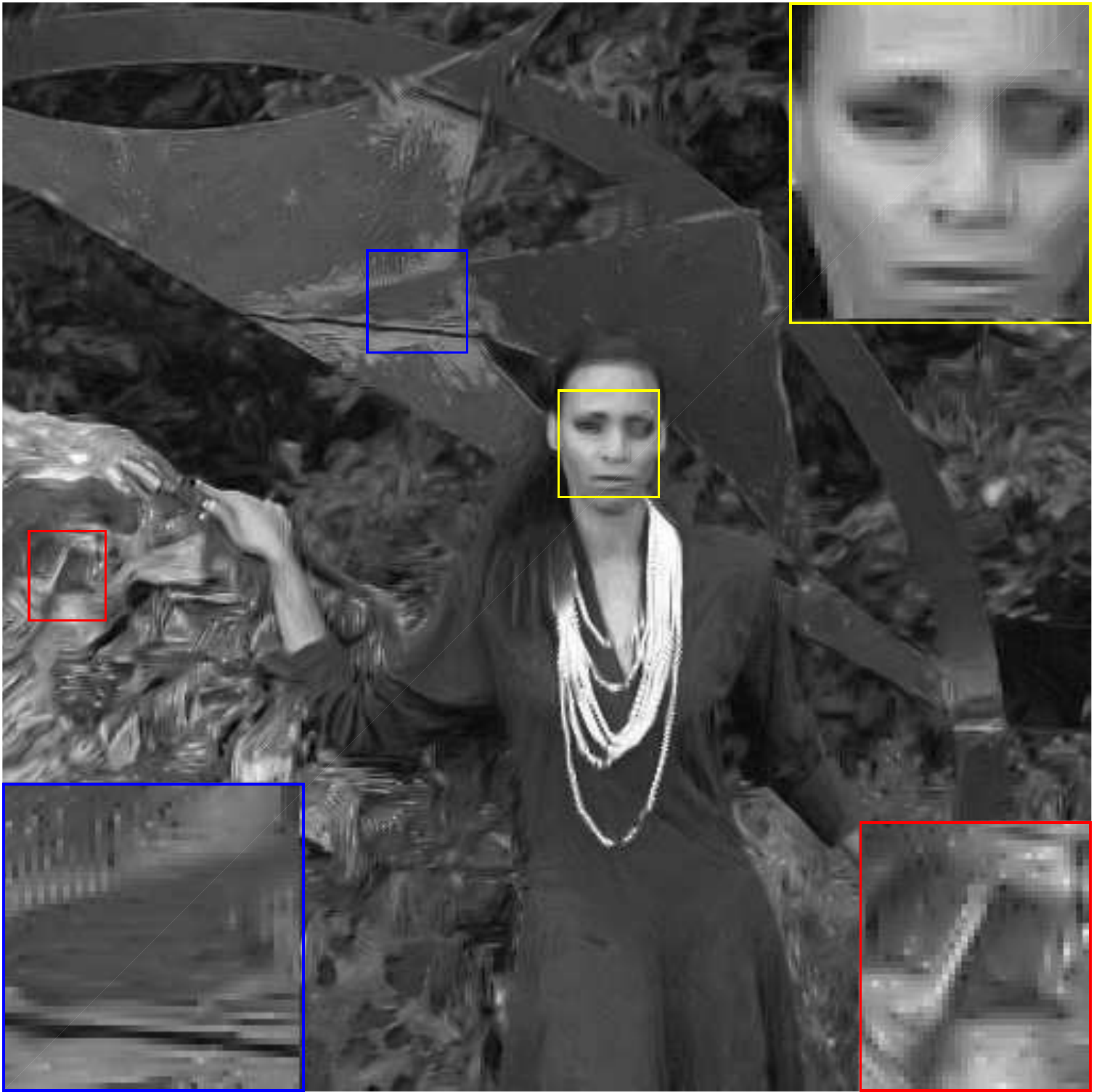}\\
Corrupted & Cubic (24.58 dB) & Smoothing (25.02 dB)\\
\includegraphics[height=1.9in]{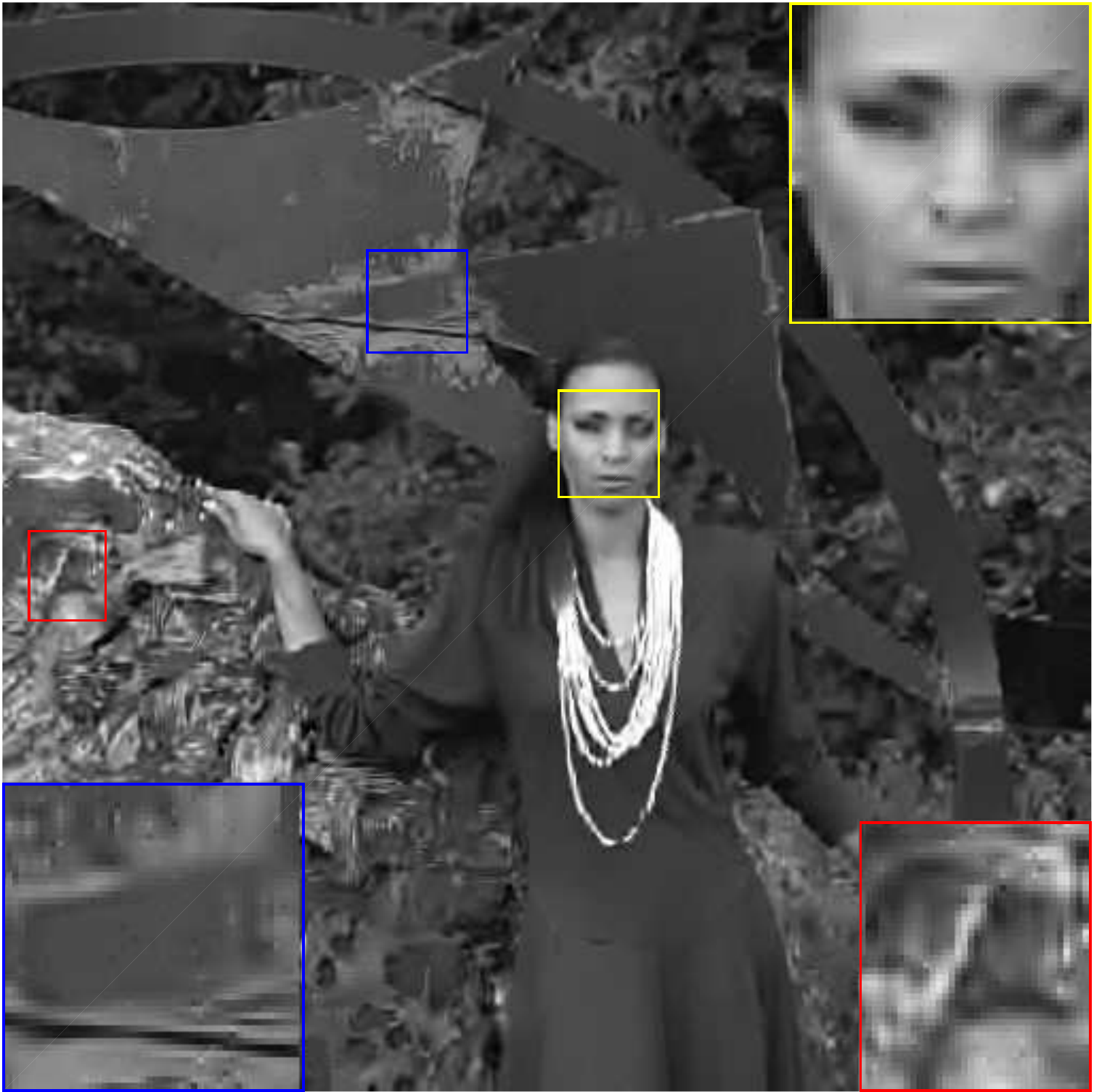}&
\includegraphics[height=1.9in]{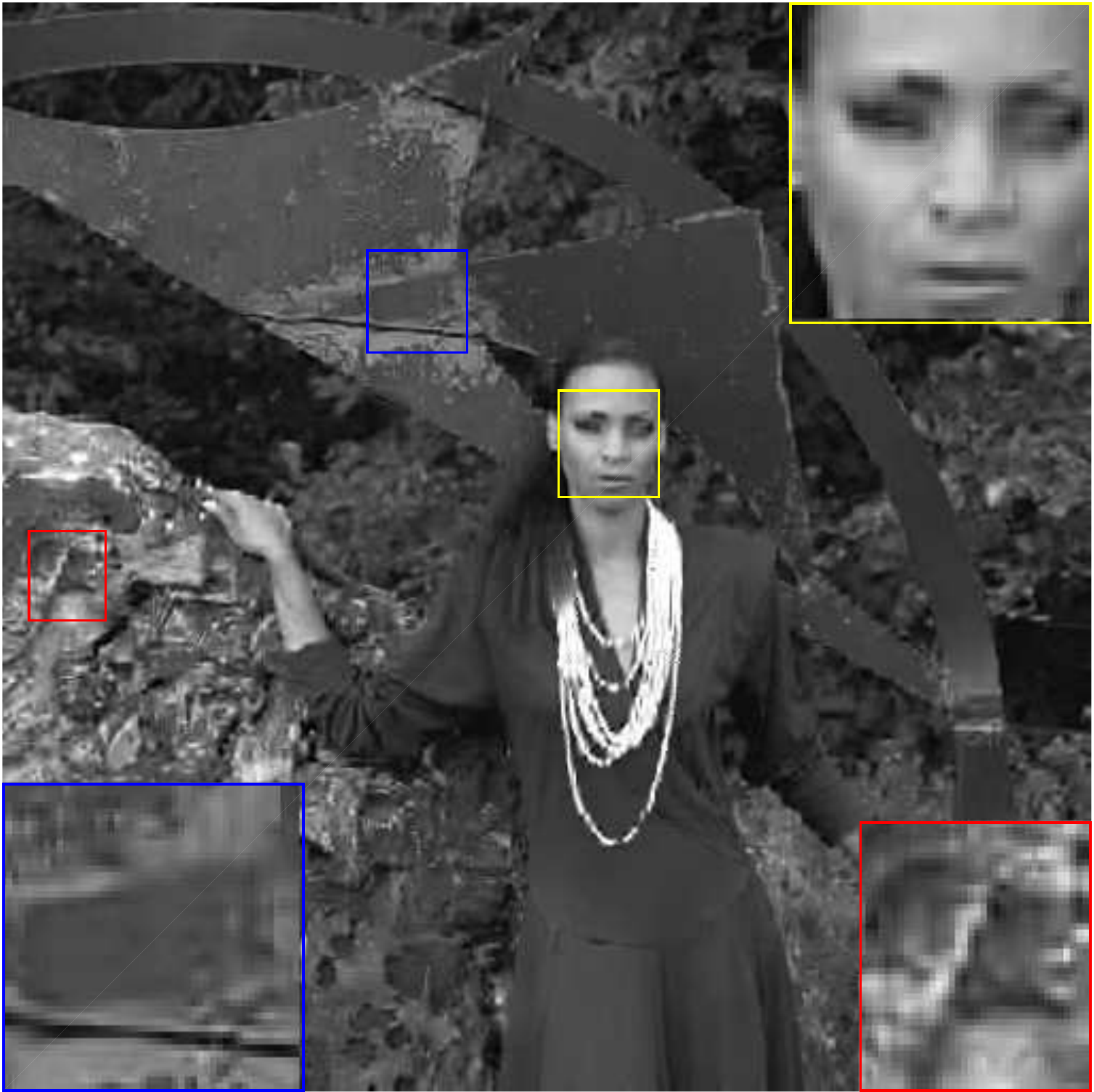}&
\includegraphics[height=1.9in]{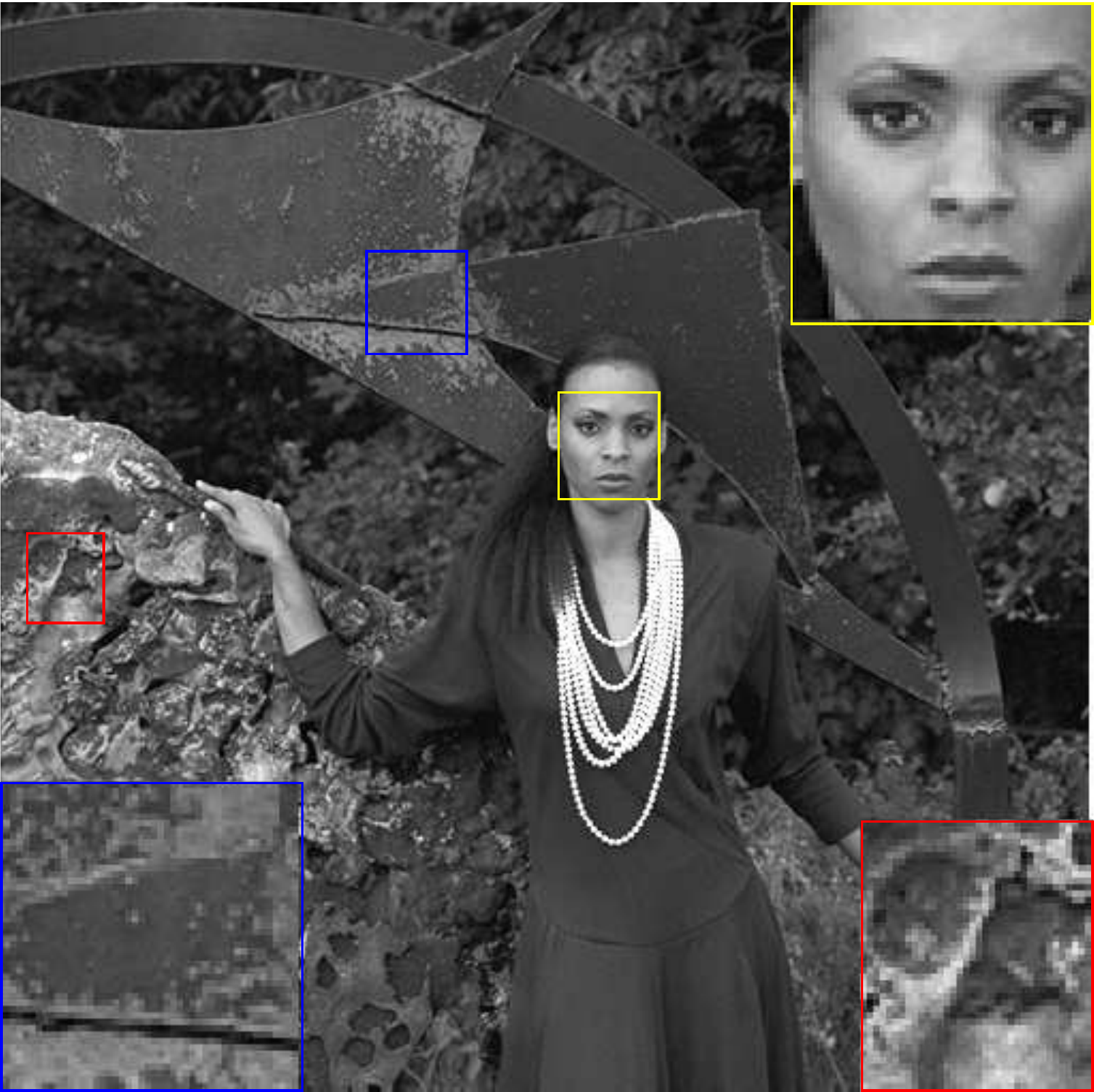}\\
 SST (25.22 dB) & FRIST (\textbf{25.41 dB}) & Original \\
\end{tabular}
\end{center}
\caption{Illutration of image inpainting results for the \textit{Kodak 18} (80\% pixels are missing), with regional zoom-in comparisons.}
\label{visual}
\end{figure}

We present preliminary results for our adaptive FRIST-based inpainting framework (based on $(\mathrm{P6})$). 
We randomly remove $80\%$ and $90\%$ of the pixels of the entire images in Fig. \ref{exp}, and simulate i.i.d. additive Gaussian noise for the sampled pixels with $\sigma = 0$, $5$, $10$, and $15$. 
We set $K = 64$, $n = 64$, and apply the proposed adaptive FRIST inpainting algorithm to reconstruct the images from the corrupted and noisy measurements. 
For comparison, we replace the adaptive FRIST in the proposed inpainting algorithm with the fixed 2D DCT, adaptive SST \cite{sabres3}, and adaptive OCTOBOS \cite{wensabres} respectively, and evaluate the inpainting performance these alternatives.
The image inpainting results obtained by the FRIST based methods are also compared with those obtained by the cubic interpolation \cite{cubicInpaint1, cubicInpaint2} and patch smoothing \cite{eladInpainting2013} methods. We used the Matlab function ``griddata" to implement the cubic interpolation, and use the publicly available implementation of the patch smoothing method. For the DCT, SST, OCTOBOS, and FRIST based methods, we initialize the image patches using the Cubic Interpolation method in noiseless cases, and using the Patch Smoothing method in noisy cases.

Table \ref{inpaintingTable} lists the image inpainting PSNR results, averaged over the images shown in Fig. \ref{testingIm}, for various fractions of sampled pixels and noise levels. The proposed adaptive FRIST inpainting scheme provides better PSNRs compared to the other inpainting methods based on interpolation, transform-domain sparsity, and spatial similarity. 
The average inpainting PSNR improvements achieved by FRIST over DCT, SST, and OCTOBOS are $0.56$ dB, $0.28$ dB, and $0.11$ dB respectively, and the standard deviations in these improvements were $0.39$ dB, $0.16$ dB, and $0.05$ dB respectively.
Importantly, adaptive FRIST provides larger improvements over the other competing methods including the learned OCTOBOS, at higher noise levels.
Figure \ref{visual} provides an illustration of the inpainting results, with regional zoom-in for visual comparisons. 
We observe that the cubic interpolation produces blur in various locations. The FRIST result is much improved, and also shows fewer artifacts compared to the patch smoothing \cite{eladInpainting2013} and adaptive SST results.
Table \ref{inpaintingTable} shows that the cubic Interpolation method is extremely sensitive to noise, whereas the FRIST based method is the most robust. 
These results indicate the benefits of adapting the highly constrained yet overcomplete FRIST data model.

\subsection{MRI Reconstruction} \label{MRIexp}

\begin{figure}
\begin{center}
\begin{tabular}{cccc}
\includegraphics[height=1.4in]{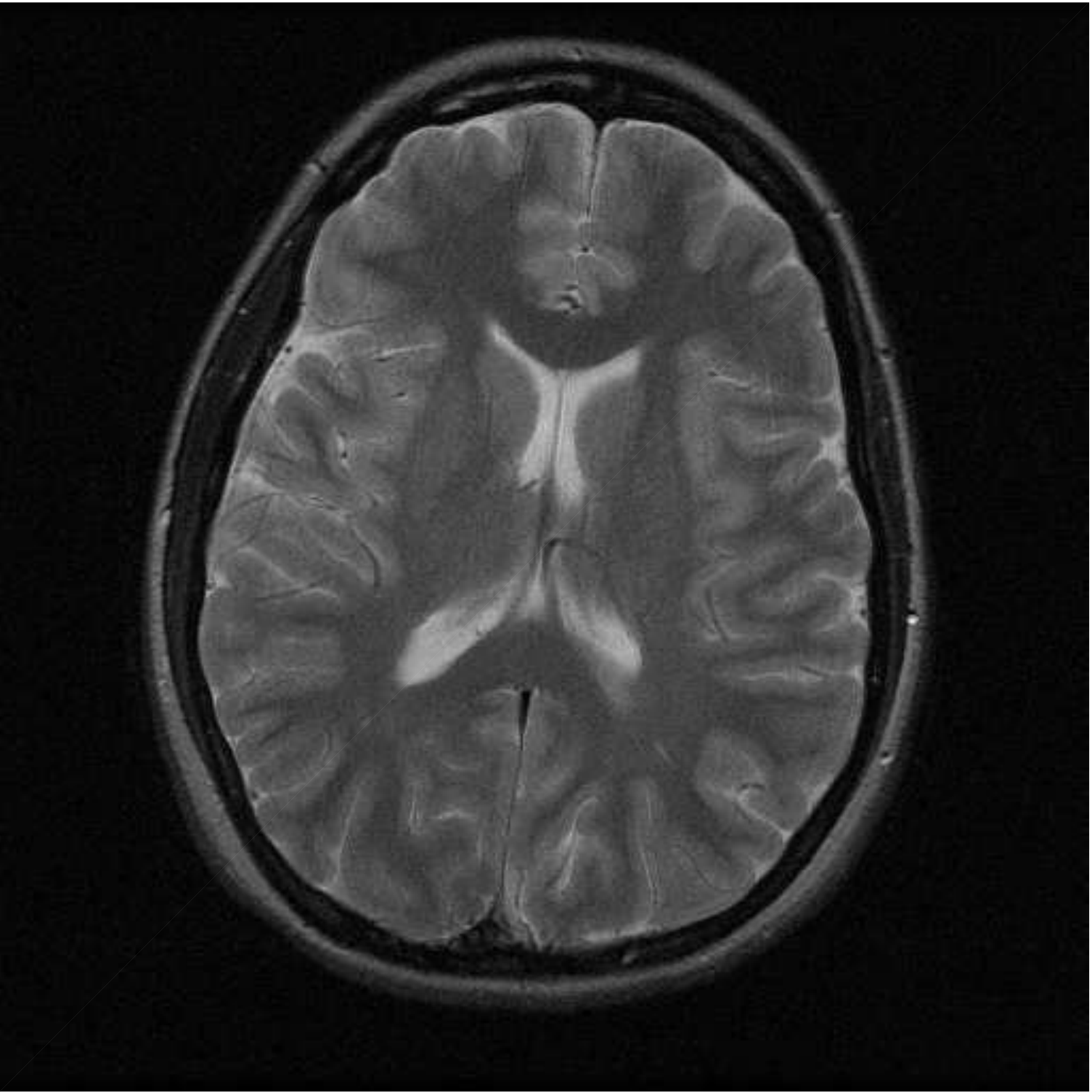}&
\includegraphics[height=1.4in]{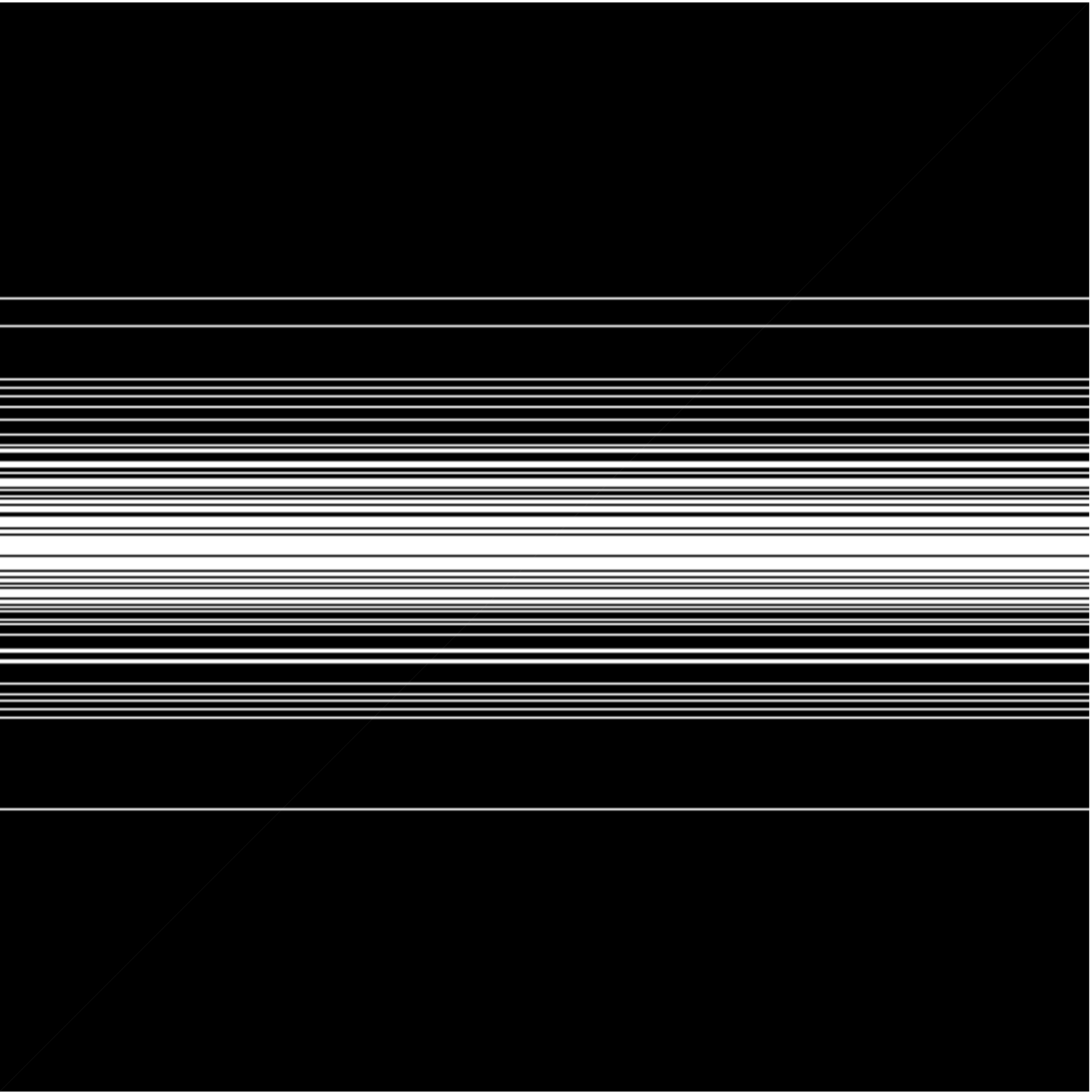}&
\includegraphics[height=1.4in]{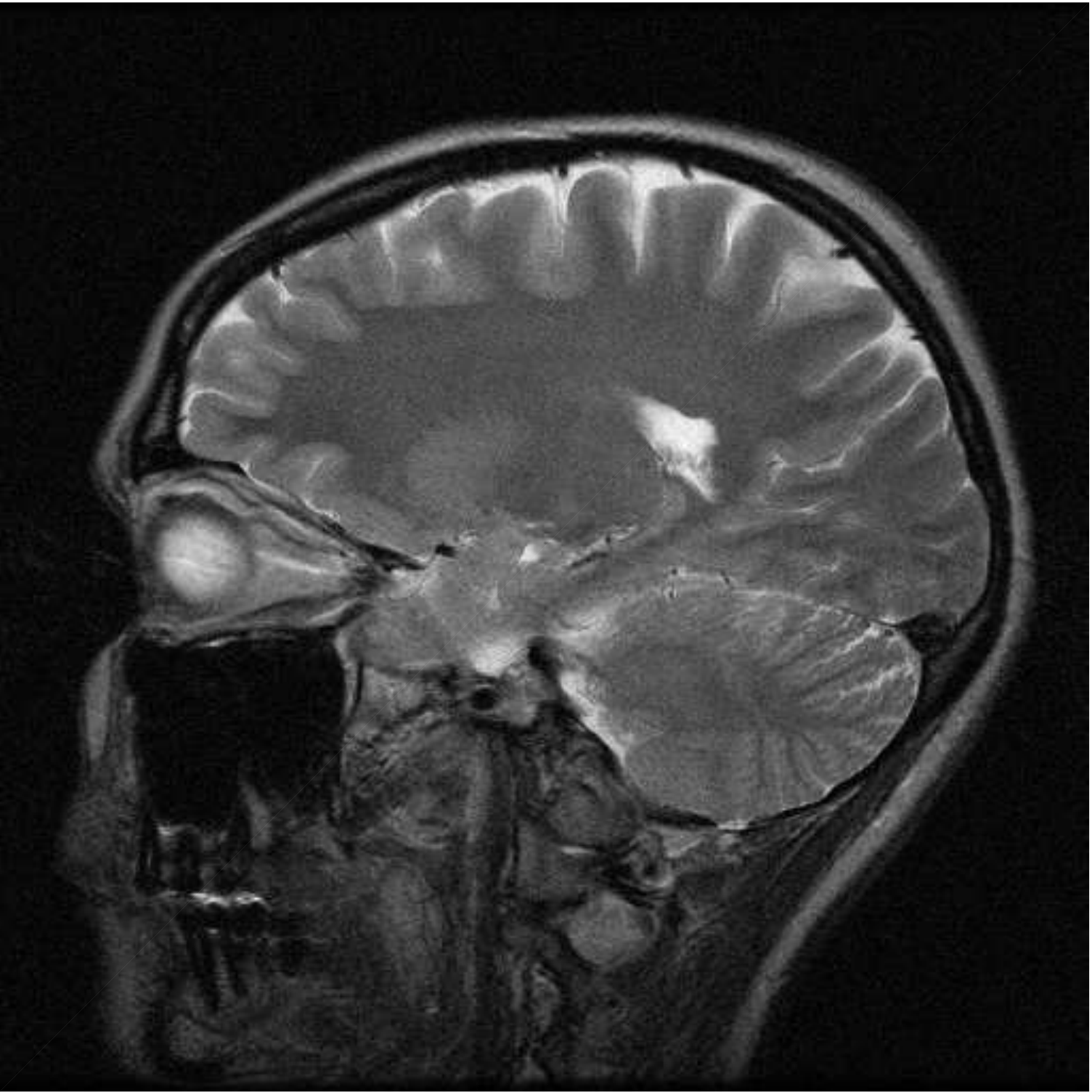}&
\includegraphics[height=1.4in]{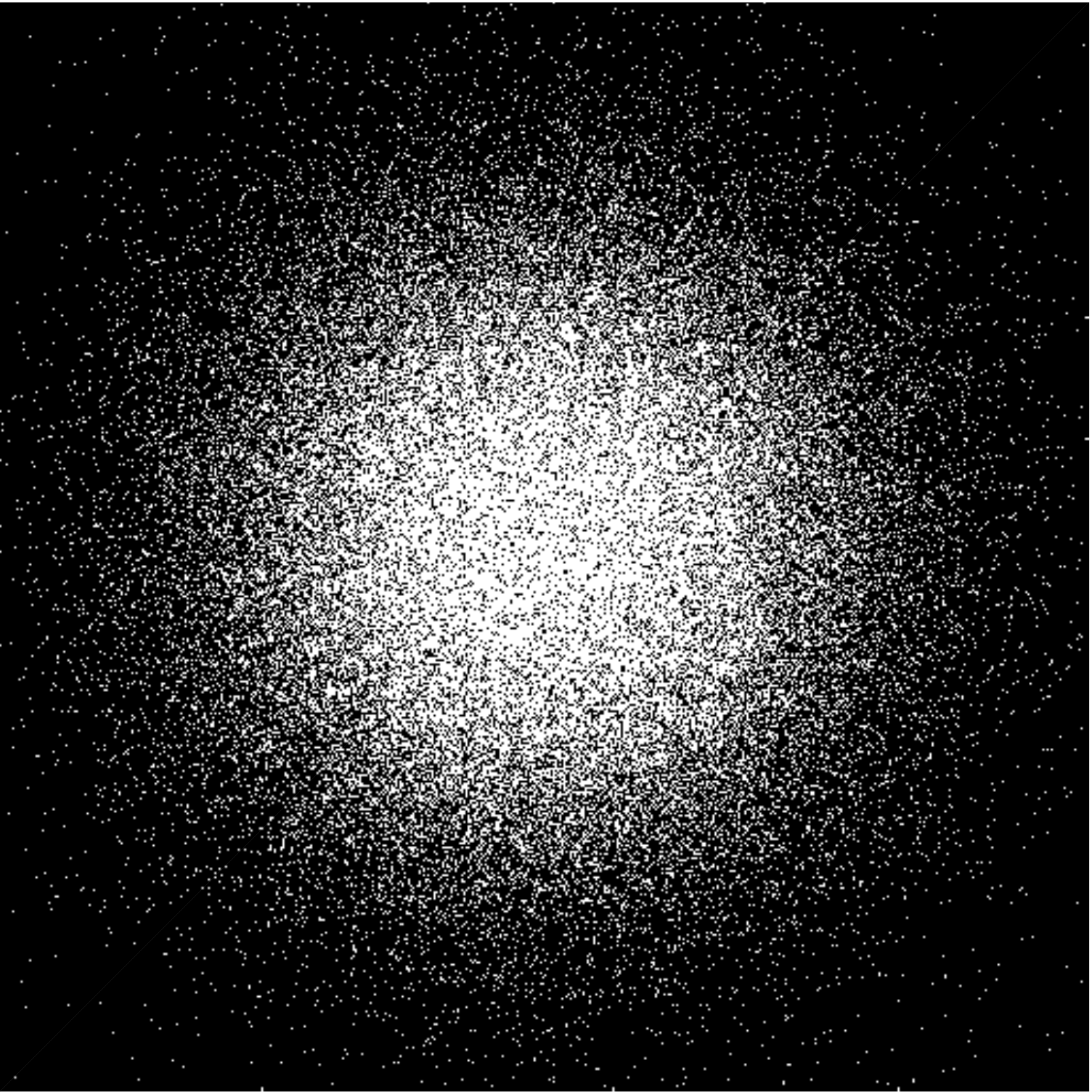}\\
(a) & (b) & (c) & (d) \\
\end{tabular}
\end{center}
\caption{Testing MRI images and their k-space sampling masks: (a) Image \textit{\textbf{1}}; (b) k-space sampling mask (Cartesian with $7 \times$ undersampling) for Image \textit{\textbf{1}}; (c) Image \textit{\textbf{2}}; (d) k-space sampling mask (2D random with $5 \times$ undersampling) for Image \textit{\textbf{2}}. }
\label{MRIimages}
\end{figure}

\begin{figure}
\begin{center}
\begin{tabular}{ccc}
\includegraphics[height=1.7in]{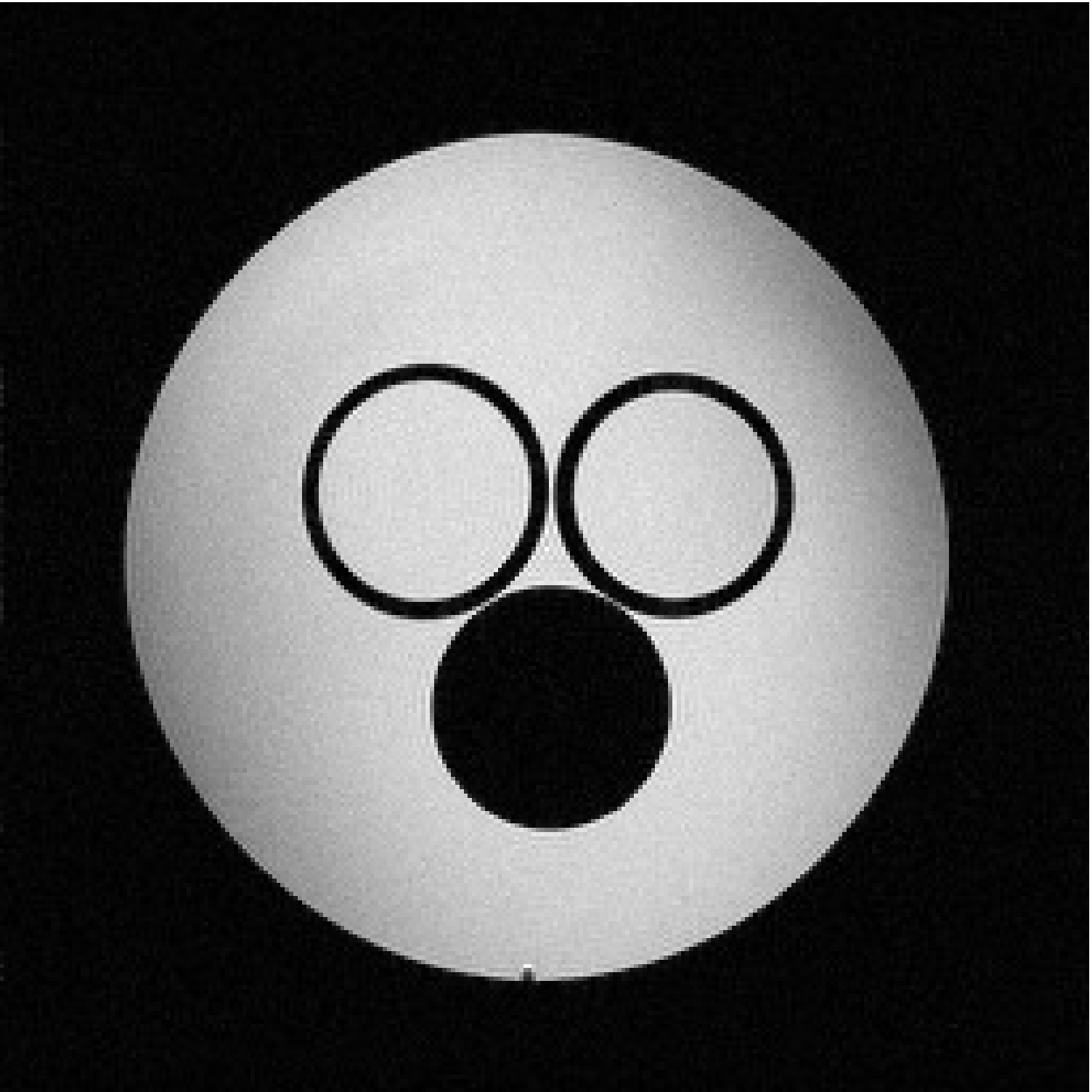}&
\includegraphics[height=1.7in]{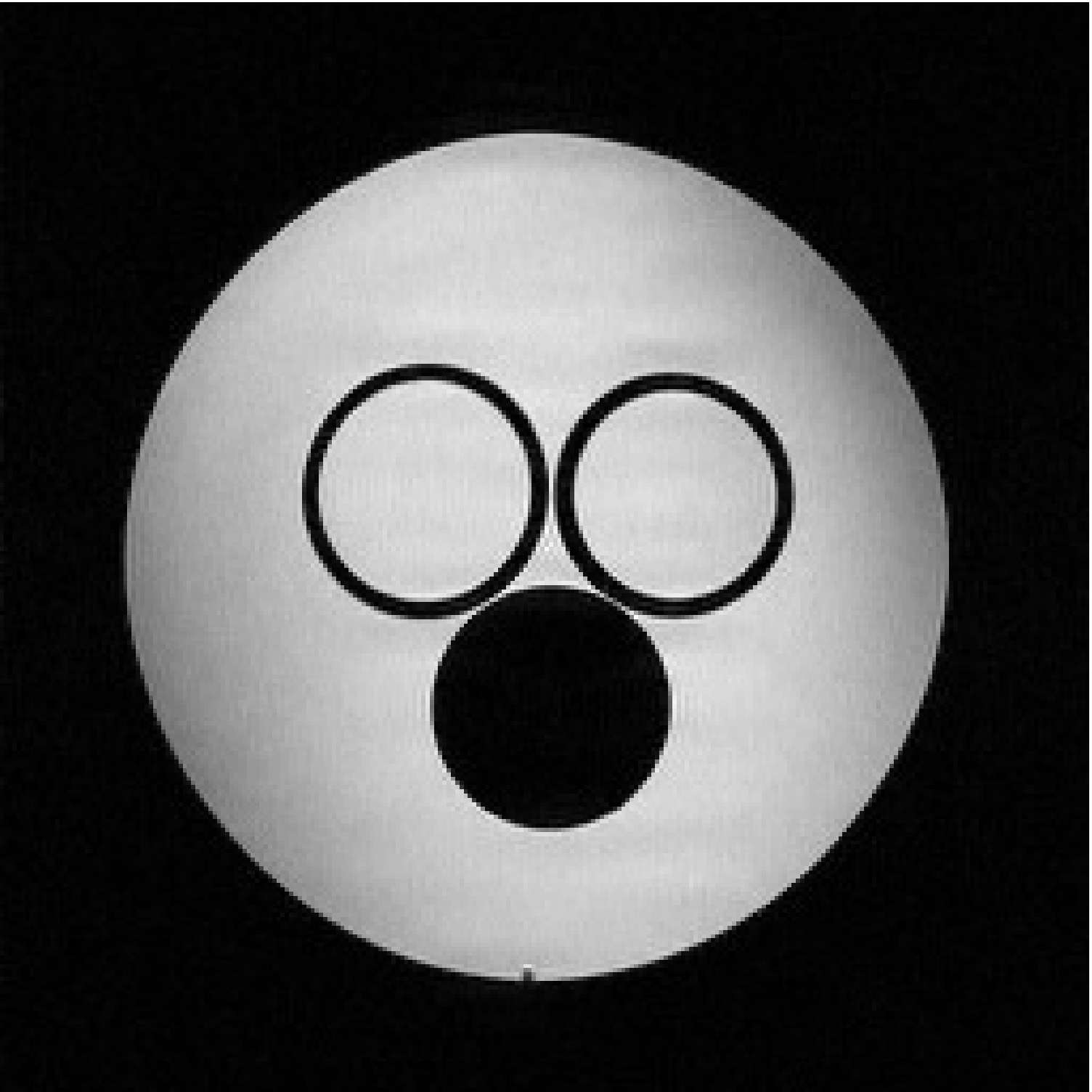}&
\includegraphics[height=1.7in]{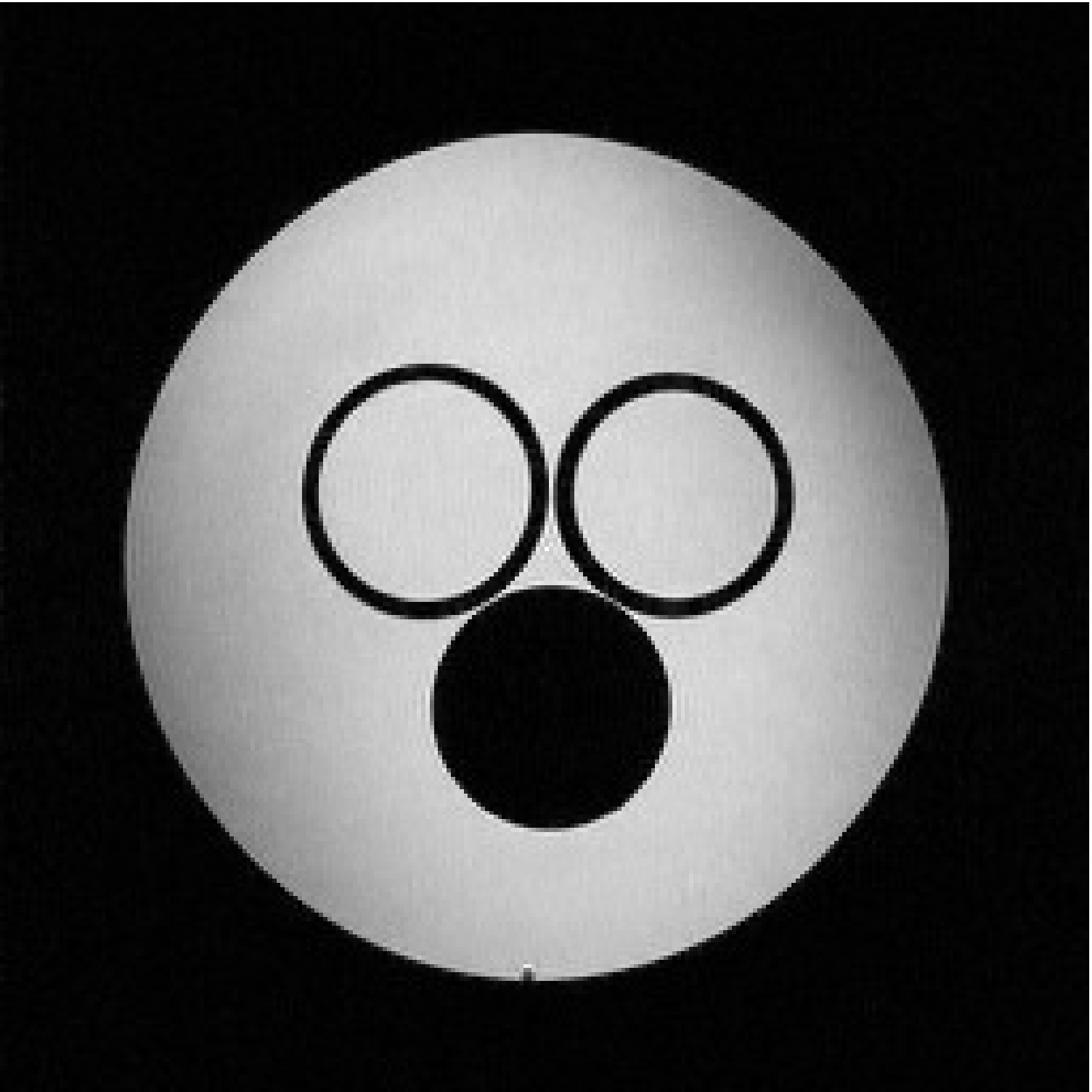}\\
(a) & (c) & (e)  \\
\includegraphics[height=1.7in]{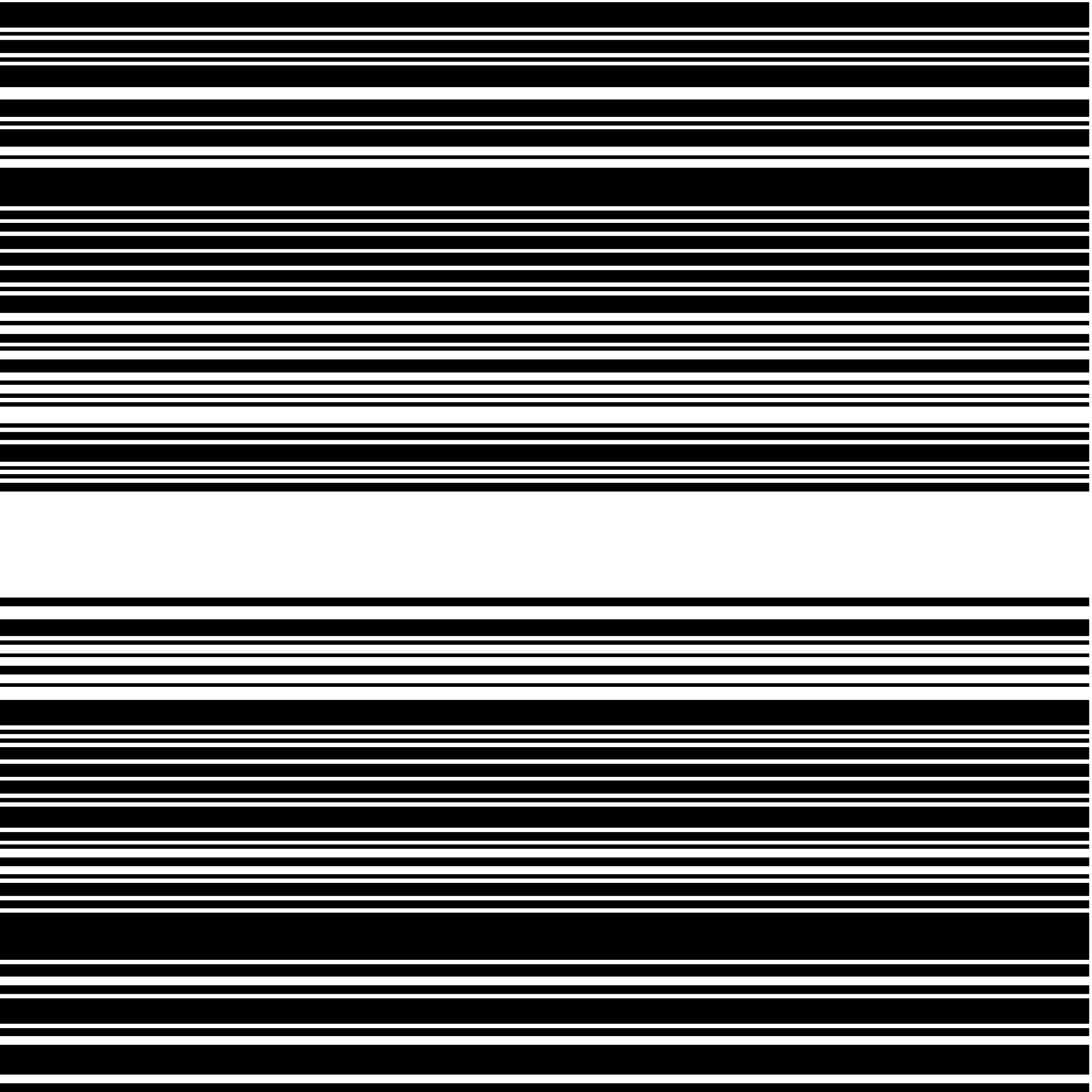}&
\includegraphics[height=1.8in]{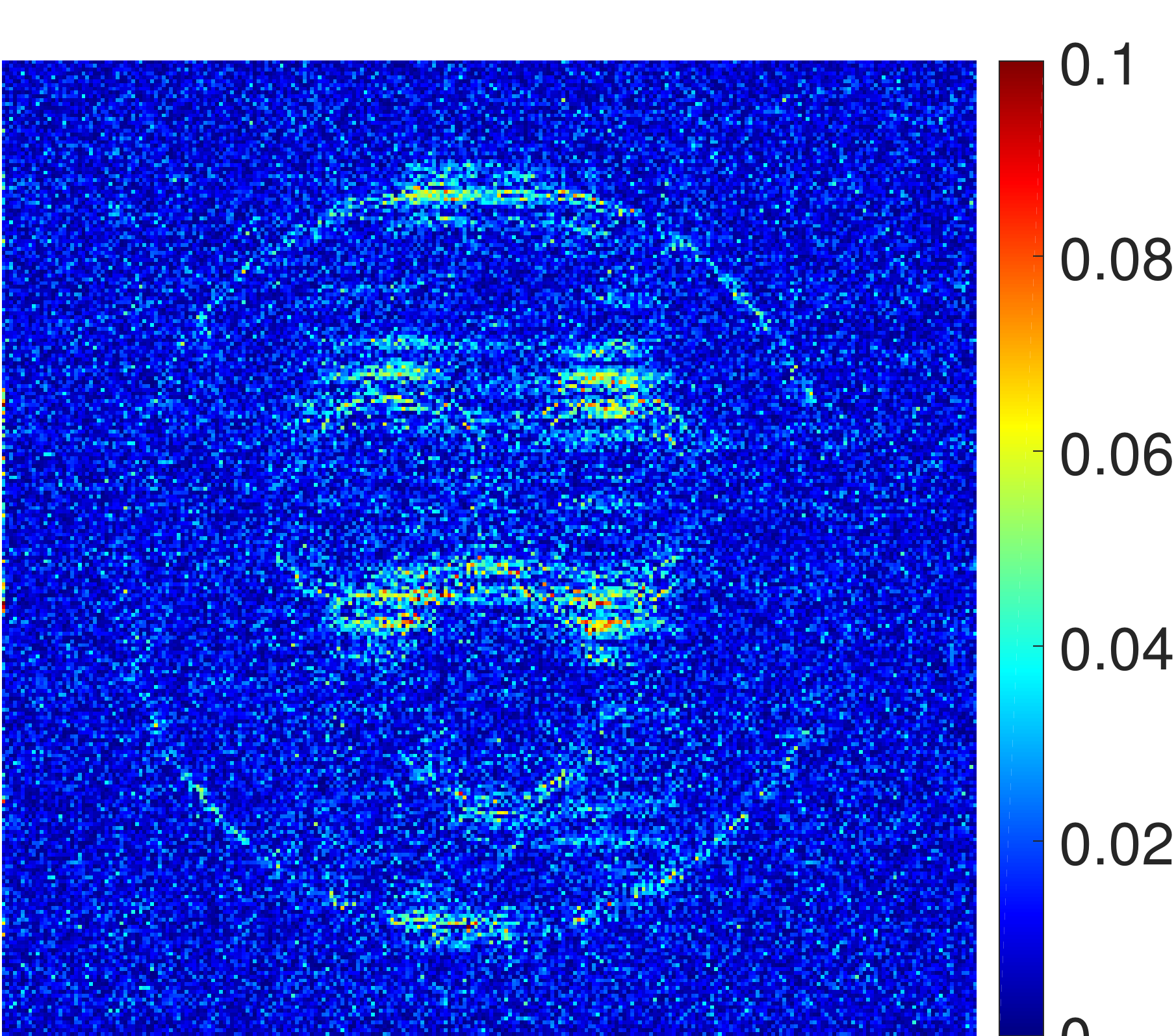}&
\includegraphics[height=1.8in]{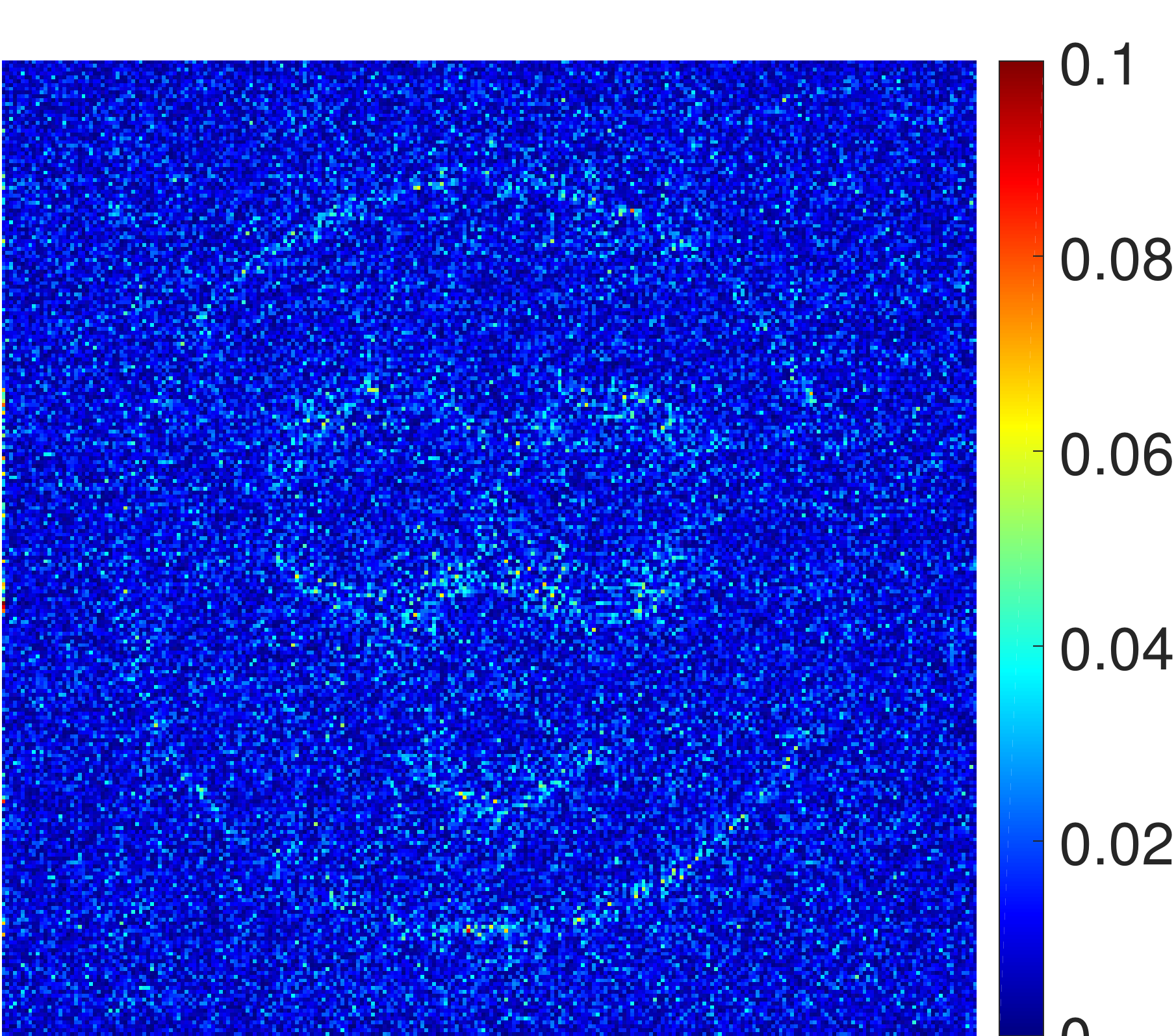}\\
(b) & (d) & (f) \\
\end{tabular}
\end{center}
\caption{Visualization of the reconstruction of Image \textit{\textbf{3}} using Cartesian sampling and $2.5 \times$ undersampling: (a) Image \textit{\textbf{3}}; (b) sampling mask in k-space; (c) TL-MRI reconstruction ($36.3$ dB); (d) magnitude of TL-MRI reconstruction error; (e) FRIST-MRI reconstruction ($36.7$ dB); and (f) magnitude of FRIST-MRI reconstruction error.}
\label{visualMRI}
\end{figure}

We present preliminary MRI reconstruction results using the proposed FRIST-MRI algorithm. 
The three complex-valued images and the corresponding k-space sampling masks used in this section are shown in Fig. \ref{MRIimages}, Fig. \ref{visualMRI}(a), and Fig. \ref{visualMRI}(b)  \footnote{The testing image data in this section were used and included in previous works \cite{syber, zhan2015fast} with the data sources.}. 
We retrospectively undersample the k-space of the reference images using the displayed sampling masks.
We set $K = 32$, the sparsity level $s = 0.05 \times nP$, and the other parameters were set similarly as for TL-MRI in \cite{syber}. 
We used a higher sparsity level $s = 0.085 \times nN$ for reconstructing Image \textit{\textbf{3}}, which worked well. 
To speed up convergence, lower sparsity levels are used in the initial iterations \cite{syber}. 
We compare our FRIST-MRI reconstruction results to those obtained using conventional or popular methods, including naive Zero-filling, Sparse MRI \cite{sparseMRI}, DL-MRI \cite{dlmri}, PBDWS \cite{pbdws}, PANO \cite{pano}, and TL-MRI \cite{syber}. 
The parameter settings for these methods are as mentioned in \cite{syber}. 
We separately tuned the sparsity parameter for TL-MRI \cite{syber} for reconstructing Image \textit{\textbf{3}} \footnote{We observed improved reconstruction PSNR compared to the result obtained using the sparsity settings in \cite{syber}.}. The reconstruction PSNRs (computed for image magnitudes) for various approaches are compared in Table \ref{mritable}.

\begin{table}[t]
\caption{Comparison of the PSNRs, corresponding to the Zero-filling, Sparse MRI \cite{sparseMRI}, DL-MRI \cite{dlmri}, PBDWS \cite{pbdws}, PANO \cite{pano}, TL-MRI \cite{syber}, and the proposed FRIST-MRI reconstructions for various images, sampling schemes, and undersampling factors. The best PSNR for each MRI image is marked in bold.}
\label{mritable}
\begin{center}
\fontsize{11}{17pt}\selectfont
\begin{tabular}{|c|c|c|c|c|c|c|c|c|c|}
\hline
\multirow{2}{*}{Image}  &  Sampling & Under- & Zero- & Sparse & DL- & \multirow{2}{*}{PBDWS} & \multirow{2}{*}{PANO} & TL- & FRIST-\\
 & Scheme & sampl. & filling & MRI & MRI & & & MRI & MRI \\
 \hline
\textit{\textbf{1}} & Cartesian & $7 \times $ & $27.9$ & $28.6$ & $30.9$ & $31.1$ & $31.1$ & $31.2$ & \textbf{31.4} \\  \hline
\textit{\textbf{2}} & 2D Random & $5 \times $ & $26.9$ & $27.9$ & $30.5$ & $30.3$ & $30.4$ & $30.6$ & \textbf{30.7} \\  \hline
\textit{\textbf{3}} & Cartesian & $2.5 \times $ & $24.9$ & $29.9$ & $36.6$ & $35.8$ & $34.8$ & $36.3$ & \textbf{36.7} \\  \hline
\end{tabular}
\end{center}
\end{table}

First, the proposed FRIST-MRI algorithm provides significant improvements over the naive Zero-filling reconstruction (the initialization of the algorithm) with $6.4$ dB better PSNR on average, as well as $4.2$ dB better PSNR (on average) over the non-adaptive Sparse MRI reconstructions. 
Compared to recently proposed popular MRI reconstruction methods, the FRIST-MRI algorithm demonstrates reasonably better performance for each testing case, with an average PSNR improvement of $0.8$ dB, $0.5$ dB, and $0.3$ dB over the non-local patch similarity-based PANO method, the partially adaptive PBDWS method, and the adaptive dictionary-based DL-MRI method. 

The proposed FRIST-MRI reconstruction quality is $0.2$ dB better than TL-MRI on average. 
As we followed a reconstruction framework and parameters similar to those used by TL-MRI \cite{syber}, the quality improvement obtained with FRIST-MRI is solely because the learned FRIST can serve as a better regularizer for MR image reconstruction compared to the single adaptive square transform in TL-MRI. 
Figure \ref{visualMRI} visualizes the reconstructions and reconstruction errors (magnitude of the difference between the magnitudes of the reconstructed and reference images) for FRIST-MRI and TL-MRI. 
The FRIST-MRI reconstruction error map clearly shows fewer artifacts, especially along the boundaries of the circles, compared to TL-MRI.

\section{Conclusion} \label{cond}

In this paper, we presented a novel framework for learning flipping and rotation invariant sparsifying transforms. 
These transforms correspond to a structured union-of-transforms, and are dubbed FRIST. The collection of transforms in FRIST are related to an underlying (or generating) parent transform by flipping and rotation operations. 
Our algorithm for FRIST learning is highly efficient, and involves optimal updates, with convergence guarantees. 
We demonstrated the ability of FRIST learning for extracting directional features in images. 
In practice, FRIST learning is insensitive to initialization, and performs better than several prior adaptive sparse modeling methods in various applications including sparse image representation, image denoising, image inpainting, and blind compressed sensing MRI reconstruction.

\section*{References}
\bibliographystyle{plain}
\bibliography{iopart-num}

\end{document}